\documentclass[11pt]{article}

\usepackage[utf8]{inputenc}
\usepackage[T1]{fontenc}

\usepackage[a4paper,margin=1in]{geometry}
\usepackage[numbers]{natbib}

\usepackage{amsmath}
\usepackage{amsfonts}
\usepackage{algorithm}
\usepackage{array}
\usepackage{algpseudocode}
\usepackage{multirow}
\usepackage{bm}
\usepackage{url}
\usepackage{mathtools}
\usepackage{graphicx}
\usepackage{booktabs}
\usepackage{authblk}
\usepackage{hyperref}
\usepackage{hyphenat}
\usepackage{adjustbox}
\usepackage{tabularx}
\usepackage{makecell}
\usepackage{pdflscape}

\newcommand{\mask}{\mathbf{m}}
\newcommand{\nfeats}{k}
\newcommand{\nprojfeats}{k_{\mathrm{proj}}}
\newcommand{\scale}{s}

\newcommand{\disp}{\psi}

\newcommand{\patchfeat}{f}
\newcommand{\pcafeat}{z}
\newcommand{\pcavolfeat}{Z}
\newcommand{\numencpatch}{P}
\newcommand{\numencfeats}{C}
\newcommand{\wplsweight}{\phi}

\newcommand{\wpls}{\text{WPLS}}
\newcommand{\pcatd}{\text{PCA3D}}
\newcommand{\gica}{\text{GICA}}
\newcommand{\bandslice}{\text{BandSlice}}
\newcommand{\dispHW}{\texttt{disp\_hw}}
\newcommand{\gridsp}{\texttt{grid\_sp}}

\title{VoxCor: Training-Free Volumetric Features \\for Multimodal Voxel Correspondence}

\author[1]{Guney Tombak}

\author[1]{Ertunc Erdil}

\author[1,2]{Ender Konukoglu}

\affil[1]{Biomedical Image Computing Group, ETH Zurich, Zurich, Switzerland}

\affil[2]{The LOOP Zurich -- Medical Research Center, Zurich, Switzerland}

\date{}

\begin{document}

\maketitle
\begin{abstract}
Cross-modal 3D medical image analysis requires voxelwise representations that remain anatomically consistent across imaging contrasts, scanners, and acquisition protocols. Recent work has shown that frozen 2D Vision Transformer (ViT) foundation models can support such representations, but typical pipelines extract features along a single anatomical axis and adapt those features inside a registration solver for one image pair at a time, leaving complementary viewing directions unused and producing representations that do not transfer to new volumes. We introduce VoxCor, a training-free fit--transform method for reusable volumetric feature representations from frozen 2D ViT foundation models. During an offline fitting phase, VoxCor combines triplanar ViT inference with a compact closed-form weighted partial least squares (WPLS) projection that uses fitting-time voxel correspondences to select modality-stable anatomical directions in the triplanar feature space. At transform time, new volumes are mapped by triplanar ViT inference and linear projection alone, without fine-tuning or registration. Voxel correspondences can then be queried directly by nearest-neighbor search. We evaluate VoxCor on intra-subject Abdomen MR--CT and inter-subject HCP T2w--T1w tasks using deformable registration, voxelwise $k$-nearest-neighbor segmentation, and segmentation-center landmark localization. VoxCor improves the hardest cross-subject, cross-modality transfer settings, reduces encoder sensitivity for dense correspondence transfer, and yields registration performance competitive with handcrafted descriptors and learned 3D features. This positions VoxCor as a reusable feature layer for downstream multimodal analysis beyond pairwise registration. Code, configuration files, and implementation details are publicly available on GitHub at \href{https://github.com/guneytombak/VoxCor}{guneytombak/VoxCor}.
\end{abstract}


\section{Introduction}
\label{sec:intro}

A central challenge in multimodal 3D medical image analysis --- spanning deformable registration, atlas-based segmentation, and population-level studies --- is to identify corresponding anatomical locations across images acquired from different subjects, modalities, scanners, and protocols~\cite{maintz1998survey,sotiras2013deformable,cabezas2011review}. Ideally, a voxelwise representation should make the same anatomical location recognizable across patients: selecting a point in one subject, such as a specific cardiac region, should retrieve or highlight the corresponding region in other subjects despite differences in anatomy, image contrast, acquisition protocol, or scanner. Building such modality- and subject-stable feature spaces remains a persistent bottleneck, since the same anatomical location can differ in appearance across contrasts and individuals.

One established way to obtain voxel correspondences is through image registration, which estimates a spatial transformation aligning anatomical locations between images. Classical methods such as SyN~\cite{avants2008symmetric}, Elastix~\cite{klein2009elastix}, and Demons~\cite{vercauteren2009diffeomorphic} have been widely used for this purpose. In multimodal settings, they address appearance differences indirectly through similarity measures such as mutual information or local structural descriptors~\cite{sotiras2013deformable,ashburner2007fast}. However, these formulations are primarily designed to solve a pairwise optimization problem: for each new image pair, correspondences must be re-estimated rather than retrieved from a shared voxelwise representation. As a result, the estimated correspondences remain pair-specific rather than forming a reusable voxelwise representation, while the hand-designed similarity measures used to guide the optimization can remain sensitive to changes in image appearance or anatomy.

Learning-based registration methods attempt to make this process faster and more robust by learning deformation predictors or task-specific similarity spaces from data~\cite{balakrishnan2019voxelmorph,vos2017end,song2022cross}. Although the field has expanded rapidly~\cite{chen2025}, these approaches typically remain tied to the modality combinations, anatomical regions, and population variability represented in their training data. When the target domain differs from this training distribution, the learned correspondences may degrade, especially if the internal feature space does not preserve anatomical neighborhoods across subjects and modalities.

Vision foundation models trained on large-scale natural image datasets have shown remarkable generalization across diverse image appearances, suggesting that their learned representations may be useful beyond the domains on which they were trained. In particular, 2D Vision Transformers (ViTs)~\cite{dosovitskiy2021an} pretrained with self-supervised objectives, such as DINO~\cite{caron2021emerging,oquab2023dinov2,simeoni2025dinov3}, or segmentation-oriented objectives, such as SAM~\cite{kirillov2023segment,ravi2024sam,carion2025sam} and its medical variants~\cite{ma2024medsam,ma2025medsam2}, have shown rich semantic structure and promising transfer to medical imaging~\cite{baharoon2023towards,cekmeceli2024vision}. Building on this idea, DINO-Reg extracts voxelwise representations from frozen DINOv2 by processing volumetric images slice-by-slice along a single anatomical axis, and combines joint-modality PCA with ConvexAdam to perform multimodal registration without encoder fine-tuning~\cite{song2024dinoreg}. More recently, Anatomix~\cite{dey2025anatomix} takes a different route by explicitly training a 3D foundation model using synthetic images generated from TotalSegmentator~\cite{wasserthal2023totalsegmentator}-derived anatomical label maps, obtaining volumetric representations for voxel correspondence rather than relying on 2D encoders applied slice-by-slice.

Despite recent progress, current ViT-based feature pipelines for medical imaging remain limited in three ways. First, they typically extract features along a single anatomical viewing direction, leaving complementary sagittal, coronal, and axial information unused~\cite{song2024dinoreg,gu2025vision}. Although 3D ViT-based foundation models for medical imaging are emerging~\cite{dey2025anatomix,he2025vista3d,du2024segvol}, many widely used large-scale ViT encoders remain 2D image models; this motivates approaches that adapt frozen 2D encoders to volumetric correspondence without training a new 3D backbone.
Second, existing feature mappings are often constructed for registering a specific image pair, so they may not generalize as standalone transformations for new data. In contrast, a reusable representation should be able to map even a single new volume into a shared, modality-stable feature space without requiring a paired image at transform time.
Third, frozen ViT features can remain sensitive when subject identity and imaging modality change simultaneously. In this regime, correspondence-agnostic compression such as joint PCA may not be sufficient to produce modality-stable anatomical neighborhoods.

In this work, we propose \textbf{VoxCor}, a training-free method that addresses these limitations. It extracts volumetric features for multimodal voxel correspondence from frozen 2D vision foundation models. Here, \emph{training-free} means that all neural backbones remain frozen: VoxCor performs no gradient-based fine-tuning and trains no task-specific deformation or segmentation network. Adaptation is limited to closed-form statistical projections fitted offline for one or more modality pairs.

VoxCor formulates feature extraction as a fit--transform process. Triplanar ViT features are adapted by a closed-form weighted partial least squares (\wpls{}) projection. This projection can be fitted once on representative paired data in voxelwise correspondence, either provided directly or derived from a fixed-parameter MIND-based registration as weak geometric supervision. At transform time, new volumes are mapped into the fitted feature space by ViT inference and linear projection alone. Direct voxel correspondences can then be obtained by nearest-neighbor search. The projection can also be fitted to a given pair of volumes, and the resulting representation can serve as input to deformable registration of that pair. In this pair-specific setting, the initial correspondences supervise the projection, while the frozen ViT features contribute broader anatomical context; the resulting representation can therefore support deformable alignment without being limited to reproducing the initial correspondence field.

We evaluate VoxCor on intra-subject Abdomen MR--CT~\cite{hering2022learn2reg} and inter-subject HCP T2w--T1w~\cite{van2013wu}, two datasets that span complementary regimes of anatomical and modality shift. Performance is measured across three correspondence tasks: deformable registration, voxelwise $k$-nearest-neighbor segmentation as a label-transfer test of feature-space neighborhoods, and registration-free correspondence as a geometric precision test. Comparisons span four frozen ViT encoders (DINOv2, DINOv3, MedSAM2, SAM3) alongside handcrafted (MIND) and learned 3D (Anatomix) descriptors. Because registration on these datasets can be adversely affected by global inter-volume misalignment, we additionally introduce \bandslice{}, a simple global initialization method for feature-based registration. \bandslice{} accounts for translation and scaling between volumes and is used to initialize the deformable registration algorithm ConvexAdam.

\paragraph{Contributions.}
\begin{enumerate}

\item We introduce \textbf{VoxCor}, a training-free fit--transform method that combines triplanar frozen ViT features with a closed-form correspondence-aware \wpls{} projection. The projection maps ViT features into a space that supports cross-subject and cross-modal correspondence.

\item We introduce \textbf{\bandslice{}}, a six-parameter per-axis scale--translation initialization method that is used to initialize ConvexAdam and improves registration accuracy across feature representations.

\item We provide \textbf{broad empirical evidence} that adapted frozen 2D ViT features support direct multimodal voxel correspondence, evaluated by deformable registration, voxelwise kNN segmentation, and registration-free correspondence. The evaluation spans four frozen ViT encoders, handcrafted and learned 3D comparators, and two complementary multimodal datasets, with the largest gains in the most challenging transfer setting, where subject identity and modality change simultaneously.

\end{enumerate}

\section{Method}
\label{sec:method}

VoxCor is composed of two phases, \emph{fit} and \emph{transform}. In the fit phase, the method determines projection matrices that map per-voxel features extracted by frozen 2D vision foundation models to a linear subspace that is shared by two given modalities. During this phase, the method uses a dataset of paired volumes in correspondence. The transform phase uses the projection matrices determined in the fit phase to map either modality to the shared linear feature subspace, at which point paired volumes in correspondence are no longer assumed. 

The projection matrices are obtained in two stages during the fit phase. In the first stage (Section~\ref{ssec:triplanar}), for each corresponding volume pair from the two modalities, initial patch features are extracted slice-by-slice using a frozen 2D ViT along each anatomical axis. A joint PCA projection is then fitted separately for each anatomical axis using features from both modalities, and used to reduce the extracted ViT features to $\nfeats$ channels. The projected features from the different axes are mapped back to the volume grid and concatenated to form \emph{per-voxel} 3D feature vectors. In the second stage (Section~\ref{ssec:wpls}), modality-specific projection matrices are fitted using weighted partial least squares (\wpls{}) and fitting-time voxel correspondences. These matrices map the per-voxel features from each modality into a shared feature space in which cross-modal voxel correspondences can be compared directly. As a correspondence-agnostic comparison (Section~\ref{ssec:pcatd}), \pcatd{} replaces \wpls{} with a second PCA projection fitted to the concatenated triplanar features. Several auxiliary components (Section~\ref{ssec:auxiliary}) are used only to make this fitting procedure well defined and robust. Foreground masks restrict PCA and \wpls{} fitting to anatomically relevant regions, while fitting-time correspondences can either be assumed from paired acquisitions or generated by fixed-parameter MIND-based registration. When large global misalignment is expected, this registration is initialized by \bandslice{}, a coarse scale--translation alignment procedure.

In the following, we describe the triplanar representation, the correspondence-aware \wpls{} projection, the correspondence-agnostic \pcatd{} comparison, and the auxiliary components used during fitting.

\begin{figure*}
    \centering
    \includegraphics[width=1\linewidth]{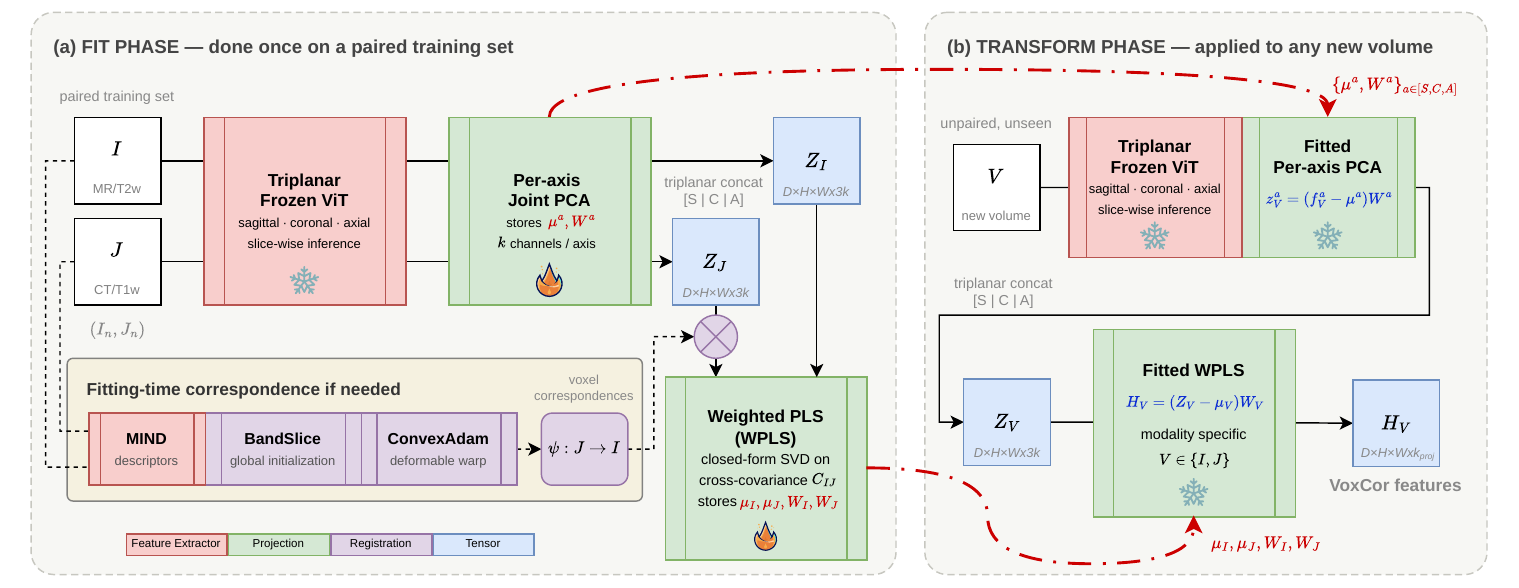}
    \caption{VoxCor pipeline, organized into a fit phase run once on a paired training set and a transform phase applied to any new volume.
    \textbf{(a) Fit phase:} paired volumes are processed by triplanar inference of a frozen 2D ViT along sagittal, coronal, and axial directions. Per-axis joint masked PCA reduces each axis to $\nfeats$ channels, and the three projected stacks are concatenated into a $3\nfeats$-channel triplanar feature volume. When voxelwise correspondences are not already available, fitting-time correspondences are generated by \bandslice{} global initialization followed by ConvexAdam deformable refinement on MIND descriptors. A correspondence-aware \wpls{} projection is then fitted by closed-form SVD of the weighted cross-covariance, yielding modality-specific projection matrices. The fitted per-axis PCA and \wpls{} projections are stored.
    \textbf{(b) Transform phase:} a new, unpaired volume is processed by the same triplanar frozen ViT, the stored per-axis PCA projections, and the stored modality-specific \wpls{} projection, producing the final $\nprojfeats$-channel VoxCor feature volume. No registration is performed for new volumes.
    For comparison, \pcatd{} replaces this final \wpls{} step with a correspondence-free PCA fitted on the concatenated triplanar features. All neural encoders remain frozen throughout. Auxiliary components used during fitting, including preprocessing and foreground masking, are described in Appendix~\ref{sec:app:method_details}.}
    \label{fig:method_pipeline}
\end{figure*}

\subsection{Triplanar ViT Feature Extraction}
\label{ssec:triplanar}
The main goal of triplanar feature extraction is to obtain 3D per-voxel features for each modality using a frozen 2D ViT.
As 2D ViTs process slices independently and do not natively encode volumetric context, we extract features along sagittal, coronal, and axial directions, and combine them per voxel. This gives each voxel complementary anatomical evidence from three orthogonal views. 

Let $(I_n, J_n) \in \mathbb{R}^{D \times H \times W},\ n = 1,\ldots,N$ be a set of $N$ paired volumes from two imaging modalities. We assume that each pair is in voxelwise correspondence during fitting. These volumes are used during the fit phase to determine the projection matrices.  

Let us denote a generic volume from either modality by $V_n \in \{I_n,J_n\}$. 
For each volume $V_n$ and each anatomical axis $a \in \{S,C,A\}$, we apply a frozen 2D ViT encoder to the corresponding stack of 2D slices. 
In order to retain high-resolution information, slices are resized by a factor $\scale>1$. This factor depends on the encoder and dataset, with memory limitations being the main constraint. Furthermore, to reduce computational cost, instead of applying the ViT to every slice, we apply it to every third slice and obtain the features of in-between slices by linear interpolation along the anatomical axis. 
We follow DINO-Reg~\cite{song2024dinoreg} in using these strategies to reduce computational cost. Aggregating patch features across all encoded slices yields
\begin{equation}
\patchfeat^a_{V_n}
\in
\mathbb{R}^{\numencpatch \times \numencfeats},
\end{equation}
where $\numencpatch$ is the number of patches along axis $a$ and $\numencfeats$ is the encoder feature dimension.

\paragraph{Per-axis joint-modality PCA:}
There are two issues with the features $\patchfeat^a_{V_n}$: the dimensionality $\numencfeats$ is often too high for downstream processing, and the features are extracted independently for each modality. DINO-Reg~\cite{song2024dinoreg} uses per-axis joint-modality PCA to address these issues. We follow the same strategy here. 

For each axis $a$, we apply PCA to patch features pooled across both modalities. 
The per-pair and global feature stacks are
\begin{equation}
\patchfeat^a_n =
\begin{bmatrix}
\patchfeat^a_{I_n} \\
\patchfeat^a_{J_n}
\end{bmatrix}
\in
\mathbb{R}^{2\numencpatch \times \numencfeats},
\qquad
\patchfeat^a =
\begin{bmatrix}
\patchfeat^a_1 \\
\vdots \\
\patchfeat^a_N
\end{bmatrix}
\in
\mathbb{R}^{2N\numencpatch \times \numencfeats}.
\end{equation}
The PCA is applied to $\patchfeat^a$ to determine a mean vector and projection matrix as

\begin{align*}
\mu^a
&=
\frac{1}{2N\numencpatch}
\sum_{m=1}^{2N\numencpatch}
(\patchfeat^a)_{m,:},
\qquad
\mu^a \in \mathbb{R}^{\numencfeats},
\\
C_{\patchfeat^a}
&=
\frac{1}{2N\numencpatch-1}
(\patchfeat^a-\mu^a)^\top(\patchfeat^a-\mu^a)
=
U^a\Lambda^a(U^a)^\top,
\\
W^a
&=
U^a_{:,1:\nfeats},
\qquad
W^a \in \mathbb{R}^{\numencfeats \times \nfeats}.
\end{align*}

$\mu^a$ and $W^a$ form the first projection stage matrices. 
Note that they are not modality-specific; they apply to both modalities. They are, however, axis-specific. 
For any volume $V_n$, the reduced patch features along axis $a$ are
\begin{equation}\label{eqn:singleAxisJointPCA}
\pcafeat^a_{V_n}
=
\bigl(\patchfeat^a_{V_n}-\mu^a\bigr)W^a
\in
\mathbb{R}^{\numencpatch \times \nfeats}.
\end{equation}
If a foreground mask is available, PCA is fitted only to foreground patches. 
Such masks can be obtained automatically; the procedure is described in Appendix~\ref{sec:app:preprocess_mask}.
At transform time, the stored mean and projection matrix are applied densely to all patch features from a test volume $V$, i.e.,
$\pcafeat^a_{V} = (\patchfeat^a_V - \mu^a)W^a$.

\paragraph{Voxel-grid reconstruction and triplanar concatenation:}
In order to aggregate features extracted along different anatomical axes, we map them back to the voxel grid.
The reduced patch-token features are placed back at their patch positions and broadcast over the patch region.
This operation, denoted by $\mathrm{unpatchify}$, gives one dense voxelwise feature volume per axis:
\begin{equation}
\pcavolfeat^a_{V_n}
=
\mathrm{unpatchify}\bigl(\pcafeat^a_{V_n}\bigr)
\in
\mathbb{R}^{D \times H \times W \times \nfeats}.
\end{equation}
The three axis-wise feature volumes are concatenated channel-wise:
\begin{equation}
\pcavolfeat_{V_n}
=
\bigl[
\pcavolfeat^S_{V_n},
\pcavolfeat^C_{V_n},
\pcavolfeat^A_{V_n}
\bigr]
\in
\mathbb{R}^{D \times H \times W \times 3\nfeats}.
\end{equation}
The triplanar representation $\pcavolfeat_{V_n}$ is the feature front-end used in this work. In principle, any encoder that directly provides voxelwise features on the volume grid, including a 3D model, could replace it, while the correspondence-aware projection described next would remain unchanged. In this work, our focus is on 2D models, specifically on using foundation models trained on very large-scale data, e.g., DINOv3. 
\subsection{Correspondence-Aware \wpls{}}
\label{ssec:wpls}

The second-stage projection matrices of VoxCor are determined using a weighted version of partial least squares (PLS)~\cite{wegelin2000survey}, which we refer to as \wpls{}. 
The main principle of PLS is to determine two projection matrices, one for each modality. 
These matrices map voxelwise features from both modalities into a common space where per-dimension correlations are maximized. 
To compute these correlations, voxelwise correspondence between $\pcavolfeat_{I_n}$ and $\pcavolfeat_{J_n}$ is assumed, i.e., the feature vector at a given voxel in $\pcavolfeat_{I_n}$ corresponds to the feature vector at the same voxel in $\pcavolfeat_{J_n}$. 
In addition, \wpls{} increases the contribution of high-gradient regions in $\pcavolfeat_{I_n}$ when estimating the projection matrices.

In principle, \wpls{} could be applied by pooling features from all voxels within the fit volumes. However, memory limitations prevent us from using all voxels. To facilitate computations, we apply average pooling to reduce the spatial dimensions of $\pcavolfeat_{V_n}$ from $D\times H\times W$ to $d\times h\times w$, obtaining $\pcavolfeat'_{V_n}\in\mathbb{R}^{d\times h\times w\times 3\nfeats}$. 
Features at different voxels in this coarser grid are then stacked to yield

\begin{align}
\pcafeat'_{I_n}
&=
\mathrm{stack}\bigl(\pcavolfeat'_{I_n}\bigr)
\in \mathbb{R}^{R \times 3\nfeats},
\\
\pcafeat'_{J_n}
&=
\mathrm{stack}\bigl(\pcavolfeat'_{J_n}\bigr)
\in \mathbb{R}^{R \times 3\nfeats},
\end{align}
where $R=dhw$ is the number of pooled feature vectors for $I_n$ and $J_n$, and ``$\mathrm{stack}$'' denotes stacking over spatial locations.
We keep the $I_n$ and $J_n$ notation here because the modality-specific roles are important for \wpls{}.

\paragraph{Modality-specific centering.}
Unlike PCA, \wpls{} uses separate means for the two modalities. We therefore compute the mean features for both modalities independently:

\begin{align*}
\mu_I
&=
\frac{1}{NR}
\sum_{n=1}^{N}
\sum_{r=1}^{R}
(\pcafeat'_{I_n})_{r,:},
\\
\mu_J
&=
\frac{1}{NR}
\sum_{n=1}^{N}
\sum_{r=1}^{R}
(\pcafeat'_{J_n})_{r,:}.
\end{align*}
These mean feature vectors are then used to center the features as 
$\bar{\pcafeat}'_{I_n} = \pcafeat'_{I_n} - \mu_I$ and $\bar{\pcafeat}'_{J_n} = \pcafeat'_{J_n} - \mu_J$.

\paragraph{Voxel weighting.}
During \wpls{} fitting, centered features are pooled from the coarse grid. Homogeneous regions may therefore dominate anatomical boundary regions in the pooled feature set. In order to emphasize the contribution of boundary locations, we implement a weighting mechanism. 
Pooled voxels are weighted by the local multichannel gradient magnitude of the features. These gradients indicate transitions in feature space, which often correspond to anatomical boundaries. 
Because the two images are assumed to be in correspondence after fitting-time alignment, we compute the weights on the fixed-modality feature grid, $\pcavolfeat'_{I_n}$, rather than on the warped moving-modality features $\pcavolfeat'_{J_n}$.
The weight is given by 

\begin{equation}
\wplsweight_{I_n,r}
=
\sqrt{
\sum_{c=1}^{3\nfeats}
\left\|
\nabla \pcavolfeat'_{I_n,r,c}
\right\|_2^{\,2}
},
\end{equation}
where $\nabla \pcavolfeat'_{I_n,r,c}$ denotes the spatial gradient of the $c$-th channel of $\pcavolfeat'_{I_n}$ at the $r$-th voxel in the coarse voxel grid. 

\paragraph{Weighted cross-covariance.} 
\wpls{} analysis relies on a weighted cross-covariance matrix computed across all pooled voxels from all fit samples. 
To this end, for each modality, we stack voxel features from all volumes row-wise to obtain $\bar{\pcafeat}'_I, \bar{\pcafeat}'_J\in \mathbb{R}^{NR\times 3\nfeats}$.
During this stacking, we respect the voxelwise correspondence between $I_n$ and $J_n$. Therefore, the corresponding rows in $\bar{\pcafeat}'_I$ and $\bar{\pcafeat}'_J$ contain corresponding feature vectors. \wpls{} relies on this correspondence to extract the projection matrices. 
In a similar fashion, we stack the weights across all volumes to obtain $\wplsweight_I \in \mathbb{R}^{NR}$.
We then estimate the weighted cross-covariance and the per-channel weighted variances of each modality:
\begin{equation}
\label{eq:wpls_xcov}
C_{IJ}
=
\frac{
\bigl( \wplsweight_I \odot \bar{\pcafeat}'_I \bigr)^{\top}
\bar{\pcafeat}'_J
}{
\sum_{r}\wplsweight_{I,r}
}
\in
\mathbb{R}^{3\nfeats \times 3\nfeats}.
\end{equation}

\begin{equation}
\sigma^2_{I,c} = \frac{\sum_{r} \wplsweight_{I,r}\,\bar{\pcafeat}'^{\,2}_{I,r,c}}{\sum_{r}\wplsweight_{I,r}},
\qquad
\sigma^2_{J,c} = \frac{\sum_{r} \wplsweight_{I,r}\,\bar{\pcafeat}'^{\,2}_{J,r,c}}{\sum_{r}\wplsweight_{I,r}},
\end{equation}
for each channel $c \in \{1,\ldots,3\nfeats\}$. The same weights $\wplsweight_I$ are used to estimate the variances of both modalities. 

\paragraph{Determination of the projection matrices by SVD.}
To prevent feature channels with large magnitude from dominating the analysis, we scale each channel to unit variance before applying the SVD.  
Let $D_I = \mathrm{diag}(\sigma_I) + \epsilon \mathbf{I}$ and $D_J = \mathrm{diag}(\sigma_J) + \epsilon\mathbf{I}$, where $\epsilon>0$ is a small ridge constant ensuring numerical stability in the inversion and $\mathbf{I}$ is an identity matrix of appropriate size. We compute the thin singular value decomposition of the scaled cross-covariance
\begin{equation}
\widetilde{C}_{IJ} = D_I^{-1}\,C_{IJ}\,D_J^{-1} = U\Sigma V^{\top},
\end{equation}
and retain the leading $\nprojfeats$ left and right singular vectors as $W_I = U_{:,1:\nprojfeats}$ and $W_J = V_{:,1:\nprojfeats}$. These modality-specific projection matrices, together with the modality-specific mean feature vectors, define the second stage of VoxCor. 

At inference time, for a new image from either modality, the second-stage projection is given, with a slight abuse of notation, by
\begin{equation}
\label{eq:wpls_proj}
H_I = \bigl( \pcavolfeat_I - \mu_I \bigr)W_I, 
\qquad
H_J = \bigl( \pcavolfeat_J - \mu_J \bigr)W_J,
\end{equation}
where $\pcavolfeat_{I}, \pcavolfeat_J\in\mathbb{R}^{D\times H\times W\times 3\nfeats}$. 
Multiplication by $W_I$ and $W_J$ from the right-hand side is applied along the last feature dimension; therefore, $H_I, H_J \in \mathbb{R}^{D \times H \times W \times \nprojfeats}$.

The scaling step equalizes per-channel marginal variances but does not decorrelate features within $I$ or within $J$; it therefore sits between standard PLS-SVD, which decomposes the raw cross-covariance $C_{IJ}$, and canonical correlation analysis, which whitens by the full within\hyp{}set covariances $\Sigma_I$ and $\Sigma_J$. This places VoxCor's final projection within the broader family of two-block PLS variants surveyed by~\cite{wegelin2000survey}, and aligns with the whitening view of cross-set decompositions developed by~\cite{jendoubi2019whitening}.  

\subsection{\pcatd{}: Correspondence-Agnostic PCA}
\label{ssec:pcatd}

VoxCor uses correspondences within \wpls{} to identify modality-specific projections. 
This stage reduces the $3\nfeats$-dimensional concatenated triplanar features to $\nprojfeats$ dimensions. 
An alternative is to use a second PCA layer to determine a modality-agnostic projection. 

To this end, we pool concatenated features from both modalities and all pairs, and compute a common mean feature $\mu \in \mathbb{R}^{3\nfeats}$. 
We then apply PCA to the centered features to determine one modality-agnostic projection matrix $W \in \mathbb{R}^{3\nfeats \times \nprojfeats}$.
In practice, we pool all $\pcafeat'_{I_n}$ and $\pcafeat'_{J_n}$ matrices row-wise to yield $\pcafeat'\in\mathbb{R}^{2NR\times 3\nfeats}$, and PCA is applied to this matrix. 

At inference time, for a volume from either modality, we obtain the projected features with 
\begin{equation}
G_V = \bigl( \pcavolfeat_V - \mu \bigr)W \in \mathbb{R}^{D \times H \times W \times \nprojfeats},
\end{equation}
using the same notation as in \wpls{} and $V\in\{I,J\}$.  \pcatd{} therefore replaces VoxCor's modality-specific means and projections with a single shared pair, and does not use correspondences during the fitting process. In our experiments, we compare the \pcatd{} features $G_I$ and $G_J$ with the \wpls{} features $H_I$ and $H_J$. 

\subsection{Auxiliary Components}
\label{ssec:auxiliary}
During the fit phase, \wpls{} assumes correspondence between pairs of volumes $(I_n, J_n)$. Furthermore, during PCA or \wpls{} fitting, background voxels can adversely influence the analysis. In this section, we provide details of two auxiliary components that support the implementation of VoxCor during the fit phase. During the transform phase, i.e., at inference time, these components are not used.

\paragraph{Foreground masking.}
In both PCA and \wpls{}, including background-voxel features during fitting may cause the fitted projections to focus on feature variation between foreground and background. This can reduce sensitivity to subtler variation within the foreground. We therefore restrict the fitting procedure to foreground regions. This masking is used only to determine the projection parameters; at transform time, the stored means and projection matrices are applied directly to all patch and voxel features without masking. This fitting-time masking changes how features are used to construct $\patchfeat_n^a$ and $(\pcafeat'_{I_n}, \pcafeat'_{J_n})$ pairs, how $\wplsweight_{I_n}$ is computed, and how the ``$\mathrm{unpatchify}$'' operation is applied.  

Let us assume that we are given foreground masks for both $I_n$ and $J_n$. When constructing $\patchfeat_{I_n}^a$, we use features only from the foreground patches in $I_n$, and likewise use foreground patches in $J_n$ to construct $\patchfeat^a_{J_n}$. Hence, $\patchfeat^a_n$ is formed only from foreground patch features from both images. This focuses the PCA analysis on the foreground. 
\wpls{} analysis, however, requires correspondence between the feature vectors. Therefore, when constructing $\pcafeat'_{I_n}$ and $\pcafeat'_{J_n}$, we use only voxels that are in the intersection of the foreground masks of $I_n$ and $J_n$ in the coarse grid. This ensures correspondence between the feature vectors. Accordingly, when computing $\wplsweight_{I_n,r}$, we normalize only over foreground voxels, and $\wplsweight_{I_n}$ is formed by stacking weights from foreground voxels only.
During fitting, the ``$\mathrm{unpatchify}$'' operation assigns zero features to background patches when constructing $\pcavolfeat^a_{V_n}$; this masking is not applied when transforming new volumes.

So far, we have not discussed how the foreground masks are generated. Several alternatives exist for creating such masks, including manual annotations. In this work, we use a fully automated method based on MIND descriptors. 
For each fit volume, a MIND descriptor map is computed and thresholded to identify near-constant background regions; the complement is used as a raw foreground mask. Enclosed background pockets inside the body are then added back to the foreground by 6-connected boundary-flood hole filling: a background voxel is kept as background only if it is reachable from the volume boundary through 6-connected background neighbors, and any background voxel that fails this reachability test is added to the foreground. This simple approach removes most background voxels, which is sufficient for the PCA and \wpls{} analyses to focus on more important feature variations in the foreground. The full procedure is detailed in Appendix~\ref{sec:app:preprocess_mask}. 
Thus, foreground masks affect only which features contribute to fitting the PCA and \wpls{} projections, and are not required at transform time.

\paragraph{Generating a fitting dataset with correspondence.}
\wpls{} analysis relies on a set of paired volumes that are in pairwise correspondence. In certain applications, such as T1-weighted and T2-weighted acquisitions in brain MRI, correspondence can be assumed between the images without any further processing. Even though that correspondence may not be perfect, several elements make VoxCor fitting robust to small deviations in correspondence, such as patch-level pooling and coarse-grid analysis for \wpls{}. 

However, correspondence cannot be assumed between different modalities in most applications, for example abdominal MRI and CT. Even when $I$ and $J$ come from the same individual, voxels may not be in correspondence. In such applications, we generate fitting-time correspondences by aligning $I_n$ and $J_n$ through non-linear registration. In this work, we use MIND features~\cite{heinrich2013ssc} with ConvexAdam~\cite{siebert2024convexadam} optimization and fixed parameters as the non-linear registration algorithm. 

In our experiments, ConvexAdam with MIND features alone was not able to account for gross global misalignments between volumes, as shown in Section~\ref{sec:experiments}. To fix this, we propose a simple algorithm to account for global misalignments between two volumes for feature-based registration methods, which we refer to as \bandslice{}. To generate fitting-time pairwise correspondences, we first use \bandslice{} with MIND features to account for global misalignment, and then apply ConvexAdam with the same features to obtain the final transformation for each $(I_n,J_n)$ pair. We refer to this combination of \bandslice{} global initialization followed by ConvexAdam refinement as Globally-Initialized ConvexAdam (\gica{}).

\paragraph{Global initialization with \bandslice{}.} 
\bandslice{} is a coarse, six-parameter scale\hyp{}translation initialization.  It estimates a restricted global transform with independent scale and translation along each anatomical axis between two given volumes $I$ and $J$. First, the feature extractor is used to extract voxelwise features from both volumes, e.g., using MIND or the triplanar features $\pcavolfeat_I$ and $\pcavolfeat_J$ described in Section~\ref{ssec:triplanar}.  For a given anatomical axis $a\in\{S,C,A\}$, slice-wise features are computed for both volumes by stacking all pixel/patch features within each slice along axis $a$. Let us denote these stacked feature vectors as $y^a_{I,i}\in\mathbb{R}^{P_I C}$ and $y^a_{J,j}\in \mathbb{R}^{P_J C}$, where $i$ and $j$ denote the slice indices along axis $a$, $P_I$ and $P_J$ denote the number of feature vectors per slice for $I$ and $J$, respectively, and $C$ is the feature dimension. 

Given the slice-wise feature vectors, the underlying principle of \bandslice{} is to determine an affine mapping between the slices in the following form 
\[ j = \sigma_a i + \delta_a, \]
where $\sigma_a$ and $\delta_a$ are axis-specific scaling and translation parameters. To determine these parameters, we compute a normalized slice-similarity matrix whose $(i,j)$ entry is given by

\begin{align}
(\bar{S}^a)_{i,j}
&=
\frac{
(y^a_{I,i})^\top y^a_{J,j}
}{
\|y^a_{I,i}\|_2\|y^a_{J,j}\|_2
},
\\
(S^a)_{i,j}
&=
\frac{1}{2}
\left(
\frac{(\bar{S}^a)_{i,j}}{\sum_{j'}(\bar{S}^a)_{i,j'}}
+
\frac{(\bar{S}^a)_{i,j}}{\sum_{i'}(\bar{S}^a)_{i',j}}
\right),
\end{align}
where the first equation simply computes the cosine similarity between the slice-wise features and the second equation applies a symmetric row and column normalization. 
\bandslice{} then determines the parameters $\sigma_a$ and $\delta_a$ by searching for the oblique line in $S^a$ that has the highest average normalized similarity. Specifically, it maximizes
\begin{equation}
  \Gamma(\sigma_a, \delta_a) =  \frac{1}{|\Omega^a(\sigma_a, \delta_a)|}\sum_{i\in\Omega^a(\sigma_a,\delta_a)} (S^a)_{i,\lfloor{\sigma_ai + \delta_a+0.5}\rfloor},
\end{equation}
where $\Omega^a(\sigma_a,\delta_a)$ defines the range of $i$ for which $\lfloor{\sigma_ai + \delta_a+0.5}\rfloor$ remains within the column range of $S^a$, and $\lfloor{\sigma_ai + \delta_a+0.5}\rfloor$ simply represents the rounding operation to determine the corresponding column index. Maximizing the similarity alone may lead to trivial solutions, such as short oblique lines that cover only a few noisy entries. To avoid such cases, we additionally require that the candidate line covers at least $\rho D_a$ slices, where $D_a \in \{D,H,W \}$ is the length along axis $a$ and $\rho=0.5$ in our experiments; lines below this overlap are excluded from the search. We also restrict the scale to $\sigma_a\in [0.8, 1.25]$ and add a regularization term that discourages $\sigma_a$ from deviating from 1, which is a realistic assumption if voxel sizes of $I$ and $J$ are made equal. 
Therefore, on each axis, \bandslice{} optimizes the regularized loss $\Gamma_R(\sigma_a, \delta_a)$ given by
\begin{equation}
    \Gamma_R(\sigma_a,\delta_a) = (1-\eta)\Gamma(\sigma_a, \delta_a) + \eta \left(1 -\frac{|\log(\sigma_a)|}{\log (1.25)}\right),
\end{equation}
where $\eta\in[0,1)$ is a weighting parameter, and the regularization term is normalized such that it takes values between 0 and 1.
In our experiments, when $I$ and $J$ come from the same individual, setting $\eta$ very close to 1, i.e., regularizing the optimization strongly, works well. When $I$ and $J$ come from different individuals, a lower $\eta$ works better. For all experiments, we use $\eta=0.99$ when $I$ and $J$ come from the same individual, effectively restricting \bandslice{} to translation-dominated initialization, and $\eta=0.1$ otherwise to allow stronger scale adaptation.

The \bandslice{} method iteratively optimizes $\Gamma_R$ along different anatomical axes, rotating through the axes. In our experiments, we start with the axial axis before proceeding through the other anatomical axes. We repeat the optimization three times going through all the axes in the same order.  

\bandslice{} can be used for any feature-based alignment method. In our experiments, we apply it to different feature representations and show that it improves MIND-based alignment as well as other feature-based alignments. 

\section{Experimental Setup}
\label{sec:experiments}

We evaluated VoxCor with two questions in mind: to what extent VoxCor's feature space supports anatomical and voxelwise correspondence, and whether that correspondence holds when subject identity and imaging modality change. To probe both questions, we adopted a three-task evaluation protocol: deformable image registration, voxelwise $k$-nearest-neighbor (kNN) segmentation, and registration-free correspondence. Registration tests whether the feature space captures enough location information to drive optimization-based alignment; kNN segmentation tests whether features carry contextual information sufficient for direct nearest-neighbor matches in the feature space to transfer anatomical labels, i.e., transduction; registration-free correspondence tests whether features are descriptive enough to allow precise voxelwise matching across images and modalities without registration.
We evaluated VoxCor in two scenarios: multimodal abdominal imaging with MRI and CT, and multi-contrast brain MRI with T1-weighted and T2-weighted images. The remainder of this section defines the tasks (Section~\ref{ssec:tasks}), the fitting regimes used to obtain VoxCor's projections (Section~\ref{ssec:fit_regimes}), the datasets (Section~\ref{ssec:datasets}), the baselines against which we compare VoxCor (Section~\ref{ssec:encoders_baselines}), and the evaluation details (Section~\ref{ssec:eval_details}).

\subsection{Tasks}
\label{ssec:tasks}

\begin{figure}[h!]
    \centering
    \includegraphics[width=0.7\linewidth]{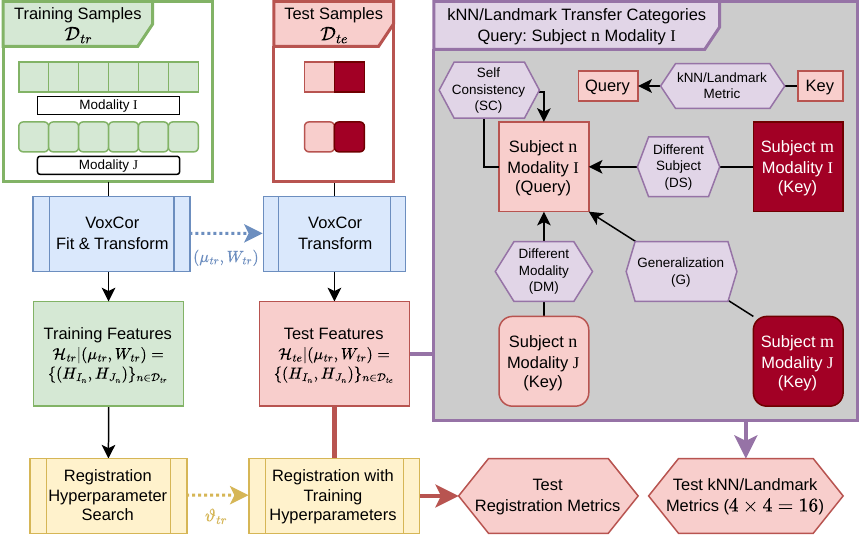}
    \caption{Evaluation protocol (shown for Abdomen MR--CT with L2OCV). \textbf{Top:} training samples are used for Dataset-Fit feature fitting and registration hyperparameter search, and the fitted projections are applied to held-out test samples. \textbf{Bottom:} registration is evaluated on held-out test pairs using either Dataset-Fit projections fitted on the training split or Pair-Fit projections fitted separately on each evaluated pair; registration hyperparameters are selected on the training split. \textbf{Right:} kNN segmentation and segmentation-center landmark localization are evaluated under the Dataset-Fit setting on all $4 \times 4 = 16$ query--key combinations within each evaluation block of two subjects ($n$, $m$) and two modalities ($I$, $J$; four volumes total), grouped into Self-Consistency (SC), Different Subject (DS), Different Modality (DM), and Generalization (G). For HCP T2w--T1w, the same protocol is used with a fixed train/test split instead of cross-validation, with subject-matched pairs for Dataset-Fit fitting and permuted inter-subject pairs for registration.}
    \label{fig:eval_protocol}
\end{figure}

The three tasks evaluate complementary properties of the same feature space.

\paragraph{Deformable registration:}
We registered two images using feature-based deformable registration and assessed feature quality through registration performance. We used ConvexAdam~\cite{siebert2024convexadam} as the deformable backend, either directly or after global initialization using \bandslice{}. We denote the direct application as \textbf{CA} and its application after global initialization as \textbf{\gica{}}. For all ViT-based methods and their variants, we applied $L_2$ feature normalization and multiplied the regularization parameter $\lambda$ by $0.1$ to account for the reduced feature magnitudes after normalization. \gica{} was our primary registration setting because it consistently improved registration performance across baselines. Performance was measured by mean Dice ($\uparrow$), 95th-percentile Hausdorff distance (HD95, mm, $\downarrow$), and the standard deviation of the log-Jacobian determinant (sdLogJ, $\downarrow$). ConvexAdam's hyperparameters were selected by random sampling of structural configurations combined with an exhaustive grid over refinement parameters, summarized in Section~\ref{ssec:eval_details} and detailed in Appendix~\ref{ssec:app:hps}. 

\paragraph{Voxelwise kNN segmentation:}
We performed voxelwise segmentation with kNN in the feature space. Voxelwise features are extracted from a query and key volume. For each voxel in the query volume, the closest $k$ voxels in the key volume are determined based on the distances in the feature space. The majority-voted label among the $k$ closest voxels is assigned to the query voxel and a voxelwise segmentation of the query volume is obtained in this manner. Segmentation accuracy indicates how well features capture contextual and semantic information that generalizes across volumes. For tractability, both query and key volume voxels were confined to the dataset-provided region-of-interest (ROI) masks, which cover the anatomical structures of interest plus some surrounding background. We scored only query voxels with a non-background segmentation label, while the key-side kNN search ranges over the full ROI; the method therefore had no explicit foreground prior on the key side.
For each query voxel, we retrieved its $k$ nearest neighbors among key-volume ROI voxels by cosine similarity and assigned the majority-voted label. Unless otherwise stated, we report $k = 7$; a sensitivity analysis over $k \in \{1,3,5,7,9,11\}$ is provided in Appendix~\ref{sec:app:knn_sensk}. Performance is reported in terms of foreground Dice ($\uparrow$).

\paragraph{Registration-free correspondence:}
This assessment is a stricter version of voxelwise kNN segmentation.
This test assesses whether, for a given voxel in one image, the single nearest voxel in the feature space lies geometrically close to the corresponding location in another image. For this evaluation, we identified ``landmarks''\footnote{These are derived from segmentation maps and are not landmarks in the strict anatomical sense.} that can be determined across modalities and patients, and performed the assessment using these landmarks. For each volume and modality, landmarks were taken as the centers of mass of selected segmentation labels (four annotated organs for Abdomen MR--CT; FreeSurfer labels listed in Appendix~\ref{ssec:app:landmark_annotations} for HCP T2w--T1w). For each query--key volume pair and landmark, we extracted the feature at the landmark voxel in the query volume, found the nearest voxel in feature space ($L_2$ distance) among the voxelwise features extracted from the key volume, and reported the Euclidean distance in millimeters between this nearest voxel and the corresponding landmark in the key volume; we report top-1 localization ($k=1$) unless stated otherwise.

\paragraph{Transfer categories:}
For kNN segmentation and registration-free correspondence, evaluation was structured around blocks of two subjects with two modalities each (four volumes per block). Each such block yields $4 \times 4 = 16$ query--key combinations. Each combination is classified into four \emph{transfer categories} by the relationship between query and key: Self-Consistency (\textbf{SC}, query $=$ key); Different Subject (\textbf{DS}, same modality, different subject); Different Modality (\textbf{DM}, same subject, different modality); and Generalization (\textbf{G}, both subject and modality differ). G is the most demanding category and the focus of our main analysis; SC mainly serves as a sanity check for feature locality. For the Self-Consistency category, we excluded the query location from the key set: in kNN segmentation, the query voxel itself was not allowed as a neighbor, and in registration-free correspondence, the query landmark voxel was removed from the candidate key voxels. This prevents the SC score from being determined by trivial self-matches. Both tasks used the Dataset-Fit setting only (Section~\ref{ssec:fit_regimes}). Figure~\ref{fig:eval_protocol} summarizes the full evaluation protocol.

\subsection{Fitting Regimes}
\label{ssec:fit_regimes}
VoxCor learns its per-axis and per-modality projection matrices during a fit phase using PCA and \wpls{}, respectively. The \pcatd{} baseline shares the initial per-axis PCA but then uses another PCA to determine its correspondence-free projection matrices. We evaluated two ways of obtaining these projections, which differ in the data used at fit time and in the downstream tasks to which they apply.

\paragraph{Dataset-Fit:}
In this regime, the projections are fitted on a separate training set and reused on unseen test volumes. This is the reusable deployment regime of VoxCor. We evaluated this regime on the kNN segmentation and registration-free correspondence tasks; no registration was run at transform time. At transform time, the test volumes need not be paired. Features can be extracted from a single volume of either modality using the projection matrices determined in the fit phase.

\paragraph{Pair-Fit:}
In applications where both modalities of a pair are available at test time, such as image registration, the projections can be fitted on the test pair itself, without relying on a separate training set. This is a pair-specific adaptation, where the feature subspace is adapted to the given pair of volumes. We evaluated this regime \emph{only on the registration task}, since it is the only task in our protocol where both volumes of a pair are available together. For the registration task, we also report the performance of the Dataset-Fit regime, so that the gap between the two alternatives can reveal how much pair-specific adaptation contributed beyond a reusable feature space.

\subsection{Datasets and Pairing Conventions}
\label{ssec:datasets}

We used two publicly available datasets in our experiments. We used all three evaluation tasks on both datasets; however, some experimental details vary to accommodate dataset-specific characteristics. We describe the specifics and the experimental variations below, and full dataset and label specifications are given in Appendix~\ref{ssec:app:dataset_details}.

\subsubsection{Abdomen MR--CT}

We used the eight paired MR--CT volumes from the Learn2Reg Abdomen MR--CT dataset~\cite{hering2022learn2reg}, resampled to $192 \times 160 \times 192$ at $2$\,mm isotropic spacing, with manual segmentations of liver, spleen, and left and right kidneys. Because only eight paired cases were available, we used Leave-2-Out Cross-Validation (L2OCV): in each fold, two pairs were held out for testing, and the remaining six pairs were used both for fitting projection matrices and for selecting ConvexAdam's hyperparameters.

\paragraph{Pairing:}
The dataset provides intra-subject abdominal MR--CT pairs. In each fold, the six training pairs were used for the Dataset-Fit procedure and registration hyperparameter search; the two held-out pairs were used for evaluation. For evaluations using registration as the downstream task, the projection matrices were fitted separately on each evaluated held-out MR--CT pair using the Pair-Fit procedure.

\paragraph{Fitting-time correspondence:}
As intra-subject Abdomen MR and CT volumes are not voxelwise aligned, we used the method described in Section~\ref{ssec:auxiliary} to generate initial correspondences between the MR and CT volumes for each subject. These correspondences were used in \wpls{} fitting, as described in Section~\ref{ssec:wpls}. Specifically, we used \bandslice{} with MIND features to account for global misalignments, followed by a fixed-parameter ConvexAdam that also uses MIND features to align the volumes. This procedure is further detailed in Appendix~\ref{sec:app:correspondence_generation} and the fixed parameters are listed in Appendix~\ref{ssec:app:internal_mind_gica}.

\subsubsection{HCP T2w--T1w}

We used T2-weighted (T2w) and T1-weighted (T1w) brain MR volumes from the Human Connectome Project (HCP)~\cite{van2013wu}, resampled to $256^3$ at $0.7$\,mm isotropic spacing. Each subject has one T2w and one T1w scan alongside fourteen FreeSurfer-derived anatomical labels~\cite{freesurfer}, which we used in the evaluations. We used six subjects to form the training split, which was used in the Dataset-Fit procedure and registration hyperparameter selection. We used twelve other subjects as the held-out evaluation set for the registration and kNN segmentation tasks. For the registration-free correspondence task only, the held-out set was expanded by twelve additional subjects (twenty-four in total), since this task was computationally cheaper than dense voxelwise kNN or registration.

\paragraph{Pairing:}
A subject's T2w--T1w MRIs in the brain dataset were already in coarse correspondence. This allowed two sets of experiments, which differ in how we construct the paired training set.

\begin{itemize}
    \item \textbf{Subject-matched pairs} were used for \emph{Dataset-Fit feature fitting}. Each subject's native T2w and T1w volumes form a pair. Anatomical correspondence between these images was assumed, so no registration was applied to obtain fitting correspondences for \wpls{}. This pairing was used for the kNN segmentation and registration-free correspondence downstream tasks.
    \item \textbf{Permuted inter-subject pairs} were used for evaluations with the registration downstream task. In these experiments, ``fixed'' and ``moving'' images came from different subjects (T2w from one subject paired with T1w from another), so the registration task was both inter-modality and inter-subject. In the Pair-Fit regime for these experiments, fitting-time correspondences were generated using the internal MIND+\gica{} registration.
\end{itemize}

\subsection{Encoders and Feature Representations}
\label{ssec:encoders_baselines}

We evaluated four frozen 2D ViT encoders alongside handcrafted and learned 3D baselines, and compared different ways of using the ViT features, including VoxCor.

\paragraph{Frozen ViT encoders:}
We used DINOv2~\cite{oquab2023dinov2}, DINOv3~\cite{simeoni2025dinov3}, MedSAM2~\cite{ma2025medsam2}, and SAM3~\cite{carion2025sam}. For DINOv2 and DINOv3, we used ViT-L backbones with patch sizes $14 \times 14$ and $16 \times 16$, respectively; MedSAM2 also uses patch size $16 \times 16$, while SAM3 uses its native fixed-resolution image encoder with patch size $14 \times 14$. In all cases, the encoder was kept frozen and dense patch-token features were extracted from the final encoder layer without task-specific heads. To obtain comparable patch-token densities across encoders, we followed the input-rescaling strategy of DINO-Reg~\cite{song2024dinoreg}; encoder- and dataset-specific scaling factors are reported in Appendix~\ref{ssec:app:data_variant_params}.

\paragraph{CNN and handcrafted baselines:}
We compared against MIND~\cite{heinrich2013ssc}, a 12-channel handcrafted local self-similarity descriptor, and Anatomix~\cite{dey2025anatomix}, a pretrained 3D CNN feature extractor. We used the publicly available Anatomix weights \texttt{anatomix.pth} for Abdomen MR--CT and \texttt{anatomix+brains.pth} for HCP T2w--T1w.

\paragraph{ViT feature variants:}
For each frozen ViT encoder, we compared three feature representations:
\begin{itemize}
    \item \textbf{Single-axis PCA}, following DINO-Reg~\cite{song2024dinoreg}, applied per-axis joint-modality PCA to one anatomical viewing direction only to determine a projection matrix. Effectively, this variant extracts features as described in Equation~\ref{eqn:singleAxisJointPCA}. During the PCA analysis, we used automatically generated foreground masks, as described in Section~\ref{ssec:auxiliary}. For the registration downstream task, we used the axial direction as the single-axis baseline. This is in accordance with~\cite{song2024dinoreg}. We refer to this variant as \emph{Axial} in our experiments. For the kNN segmentation and registration-free correspondence downstream tasks, we additionally report the best single-axis result, assuming an oracle that selects the best of sagittal, coronal, or axial PCA features separately for each evaluation group. We refer to this variant as \emph{Best Axis PCA} in our experiments. In all cases, we used the 24 dimensions with the highest variance found during PCA, i.e., $\nfeats=24$.
    \item \textbf{\pcatd{}} applies a correspondence-free PCA projection to the concatenated triplanar features, testing whether triplanar aggregation alone can recover modality-stable correspondences. \pcatd{} reduces 72 dimensions to 24, so that it can be fairly compared to the single-axis PCA features, i.e., $\nprojfeats = 24$. \pcatd{} forms the most natural alternative to \wpls{}. 
    \item \textbf{\wpls{}} uses the same triplanar input as \pcatd{} but fits a correspondence-aware projection as described in Section~\ref{ssec:wpls}, testing whether explicit geometric supervision improves modality-stable correspondence compared to \pcatd{}. In accordance with the other alternatives, we used 24 \wpls{} dimensions, i.e., $\nprojfeats = 24$. All results indicated with \wpls{} in the next section correspond to variants of the proposed VoxCor method. 
\end{itemize}

\paragraph{+MIND hybrids (registration only):}
For the registration downstream task, Dey et al.~\cite{dey2025anatomix} have shown that combining Anatomix with MIND features leads to clear improvements in registration performance. Following the same approach, we additionally evaluated +MIND hybrids that concatenate the first 16 dimensions of ViT features with a 12-channel MIND descriptor, giving 28 channels. These dimensions were chosen to conform with the experimental setup in~\cite{dey2025anatomix} to facilitate comparisons. The same construction was applied to Anatomix (Anatomix+MIND) and to each ViT feature variant (Axial+MIND, \pcatd{}+MIND, \wpls{}+MIND). Because ViT and MIND features have different magnitude scales, the selected ViT channels were scaled by $0.1$ before concatenation. Per-voxel feature normalization, regularization scaling, and the MIND parameters used inside +MIND hybrids are reported in further detail in Appendix~\ref{ssec:app:feature_norm}. Among the +MIND hybrids, only Anatomix+MIND was evaluated on the kNN segmentation and registration-free correspondence tasks; the ViT+MIND hybrids were evaluated only for registration.

\subsection{Evaluation Details}
\label{ssec:eval_details}

\begin{table}
\centering
\caption{Experimental configurations. Pair-Fit was used only for registration, while kNN segmentation and landmark localization were evaluated only under Dataset-Fit. For HCP T2w--T1w, Dataset-Fit fitting uses subject-matched pairs, whereas registration hyperparameter search, Pair-Fit fitting, and registration evaluation use permuted inter-subject pairs.}
\label{tab:exp_grid}
\vspace{4pt}
\begin{tabular}{p{5cm}|p{10cm}}
\hline
\textbf{Factor} & \textbf{Levels} \\
\hline
\multicolumn{2}{l}{\emph{Baseline methods}} \\
\hline
Features & MIND (12\,ch), Anatomix (16\,ch), Anatomix+MIND (28\,ch) \\
Registration & CA, \gica{} \\
\hline
\multicolumn{2}{l}{\emph{ViT methods (per encoder)}} \\
\hline
Registration features & Axial PCA, \pcatd{}, \wpls{} (24\,ch; each) \\
Registration feature setting & Base, +MIND \\
kNN / landmark features & Best Axis PCA, \pcatd{}, \wpls{} \\
Fitting & Registration: Pair-Fit and Dataset-Fit; kNN / landmark: Dataset-Fit only \\
Registration & CA, \gica{} \\
\hline
\end{tabular}
\end{table}

\paragraph{Registration hyperparameter selection:}
All deformable registration experiments used ConvexAdam~\cite{siebert2024convexadam} as the feature-based registration algorithm. ConvexAdam requires hyperparameter selection on a validation set. We sampled $N = 400$ structural configurations uniformly at random from a memory-feasible subset of the search space, and evaluated each over an exhaustive $4 \times 4$ grid of refinement parameters, yielding $6{,}400$ evaluated configurations per method setting. The search space additionally included \gica{} variants that skip the Adam refinement (convex-only) and that skip both refinement stages (global-initialization only), so that hyperparameter selection could collapse to a coarser or fully affine output when feature-driven deformation did not improve validation Dice. For Abdomen MR--CT, the best configuration was selected by L2OCV Dice within each fold; for HCP T2w--T1w, the best configuration was selected on the training split and applied to the held-out inter-subject registration tasks. The full search space, sampling protocol, ConvexAdam-MIND extension, and VRAM-aware pre-screening are further detailed in Appendix~\ref{ssec:app:hps}--\ref{ssec:app:vram}.

\paragraph{kNN and registration-free correspondence protocols:}
The kNN segmentation and registration\hyp{}free correspondence tasks have no hyperparameter selection. For kNN segmentation, $k = 7$ is fixed in the main results (with a sensitivity analysis in Appendix~\ref{sec:app:knn_sensk}), and registration-free correspondence uses fixed top-1 matching with $L_2$ feature distance.

\paragraph{Implementation:}
Experiments were implemented in PyTorch, with frozen ViT encoders run in bfloat16 and xFormers~\cite{xFormers2022} memory-efficient attention enabled where supported. Most evaluations were performed on a single NVIDIA A6000 GPU (48\,GB); a small number of large registration configurations required an A100 (80\,GB). Evaluation scripts support checkpointed execution for SLURM workflows. Full software and hardware details are given in Appendix~\ref{ssec:app:software_hardware}.

\paragraph{Summary:}
Table~\ref{tab:exp_grid} summarizes the experimental grid across encoders, feature representations, fitting regimes, and registration settings.

\section{Results}
\label{sec:results}

We report results in four stages. We first present results on the deformable registration task, which tests whether VoxCor features support optimization-based anatomical alignment as well as the established alternatives. We then present results on voxelwise kNN segmentation (label transfer by nearest-neighbor matching in the feature space) and registration-free correspondence (correspondence based on nearest-neighbor matching in the feature space). Finally, we present computational characteristics in terms of fitting time, transform-time feature extraction cost, and peak GPU memory.

\subsection{Deformable Registration Performance}
\label{ssec:results_reg}

\begin{figure}[h!]
    \centering
    \includegraphics[width=0.7\linewidth]{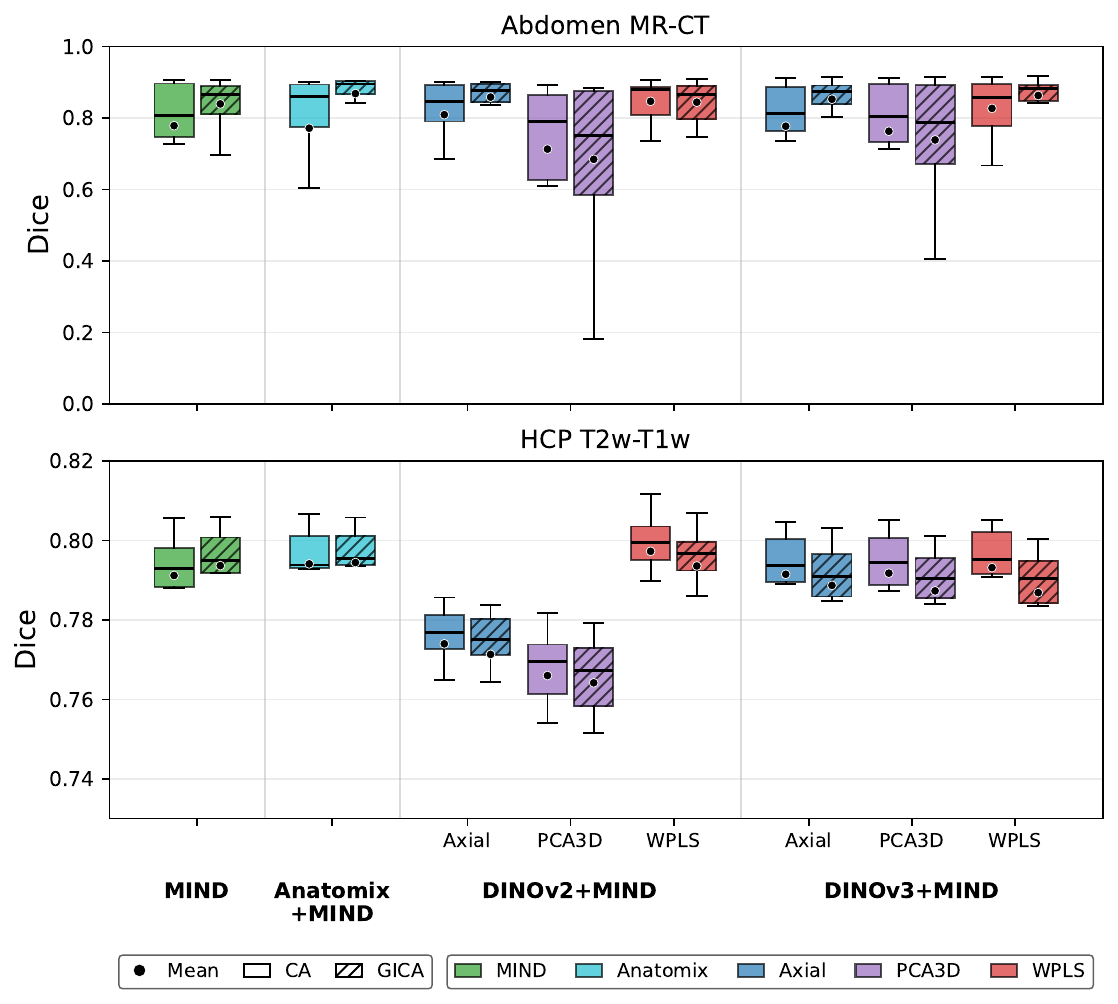}
    \caption{Direct ConvexAdam (CA, plain boxes) versus Globally-Initialized ConvexAdam (\gica{}, hatched boxes) under the MIND-augmented registration setting. Boxes show the interquartile range of Dice scores; center lines mark medians, whiskers cover the non-outlier range, and black dots mark means. CNN rows show MIND and Anatomix+MIND; ViT-based rows show Pair-Fit\,+\,MIND features. Global initialization is most beneficial on Abdomen MR--CT, where coarse field-of-view mismatch is a real factor; on HCP T2w--T1w, the two settings are nearly indistinguishable.}
    \label{fig:reg_gica_compare}
\end{figure}

\begin{figure}[h!]
    \centering
    \includegraphics[width=0.7\linewidth]{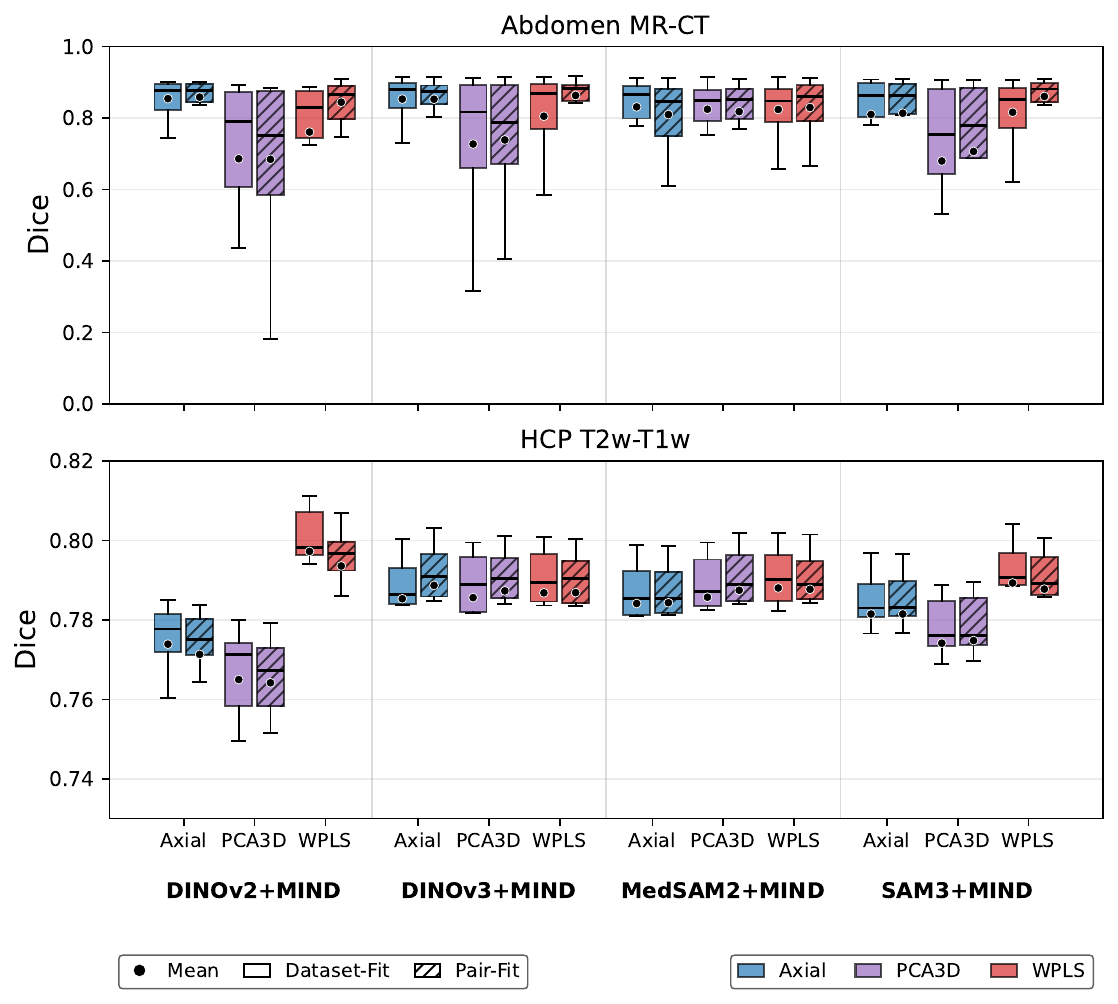}
    \caption{Reusable Dataset-Fit (plain boxes) versus pair-specific Pair-Fit (hatched boxes) under \gica{} for ViT-based +MIND configurations. Boxes show the interquartile range of Dice scores; center lines mark medians, whiskers cover the non-outlier range, and black dots mark means. The two regimes are nearly indistinguishable on HCP T2w--T1w; on Abdomen MR--CT, Pair-Fit measurably improves \wpls{}+MIND for several encoders.}
    \label{fig:reg_fit_compare}
\end{figure}

The main registration finding is that VoxCor features were competitive with the compared handcrafted descriptors and learned 3D features, while the relative behavior depended strongly on whether MIND features were concatenated. Tables~\ref{tab:reg_main_base} and~\ref{tab:reg_main_mind} report the main quantitative results,
the former containing results obtained with different feature representations and the latter when MIND features are concatenated to base features.  
For CNN-based baselines, the first table reports MIND and Anatomix, while the second table repeats MIND results and reports new results for Anatomix+MIND. For ViT-based methods, the tables report results under the Pair-Fit procedure; the reusable Dataset-Fit procedure is analyzed separately in Fig.~\ref{fig:reg_fit_compare}. The registration results in Tables~\ref{tab:reg_main_base},~\ref{tab:reg_main_mind} and Fig.~\ref{fig:reg_fit_compare} use \bandslice{} for global initialization. In Fig.~\ref{fig:reg_gica_compare}, we present a comparison of registration performance with and without \bandslice{}, i.e., CA versus \gica{}, for CNN baselines and DINO-based ViT methods. Complete numerical results are provided in Appendix~\ref{sec:app:reg_all}.

\begin{table}
\centering
\caption{Deformable registration performance for base feature representations (no MIND concatenation). CNN baselines are evaluated under \textbf{\gica{}}; ViT-based methods are evaluated under \textbf{Pair-Fit\,+\,\gica{}}. Values are mean$\pm$standard deviation, averaged across L2OCV folds for Abdomen MR--CT and across held-out test pairs for HCP T2w--T1w. Dice ($\uparrow$), HD95 [mm] ($\downarrow$), and sdLogJ ($\downarrow$). Bold marks the best mean Dice and HD95 per column; lower sdLogJ indicates smoother fields but should be interpreted jointly with Dice and HD95. Some entries report sdLogJ = 0.000; in these cases the selected hyperparameter configuration's refinement stage did not improve over the BandSlice global initialization, so the final displacement reduces to an affine transform and the log-Jacobian determinant is constant.}
\label{tab:reg_main_base}
\vspace{4pt}
\setlength{\tabcolsep}{3.2pt}
\begin{adjustbox}{width=\textwidth}
\begin{tabular}{l|l|ccc||ccc}
\hline
\multirow{2}{*}{\textbf{Encoder}} & \multirow{2}{*}{\textbf{Method}} & \multicolumn{3}{c||}{\textbf{Abdomen MR--CT}} & \multicolumn{3}{c}{\textbf{HCP T2w--T1w}} \\
 & & Dice $\uparrow$ & HD95 [mm] $\downarrow$ & sdLogJ $\downarrow$ & Dice $\uparrow$ & HD95 [mm] $\downarrow$ & sdLogJ $\downarrow$ \\
\hline\hline
-- & Initial & $0.373{\pm}0.172$ & $40.347{\pm}21.574$ & -- & $0.548{\pm}0.073$ & $\phantom{0}4.864{\pm}\phantom{0}1.056$ & -- \\
\hline
-- & MIND \cite{heinrich2013ssc} & $0.839{\pm}0.073$ & $13.874{\pm}10.840$ & $0.163{\pm}0.030$ & $0.794{\pm}0.011$ & $\phantom{0}1.934{\pm}\phantom{0}0.202$ & $0.067{\pm}0.007$ \\
\multirow{1}{*}{--} & Anatomix \cite{dey2025anatomix} & $0.803{\pm}0.119$ & $14.036{\pm}\phantom{0}8.377$ & $0.119{\pm}0.015$ & $0.736{\pm}0.018$ & $\phantom{0}2.345{\pm}\phantom{0}0.271$ & $0.046{\pm}0.008$ \\
\hline
\multirow{3}{*}{DINOv2} & Axial \cite{song2024dinoreg} & $\mathbf{0.841{\pm}0.058}$ & $11.471{\pm}\phantom{0}6.727$ & $0.154{\pm}0.025$ & $0.742{\pm}0.014$ & $\phantom{0}2.278{\pm}\phantom{0}0.238$ & $0.058{\pm}0.007$ \\
 & \pcatd{} & $0.703{\pm}0.230$ & $22.712{\pm}19.323$ & $0.164{\pm}0.027$ & $0.749{\pm}0.014$ & $\phantom{0}2.188{\pm}\phantom{0}0.255$ & $0.055{\pm}0.008$ \\
 & \wpls{} & $0.839{\pm}0.060$ & $11.454{\pm}\phantom{0}7.529$ & $0.150{\pm}0.020$ & $0.784{\pm}0.014$ & $\phantom{0}1.970{\pm}\phantom{0}0.222$ & $0.061{\pm}0.007$ \\
\hline
\multirow{3}{*}{DINOv3} & Axial & $0.781{\pm}0.128$ & $16.798{\pm}11.572$ & $0.150{\pm}0.036$ & $0.722{\pm}0.016$ & $\phantom{0}2.477{\pm}\phantom{0}0.267$ & $0.051{\pm}0.006$ \\
 & \pcatd{} & $0.662{\pm}0.248$ & $26.160{\pm}24.899$ & $0.127{\pm}0.031$ & $0.731{\pm}0.025$ & $\phantom{0}2.391{\pm}\phantom{0}0.322$ & $0.049{\pm}0.007$ \\
 & \wpls{} & $0.826{\pm}0.073$ & $12.091{\pm}\phantom{0}5.477$ & $0.138{\pm}0.014$ & $0.772{\pm}0.015$ & $\phantom{0}2.069{\pm}\phantom{0}0.226$ & $0.050{\pm}0.007$ \\
\hline
\multirow{3}{*}{MedSAM2} & Axial & $0.835{\pm}0.097$ & $12.401{\pm}\phantom{0}6.954$ & $0.149{\pm}0.033$ & $0.645{\pm}0.031$ & $\phantom{0}3.302{\pm}\phantom{0}0.542$ & $0.077{\pm}0.010$ \\
 & \pcatd{} & $0.822{\pm}0.109$ & $14.878{\pm}\phantom{0}9.038$ & $0.206{\pm}0.078$ & $0.636{\pm}0.040$ & $\phantom{0}3.560{\pm}\phantom{0}0.630$ & $0.000{\pm}0.000$ \\
 & \wpls{} & $0.824{\pm}0.132$ & $\mathbf{10.967{\pm}\phantom{0}7.465}$ & $0.142{\pm}0.027$ & $\mathbf{0.797{\pm}0.011}$ & $\mathbf{\phantom{0}1.912{\pm}\phantom{0}0.191}$ & $0.094{\pm}0.011$ \\
\hline
\multirow{3}{*}{SAM3} & Axial & $0.778{\pm}0.110$ & $17.223{\pm}\phantom{0}7.605$ & $0.141{\pm}0.012$ & $0.657{\pm}0.031$ & $\phantom{0}3.284{\pm}\phantom{0}0.478$ & $0.056{\pm}0.006$ \\
 & \pcatd{} & $0.610{\pm}0.255$ & $30.934{\pm}22.572$ & $0.163{\pm}0.029$ & $0.617{\pm}0.041$ & $\phantom{0}3.573{\pm}\phantom{0}0.532$ & $0.046{\pm}0.008$ \\
 & \wpls{} & $0.794{\pm}0.132$ & $13.212{\pm}\phantom{0}6.245$ & $0.145{\pm}0.023$ & $0.761{\pm}0.021$ & $\phantom{0}2.244{\pm}\phantom{0}0.246$ & $0.077{\pm}0.011$ \\
\hline
\end{tabular}
\end{adjustbox}
\end{table}

\begin{table}
\centering
\caption{Deformable registration performance for MIND-augmented feature representations. CNN baselines are evaluated under \textbf{\gica{}}; ViT-based methods are evaluated under \textbf{Pair-Fit\,+\,\gica{}}. Format and metrics as in Table~\ref{tab:reg_main_base}.}
\label{tab:reg_main_mind}
\vspace{4pt}
\setlength{\tabcolsep}{3.2pt}
\begin{adjustbox}{width=\textwidth}
\begin{tabular}{l|l|ccc||ccc}
\hline
\multirow{2}{*}{\textbf{Encoder}} & \multirow{2}{*}{\textbf{Method}} & \multicolumn{3}{c||}{\textbf{Abdomen MR--CT}} & \multicolumn{3}{c}{\textbf{HCP T2w--T1w}} \\
 & & Dice $\uparrow$ & HD95 [mm] $\downarrow$ & sdLogJ $\downarrow$ & Dice $\uparrow$ & HD95 [mm] $\downarrow$ & sdLogJ $\downarrow$ \\
\hline\hline
-- & Initial & $0.373{\pm}0.172$ & $40.347{\pm}21.574$ & -- & $0.548{\pm}0.073$ & $\phantom{0}4.864{\pm}\phantom{0}1.056$ & -- \\
\hline
-- & MIND \cite{heinrich2013ssc} & $0.839{\pm}0.073$ & $13.874{\pm}10.840$ & $0.163{\pm}0.030$ & $0.794{\pm}0.011$ & $\phantom{0}1.934{\pm}\phantom{0}0.202$ & $0.067{\pm}0.007$ \\
\multirow{1}{*}{--} & Anatomix+MIND \cite{dey2025anatomix} & $\mathbf{0.868{\pm}0.059}$ & $\phantom{0}9.556{\pm}\phantom{0}5.655$ & $0.155{\pm}0.018$ & $\mathbf{0.794{\pm}0.011}$ & $\phantom{0}1.933{\pm}\phantom{0}0.219$ & $0.071{\pm}0.009$ \\
\hline
\multirow{3}{*}{DINOv2} & Axial+MIND & $0.858{\pm}0.055$ & $\phantom{0}9.965{\pm}\phantom{0}5.373$ & $0.134{\pm}0.017$ & $0.771{\pm}0.013$ & $\phantom{0}2.101{\pm}\phantom{0}0.236$ & $0.048{\pm}0.007$ \\
 & \pcatd{}+MIND & $0.684{\pm}0.242$ & $25.904{\pm}21.856$ & $0.146{\pm}0.023$ & $0.764{\pm}0.014$ & $\phantom{0}2.102{\pm}\phantom{0}0.245$ & $0.052{\pm}0.008$ \\
 & \wpls{}+MIND & $0.844{\pm}0.061$ & $11.004{\pm}\phantom{0}6.126$ & $0.139{\pm}0.016$ & $0.794{\pm}0.012$ & $\mathbf{\phantom{0}1.928{\pm}\phantom{0}0.212}$ & $0.062{\pm}0.008$ \\
\hline
\multirow{3}{*}{DINOv3} & Axial+MIND & $0.853{\pm}0.066$ & $10.146{\pm}\phantom{0}5.721$ & $0.100{\pm}0.020$ & $0.789{\pm}0.012$ & $\phantom{0}1.955{\pm}\phantom{0}0.226$ & $0.047{\pm}0.005$ \\
 & \pcatd{}+MIND & $0.739{\pm}0.192$ & $20.115{\pm}18.391$ & $0.106{\pm}0.032$ & $0.787{\pm}0.013$ & $\phantom{0}1.967{\pm}\phantom{0}0.238$ & $0.044{\pm}0.005$ \\
 & \wpls{}+MIND & $0.863{\pm}0.056$ & $\phantom{0}9.463{\pm}\phantom{0}4.739$ & $0.099{\pm}0.022$ & $0.787{\pm}0.013$ & $\phantom{0}1.962{\pm}\phantom{0}0.225$ & $0.038{\pm}0.004$ \\
\hline
\multirow{3}{*}{MedSAM2} & Axial+MIND & $0.810{\pm}0.102$ & $13.559{\pm}\phantom{0}6.649$ & $0.075{\pm}0.019$ & $0.784{\pm}0.012$ & $\phantom{0}1.996{\pm}\phantom{0}0.235$ & $0.039{\pm}0.004$ \\
 & \pcatd{}+MIND & $0.818{\pm}0.108$ & $12.799{\pm}\phantom{0}6.802$ & $0.077{\pm}0.025$ & $0.787{\pm}0.013$ & $\phantom{0}1.980{\pm}\phantom{0}0.235$ & $0.045{\pm}0.004$ \\
 & \wpls{}+MIND & $0.829{\pm}0.085$ & $12.413{\pm}\phantom{0}6.297$ & $0.090{\pm}0.017$ & $0.788{\pm}0.012$ & $\phantom{0}1.972{\pm}\phantom{0}0.236$ & $0.045{\pm}0.005$ \\
\hline
\multirow{3}{*}{SAM3} & Axial+MIND & $0.813{\pm}0.151$ & $12.686{\pm}\phantom{0}9.633$ & $0.085{\pm}0.027$ & $0.781{\pm}0.013$ & $\phantom{0}2.045{\pm}\phantom{0}0.261$ & $0.044{\pm}0.006$ \\
 & \pcatd{}+MIND & $0.706{\pm}0.281$ & $26.974{\pm}36.991$ & $0.100{\pm}0.030$ & $0.775{\pm}0.014$ & $\phantom{0}2.088{\pm}\phantom{0}0.245$ & $0.038{\pm}0.006$ \\
 & \wpls{}+MIND & $0.860{\pm}0.061$ & $\mathbf{\phantom{0}9.307{\pm}\phantom{0}4.437}$ & $0.103{\pm}0.032$ & $0.788{\pm}0.013$ & $\phantom{0}1.970{\pm}\phantom{0}0.225$ & $0.043{\pm}0.005$ \\
\hline
\end{tabular}
\end{adjustbox}
\end{table}

Based on the results in Tables~\ref{tab:reg_main_base} and~\ref{tab:reg_main_mind}, we would like to highlight three results. First, VoxCor, i.e., WPLS with different ViT-based encoders, matched the handcrafted and learned alternatives in registration accuracy, i.e., MIND~\cite{heinrich2013ssc}, Anatomix~\cite{dey2025anatomix}, and DINOv2 Axial~\cite{song2024dinoreg}. On HCP T2w--T1w in the base-feature setting in Table~\ref{tab:reg_main_base}, MedSAM2 with \wpls{} reached Dice $0.797$ and HD95 $1.91$\,mm, slightly exceeding MIND ($0.794$ Dice, $1.93$\,mm). On Abdomen MR--CT, the strongest MIND-augmented results were obtained by Anatomix+MIND ($0.868$ Dice) and ViT+\wpls{}+MIND variants, with DINOv3 \wpls{}+MIND at $0.863$ and SAM3 \wpls{}+MIND at $0.860$ Dice. Thus, VoxCor features were competitive with both handcrafted local descriptors and learned 3D features for deformable registration. We also note that Dice scores for the brain dataset are generally lower than those for the abdomen dataset. This likely reflects the more challenging label structure in these images, which are used to compute the Dice scores. 

Second, when CNN- or ViT-based features were concatenated with MIND, the performance increased for almost all of them. This can be directly observed by comparing the corresponding rows in Tables~\ref{tab:reg_main_base} and~\ref{tab:reg_main_mind}. 
These results suggest that CNN- or ViT-based features do not capture local details as well as MIND for the ConvexAdam algorithm. At the same time, the fact that they reached higher registration accuracy compared to MIND alone, suggests that encoder-based features provide useful contextual information complementary to MIND. 

Third, \wpls{} in almost all the cases improved registration accuracy over \pcatd{}. The improvement was consistent for different datasets and ViT backbones. This is a direct demonstration of the benefits of using correspondence-aware and modality-specific projections. The strategy in VoxCor, i.e., using \wpls{}, was able to determine a common feature subspace that yielded better registration performance. 

\paragraph{Effect of global initialization:}
Figure~\ref{fig:reg_gica_compare} compares directly using ConvexAdam (CA) with first using \bandslice{} to account for global misalignments and then applying ConvexAdam, i.e., Globally-Initialized ConvexAdam (\gica{}). 
Results are presented only for the MIND-augmented setting, which led to better results. First of all, the effect was dataset-dependent. On Abdomen MR--CT, \gica{} improved most of the configurations, except \pcatd{} results.  
The global misalignments due to variations in field-of-view were causing ConvexAdam-based deformable registration to fail. 
Accounting for these misalignments with \bandslice{} improved the following ConvexAdam's performance. 
On HCP T2w--T1w, on the other hand, the two settings were nearly indistinguishable across methods. 
This is not surprising because global misalignments in the HCP dataset were less severe than those in the abdomen dataset. 
ConvexAdam was able to address these less severe misalignments, and the contribution of \bandslice{} was not important. 
These results suggest that \bandslice{} can be a good initialization for feature-based deformable registration using the ConvexAdam algorithm, specifically when global misalignments are expected. 

\paragraph{Pair-Fit versus Dataset-Fit:}
Figure~\ref{fig:reg_fit_compare} compares reusable Dataset-Fit with pair-specific Pair-Fit under \gica{} for ViT-based +MIND configurations. 
On Abdomen MR--CT, Pair-Fit performed better than Dataset-Fit for \wpls{}+MIND using DINOv2, DINOv3, and SAM3 features. 
Axial+MIND and MedSAM2 variants did not show a substantial difference between Pair-Fit and Dataset-Fit.  
On HCP T2w--T1w, the two regimes were nearly indistinguishable across all encoders and feature types.
This indicates that reusable fitted representations transferred well in the HCP setting, where field-of-views of different images were not different. 
In this light, the larger Abdomen-specific Pair-Fit gains, especially for \wpls{}, suggest that correspondence-aware projections may be more sensitive to anatomical and field-of-view variability. 
In practical terms, Dataset-Fit is a reliable reusable setting for normalized data such as HCP, while pair-specific fitting can still improve performance under more heterogeneous Abdomen MR--CT registration.

\subsection{Feature Quality via kNN Segmentation}
\label{ssec:results_knn}

The main finding of the kNN segmentation experiments is that \wpls{} showed a clear advantage in the Different Modality (DM) and Generalization (G) categories, where modality changes between the key and query volumes in the former, and both subject and modality change in the latter. The advantage was especially clear when \wpls{} was used with DINOv3. All kNN segmentation results are reported in Table~\ref{tab:knn_all_groups}, under the \textbf{Dataset-Fit} setting: the per-axis PCA, \pcatd{}, and \wpls{} projections were fitted only on the training split and then applied to held-out test volumes, specifically, to the key and query volumes. This corresponds to the reusable deployment regime of VoxCor rather than pair-specific adaptation. A sensitivity analysis over $k \in \{1,3,5,7,9,11\}$ is provided in Appendix~\ref{sec:app:knn_sensk}; the main conclusions were stable across this range. For ViT-based representations, variation across $k$ was small relative to the cross-method differences discussed here.
Higher sensitivities were observed in self-consistency for descriptors with a strong local component, namely MIND on both datasets and Anatomix+MIND on HCP, and did not affect the transfer-category ordering.

\begin{figure}[h!]
    \centering
    \includegraphics[width=0.8\linewidth]{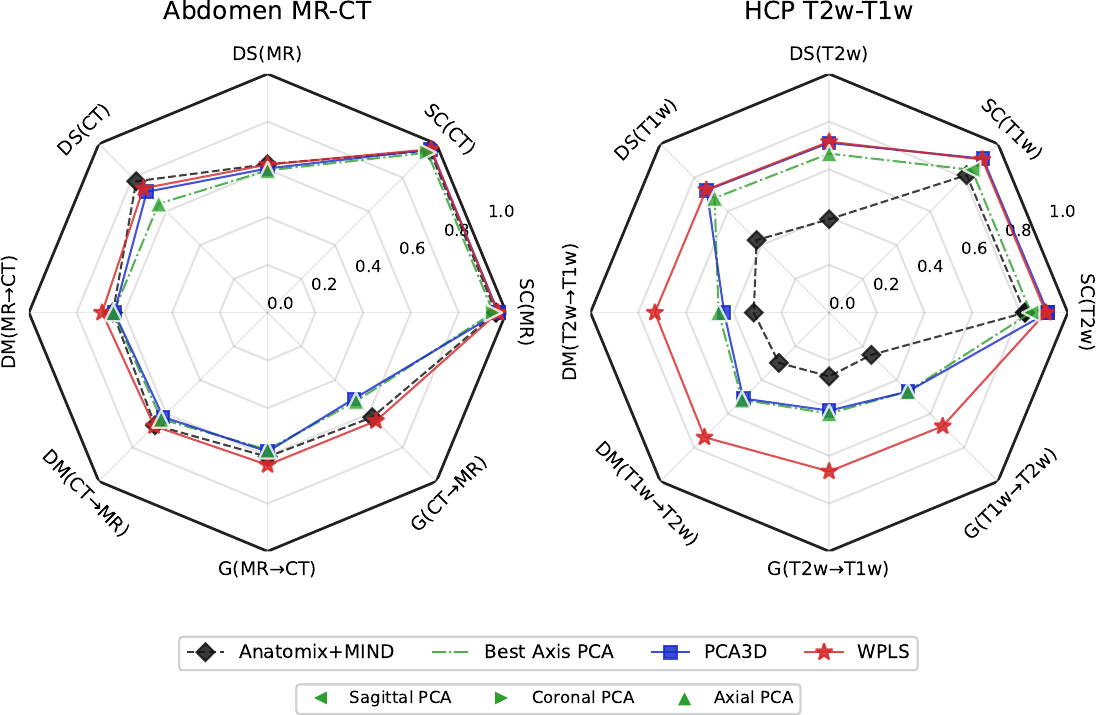}
    \caption{DINOv3 voxelwise kNN segmentation Dice radar plots under the \textbf{Dataset-Fit} setting with $k=7$. Axes show direction-specific Self-Consistency (SC), Different Subject (DS), Different Modality (DM), and Generalization (G) groups for Abdomen MR--CT and HCP T2w--T1w. Marker shapes on the Best Axis PCA curve indicate which single-axis PCA feature was selected for each direction-specific group. \wpls{} dominates the DM and G categories on both datasets, with the difference from \pcatd{} largest in the cross-modality categories.}
    \label{fig:knn_radar}
\end{figure}

Unlike registration, this experiment obtained correspondences by nearest-neighbor search in the transformed feature space, and tested whether these neighbors carried the same anatomical labels across changes in subject identity and imaging modality. The setting therefore also differs from evaluating within-image feature self-consistency: it probes how features generalize across volumes. For Abdomen MR--CT, the most informative categories are DS, DM, and G; note that even same-patient MR--CT pairs were misaligned in the original image space. For HCP T2w--T1w, the DS and G categories were arguably more informative than DM, because different contrasts of the same subject are largely aligned. The DINOv3 direction-specific results are shown in further detail in Fig.~\ref{fig:knn_radar}; the corresponding radar plots and direction-specific tables for all encoders are provided in Appendix~\ref{sec:app:knn}.

\begin{table}
\centering
\caption{Voxelwise kNN segmentation Dice ($k=7$) under the \textbf{Dataset-Fit} setting. Each cell reports mean$\pm$standard deviation pooled across both transfer directions within the category (e.g.\ MR$\to$CT and CT$\to$MR for Abdomen DM). The table reports DS, DM, and G transfer categories; Self-Consistency is omitted because it is uniformly high for ViT-based methods. Best Axis PCA selects the best of sagittal, coronal, or axial PCA per dataset and category. Bold marks the best result per column.}
\label{tab:knn_all_groups}
\vspace{4pt}
\setlength{\tabcolsep}{3.2pt}
\begin{adjustbox}{width=\textwidth}
\begin{tabular}{l|l|ccc||ccc}
\hline
\multirow{2}{*}{\textbf{Encoder}} & \multirow{2}{*}{\textbf{Method}} & \multicolumn{3}{c||}{\textbf{Abdomen MR--CT}} & \multicolumn{3}{c}{\textbf{HCP T2w--T1w}} \\
 & & DS $\uparrow$ & DM $\uparrow$ & G $\uparrow$ & DS $\uparrow$ & DM $\uparrow$ & G $\uparrow$ \\
\hline\hline
\multirow{3}{*}{--} & MIND \cite{heinrich2013ssc} & 0.063$\pm$0.040 & 0.064$\pm$0.020 & 0.043$\pm$0.008 & 0.104$\pm$0.006 & 0.116$\pm$0.007 & 0.085$\pm$0.004 \\
 & Anatomix \cite{dey2025anatomix} & 0.551$\pm$0.155 & 0.557$\pm$0.139 & 0.461$\pm$0.134 & 0.275$\pm$0.019 & 0.195$\pm$0.018 & 0.186$\pm$0.017 \\
 & Anatomix+MIND \cite{dey2025anatomix} & \textbf{0.700$\pm$0.145} & 0.658$\pm$0.174 & 0.612$\pm$0.141 & 0.410$\pm$0.029 & 0.307$\pm$0.021 & 0.259$\pm$0.017 \\
\hline
\multirow{3}{*}{DINOv2} & Best Axis PCA & 0.624$\pm$0.135 & 0.576$\pm$0.127 & 0.523$\pm$0.092 & 0.659$\pm$0.014 & 0.462$\pm$0.034 & 0.420$\pm$0.026 \\
 & \pcatd{} & 0.666$\pm$0.147 & 0.629$\pm$0.160 & 0.577$\pm$0.113 & 0.716$\pm$0.016 & 0.522$\pm$0.040 & 0.483$\pm$0.037 \\
 & \wpls{} & 0.689$\pm$0.147 & 0.655$\pm$0.150 & 0.605$\pm$0.116 & \textbf{0.724$\pm$0.016} & 0.724$\pm$0.011 & 0.665$\pm$0.014 \\
\hline
\multirow{3}{*}{DINOv3} & Best Axis PCA & 0.621$\pm$0.128 & 0.638$\pm$0.129 & 0.549$\pm$0.131 & 0.672$\pm$0.016 & 0.490$\pm$0.038 & 0.445$\pm$0.036 \\
 & \pcatd{} & 0.660$\pm$0.166 & 0.630$\pm$0.172 & 0.547$\pm$0.126 & 0.718$\pm$0.020 & 0.474$\pm$0.042 & 0.439$\pm$0.037 \\
 & \wpls{} & 0.678$\pm$0.156 & \textbf{0.683$\pm$0.155} & \textbf{0.640$\pm$0.159} & 0.722$\pm$0.017 & \textbf{0.734$\pm$0.013} & \textbf{0.669$\pm$0.016} \\
\hline
\multirow{3}{*}{MedSAM2} & Best Axis PCA & 0.359$\pm$0.152 & 0.329$\pm$0.094 & 0.248$\pm$0.069 & 0.540$\pm$0.040 & 0.114$\pm$0.011 & 0.108$\pm$0.010 \\
 & \pcatd{} & 0.369$\pm$0.177 & 0.319$\pm$0.127 & 0.243$\pm$0.084 & 0.579$\pm$0.042 & 0.110$\pm$0.008 & 0.105$\pm$0.008 \\
 & \wpls{} & 0.430$\pm$0.205 & 0.394$\pm$0.166 & 0.330$\pm$0.164 & 0.585$\pm$0.044 & 0.466$\pm$0.019 & 0.387$\pm$0.018 \\
\hline
\multirow{3}{*}{SAM3} & Best Axis PCA & 0.296$\pm$0.136 & 0.241$\pm$0.066 & 0.195$\pm$0.054 & 0.490$\pm$0.031 & 0.164$\pm$0.012 & 0.142$\pm$0.012 \\
 & \pcatd{} & 0.284$\pm$0.128 & 0.222$\pm$0.066 & 0.183$\pm$0.055 & 0.546$\pm$0.039 & 0.197$\pm$0.022 & 0.170$\pm$0.018 \\
 & \wpls{} & 0.319$\pm$0.142 & 0.253$\pm$0.072 & 0.220$\pm$0.063 & 0.598$\pm$0.040 & 0.499$\pm$0.018 & 0.422$\pm$0.025 \\
\hline
\end{tabular}
\end{adjustbox}
\end{table}

The main separation between methods appeared in the transfer categories involving modality change, DM and G, while DS showed that strong CNN-based feature combinations could still be competitive when only subject identity changed. MIND alone gave low kNN Dice in all transfer categories, suggesting that a local self-similarity descriptor that is effective for deformable registration does not by itself define a useful feature space for direct nearest-neighbor label transfer. Anatomix performed comparatively better, consistent with its more semantic, segmentation-oriented representation. Combining Anatomix with MIND further improved all DS, DM, and G scores on both datasets. This suggests that Anatomix+MIND benefits from combining more global anatomical information with local self-similarity cues. However, this combination was still not sufficient to provide the highest consistency under modality transfer. 

Among the ViT-based representations, \wpls{} consistently improved over both Best Axis PCA and \pcatd{} in the hardest categories, DM and G. Larger differences were observed for HCP T2w--T1w tests. 
The DS category was more mixed. 
\wpls{} still improved over the Best Axis PCA and \pcatd{}. However, Anatomix+MIND achieved strong kNN segmentation performance in the Abdomen MR--CT dataset. For the HCP T2w--T1w dataset, the \wpls{} method showed an advantage even in the DS category. 
Overall, these results suggest that \wpls{} provided its clearest advantage when modality transfer was required, while also preserving strong cross-subject performance; the advantage over modality-agnostic PCA projections was especially large on HCP, where \wpls{} produced feature neighborhoods that were less contrast-dominated and more anatomically consistent.

Encoder choice also had a strong effect. Features extracted by MedSAM2 and SAM3 did not perform as well as DINO features. \wpls{} provided some improvements for these encoders compared to using Best Axis PCA and \pcatd{}, but the differences between encoders were more dominant. 
This indicates that correspondence-aware projection and encoder choice were complementary: \wpls{} aligned modality-specific feature spaces, but the best cross-subject, cross-modality neighborhoods were obtained when the backbone already provided transferable anatomical structure.

\subsection{Geometric Precision of Registration-free Correspondence}
\label{ssec:results_lm}
 
The registration-free correspondence experiment tested the same claim: whether, for a given voxel in one image, the voxel in the other image that is closest in the feature space lies geometrically close to the corresponding anatomical reference point. 
This test is similar in essence to the kNN segmentation test but stricter as it is pointwise. 
Instead of manually annotating landmarks, we derived reference points automatically as the centers of mass of selected segmentation labels in each volume. All results are reported in Table~\ref{tab:landmark_all_groups_medianpair} under the \textbf{Dataset-Fit} setting with $k=1$, i.e., only considering the closest voxel, and $L_2$ distance in the feature space. 
As in the kNN setting, the main table omits Self-Consistency and reports DS, DM, and G settings; the appendix includes the full direction-specific results.

\begin{figure*}
    \centering
    \includegraphics[width=\linewidth]{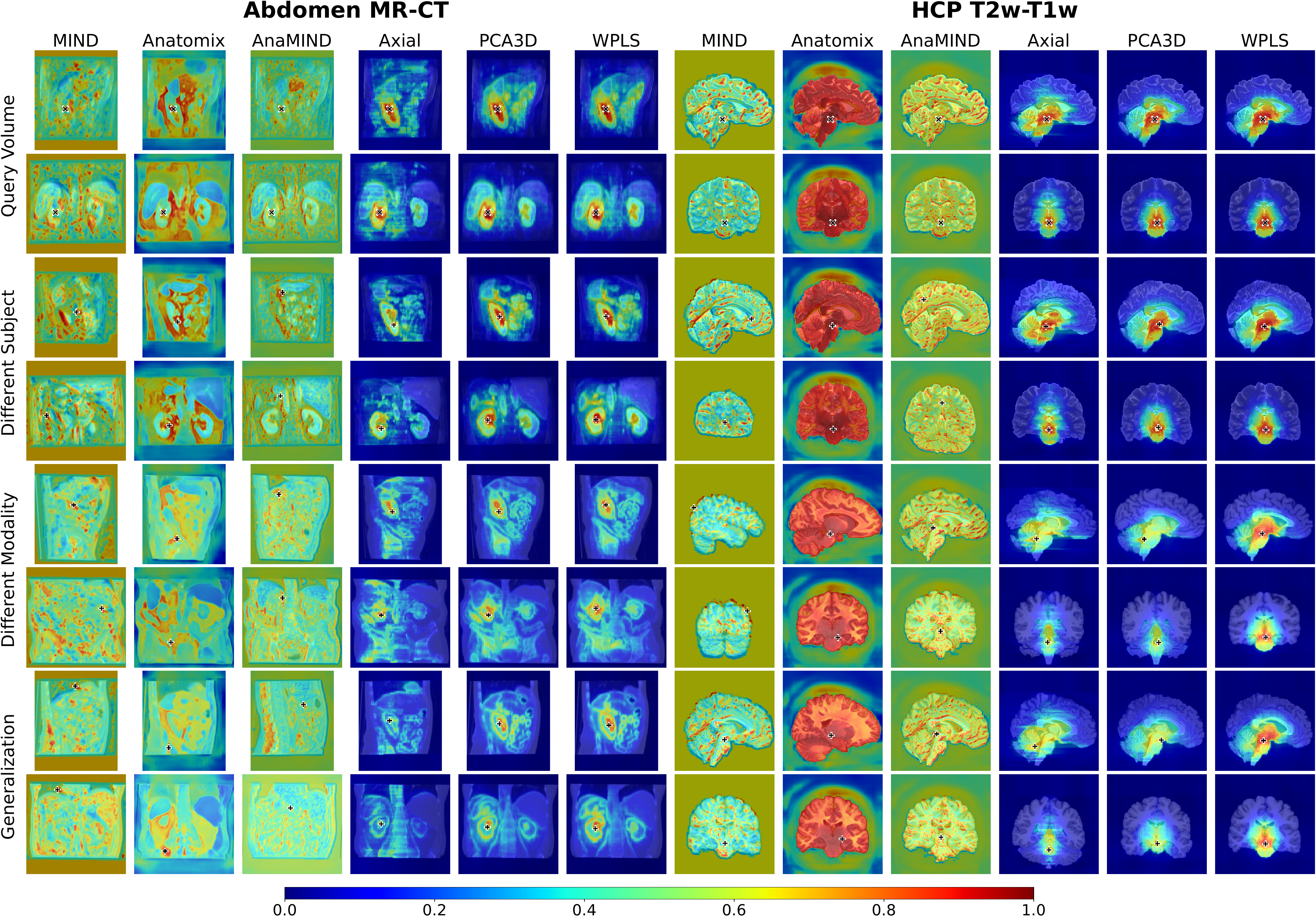}
    \caption{Qualitative direct feature-space correspondence on Abdomen MR--CT (right kidney) and HCP T2w--T1w (hippocampus). Columns are grouped by dataset and method, and rows are grouped by transfer category, with two adjacent rows per category showing sagittal and coronal views. The cross markers indicate the query landmarks and the plus markers indicate the maximum-similarity voxels in the target volumes. Columns compare MIND, Anatomix, Anatomix+MIND (AnaMIND), and DINOv3 Dataset-Fit Axial PCA, \pcatd{}, and \wpls{} within each dataset. Similarity maps are sharpened for visualization only by mapping cosine similarity $c$ to $(\exp(\tau(c+1)/2)-1)/(\exp(\tau)-1)$ with $\tau{=}5$. The figure is best viewed in color.}
    \label{fig:qual_correspondence_maps}
\end{figure*}

Figure~\ref{fig:qual_correspondence_maps} provides a qualitative example of the registration-free correspondence task. For a fixed query landmark, the feature vector at the query location was compared with all voxels in the target volume. The resulting feature similarity maps showed that MIND and Anatomix-based descriptors often produced broad or displaced high-similarity regions, whereas DINOv3 \pcatd{} and \wpls{} produced more compact maps centered around the corresponding anatomical regions, with \wpls{} maintaining sharper compactness in the Different Subject and Generalization columns.

\begin{table}
\centering
\caption{Registration-free correspondence error under the \textbf{Dataset-Fit} setting using top-$1$ nearest-neighbor matching in feature space ($k=1$, $L_2$ distance). Distances are reported in millimeters as mean$\pm$standard deviation across landmarks, using isotropic voxel sizes of $2.0\times2.0\times2.0$\,mm$^3$ for Abdomen MR--CT and $0.7\times0.7\times0.7$\,mm$^3$ for HCP T2w--T1w. Each landmark value is first computed as the median over held-out query--key pairs within each category (median-pair aggregation; pooled-mean aggregation in Appendix~\ref{sec:app:landmark_pooledmean}). HCP T2w--T1w results use the expanded 24-subject held-out set described in Section~\ref{ssec:datasets}. The table reports DS, DM, and G; Self-Consistency is omitted. Lower is better. Bold marks the best result per column.}
\label{tab:landmark_all_groups_medianpair}
\vspace{4pt}
\setlength{\tabcolsep}{2.1pt}
\begin{adjustbox}{width=\textwidth}
\begin{tabular}{l|l|ccc||ccc}
\hline
\multirow{2}{*}{\textbf{Encoder}} & \multirow{2}{*}{\textbf{Method}} & \multicolumn{3}{c||}{\textbf{Abdomen MR--CT}} & \multicolumn{3}{c}{\textbf{HCP T2w--T1w}} \\
 & & DS $\downarrow$ & DM $\downarrow$ & G $\downarrow$ & DS $\downarrow$ & DM $\downarrow$ & G $\downarrow$ \\
\hline\hline
\multirow{3}{*}{--} & MIND \cite{heinrich2013ssc}& 167.68$\pm$\phantom{0}27.34 & 165.26$\pm$\phantom{0}40.17 & 156.83$\pm$\phantom{00}8.16 & 55.90$\pm$\phantom{0}7.48 & 55.66$\pm$\phantom{0}7.34 & 57.67$\pm$\phantom{0}8.15 \\
 & Anatomix \cite{dey2025anatomix} & \phantom{0}41.33$\pm$\phantom{0}16.81 & \phantom{0}40.78$\pm$\phantom{0}21.89 & \phantom{0}51.79$\pm$\phantom{0}14.33 & 18.62$\pm$10.64 & 20.91$\pm$13.02 & 27.31$\pm$11.73 \\
 & Anatomix+MIND \cite{dey2025anatomix} & \phantom{0}44.62$\pm$\phantom{0}11.51 & \phantom{0}44.52$\pm$\phantom{0}15.80 & \phantom{0}69.68$\pm$\phantom{0}26.45 & 23.46$\pm$12.10 & 26.00$\pm$12.38 & 28.89$\pm$10.68 \\
\hline
\multirow{3}{*}{DINOv2} & Best Axis PCA & \phantom{0}30.10$\pm$\phantom{00}8.03 & \phantom{0}28.07$\pm$\phantom{0}10.14 & \phantom{0}41.62$\pm$\phantom{0}20.01 & \phantom{0}4.31$\pm$\phantom{0}0.91 & \phantom{0}6.87$\pm$\phantom{0}3.63 & \phantom{0}7.77$\pm$\phantom{0}3.31 \\
 & \pcatd{} & \phantom{0}27.86$\pm$\phantom{0}11.57 & \phantom{0}29.17$\pm$\phantom{0}18.08 & \phantom{0}32.15$\pm$\phantom{0}11.78 & \phantom{0}4.05$\pm$\phantom{0}0.89 & \phantom{0}6.19$\pm$\phantom{0}3.19 & \phantom{0}7.26$\pm$\phantom{0}2.77 \\
 & \wpls{} & \phantom{0}29.50$\pm$\phantom{00}9.58 & \phantom{0}24.12$\pm$\phantom{00}5.71 & \phantom{0}26.07$\pm$\phantom{00}8.25 & \phantom{0}4.06$\pm$\phantom{0}0.88 & \phantom{0}3.93$\pm$\phantom{0}1.26 & \phantom{0}5.76$\pm$\phantom{0}1.51 \\
\hline
\multirow{3}{*}{DINOv3} & Best Axis PCA & \phantom{0}28.47$\pm$\phantom{00}5.90 & \phantom{0}24.58$\pm$\phantom{00}7.78 & \phantom{0}32.87$\pm$\phantom{00}9.14 & \phantom{0}4.29$\pm$\phantom{0}1.16 & \phantom{0}5.39$\pm$\phantom{0}2.59 & \phantom{0}7.21$\pm$\phantom{0}2.50 \\
 & \pcatd{} & \phantom{0}22.95$\pm$\phantom{00}2.94 & \phantom{0}20.58$\pm$\phantom{00}2.80 & \phantom{0}25.00$\pm$\phantom{00}7.05 & \phantom{0}3.76$\pm$\phantom{0}1.01 & \phantom{0}5.52$\pm$\phantom{0}2.84 & \phantom{0}6.60$\pm$\phantom{0}2.45 \\
 & \wpls{} & \textbf{\phantom{0}20.73$\pm$\phantom{00}3.87} & \textbf{\phantom{0}19.52$\pm$\phantom{00}0.69} & \textbf{\phantom{0}24.45$\pm$\phantom{00}4.74} & \textbf{\phantom{0}3.60$\pm$\phantom{0}0.95} & \textbf{\phantom{0}2.86$\pm$\phantom{0}0.56} & \textbf{\phantom{0}4.39$\pm$\phantom{0}0.84} \\
\hline
\multirow{3}{*}{MedSAM2} & Best Axis PCA & \phantom{0}84.59$\pm$\phantom{0}38.97 & \phantom{0}92.65$\pm$\phantom{0}22.15 & 101.09$\pm$\phantom{0}32.35 & \phantom{0}9.85$\pm$\phantom{0}6.80 & 34.22$\pm$12.73 & 36.95$\pm$\phantom{0}9.70 \\
 & \pcatd{} & \phantom{0}92.31$\pm$\phantom{0}37.01 & \phantom{0}99.19$\pm$\phantom{0}33.93 & 102.04$\pm$\phantom{0}10.56 & \phantom{0}8.31$\pm$\phantom{0}3.45 & 37.34$\pm$12.07 & 37.86$\pm$11.41 \\
 & \wpls{} & \phantom{0}73.59$\pm$\phantom{0}37.38 & \phantom{0}87.27$\pm$\phantom{0}29.23 & \phantom{0}70.16$\pm$\phantom{0}14.17 & \phantom{0}6.65$\pm$\phantom{0}1.70 & 17.99$\pm$\phantom{0}9.74 & 25.40$\pm$\phantom{0}9.33 \\
\hline
\multirow{3}{*}{SAM3} & Best Axis PCA & 114.54$\pm$\phantom{0}28.93 & 118.06$\pm$\phantom{0}14.65 & 121.51$\pm$\phantom{0}27.79 & 11.37$\pm$\phantom{0}5.20 & 29.71$\pm$11.05 & 30.65$\pm$\phantom{0}7.61 \\
 & \pcatd{} & 100.28$\pm$\phantom{0}11.90 & 118.61$\pm$\phantom{0}17.97 & 123.90$\pm$\phantom{0}44.34 & \phantom{0}8.49$\pm$\phantom{0}2.88 & 23.27$\pm$10.86 & 26.22$\pm$\phantom{0}7.92 \\
 & \wpls{} & 108.63$\pm$\phantom{0}14.14 & 110.49$\pm$\phantom{0}11.77 & 126.21$\pm$\phantom{0}32.19 & \phantom{0}6.32$\pm$\phantom{0}1.51 & 14.11$\pm$\phantom{0}6.14 & 16.08$\pm$\phantom{0}5.38 \\
\hline
\end{tabular}
\end{adjustbox}
\end{table}

\begin{figure*}
    \centering
    \includegraphics[width=\textwidth]{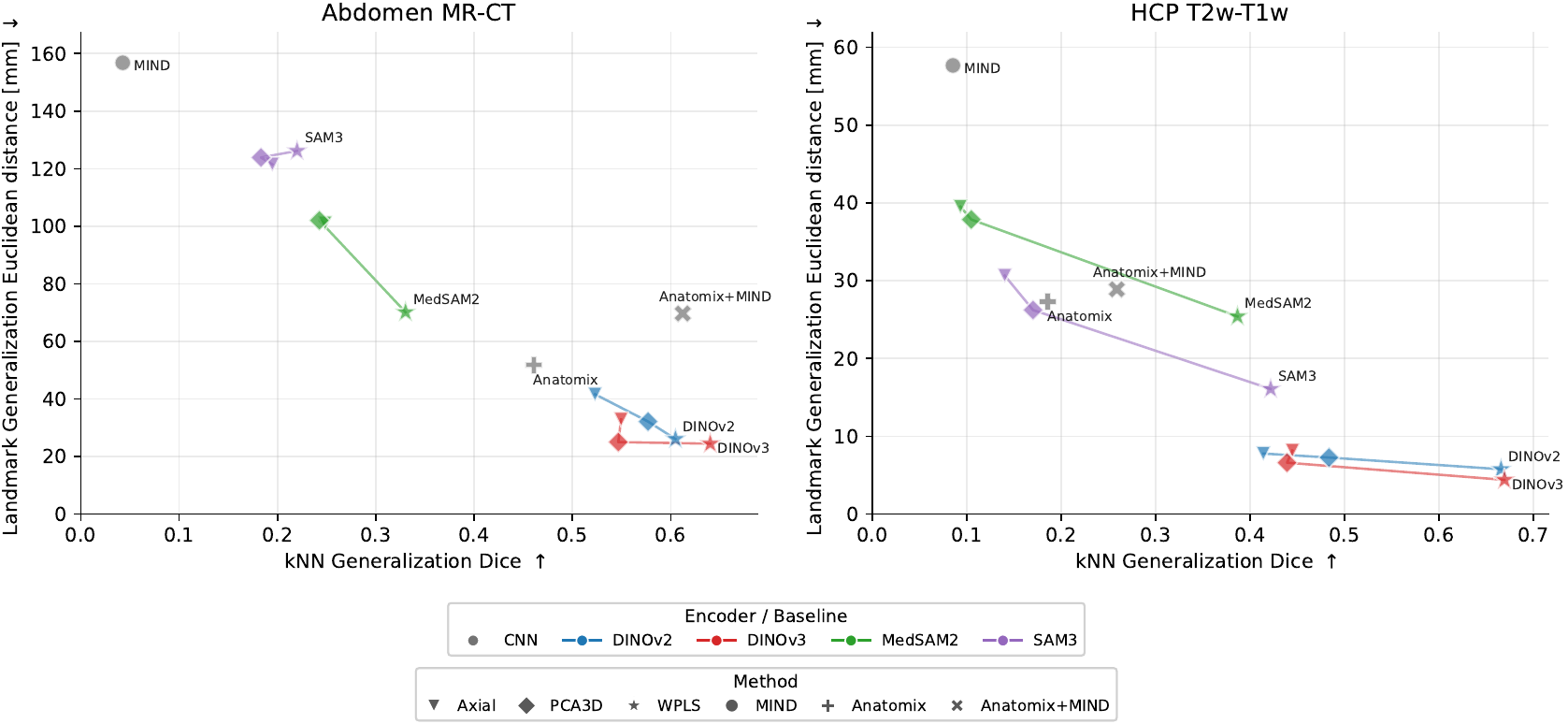}
    \caption{Semantic-versus-geometric correspondence in the Generalization (G) category. Each panel plots voxelwise kNN G-Dice ($x$-axis, higher is better) against median-pair landmark G-distance in millimeters ($y$-axis, lower is better); the lower-right corner is best. Each colored trajectory connects Axial PCA $\rightarrow$ \pcatd{} $\rightarrow$ \wpls{} for one frozen encoder, with stars marking the \wpls{} endpoint. Grey markers show CNN baselines. \wpls{} pushes every ViT encoder toward higher G-Dice and lower G-distance simultaneously.}
    \label{fig:knn_lm_medpair_g_trajectory}
\end{figure*}

Table~\ref{tab:landmark_all_groups_medianpair} reports correspondence errors for DS, DM, and G under \emph{median-pair aggregation}: for each anatomical landmark, errors are first summarized by the median over held-out query--key pairs in the corresponding category, and the table value is then the mean and standard deviation of these per-landmark medians. Figure~\ref{fig:knn_lm_medpair_g_trajectory} additionally plots G-category landmark errors against G-category kNN segmentation Dice using the same aggregation, with the projection variants for each ViT encoder linked by lines along the trajectory Axial PCA $\rightarrow$ \pcatd{} $\rightarrow$ \wpls{}.

Results on the registration-free correspondence task confirmed the main trend that was also observed in kNN segmentation, but under a stricter pointwise criterion. On HCP T2w--T1w, the error in the G category decreased along the projection trajectory for the DINO-based encoders: DINOv3 improved from $7.21$\,mm with Best Axis PCA to $6.60$\,mm with \pcatd{} and further to $4.39$\,mm with \wpls{}, while DINOv2 improved from $7.77$\,mm to $7.26$\,mm and $5.76$\,mm. Thus, triplanar aggregation improved over the best single-axis representation, and correspondence-aware projection provided the largest additional gain in the hardest cross-subject, cross-modality setting. Together with the kNN results, these findings indicate that \wpls{} can improve not only label-level neighborhood consistency but also the geometric precision of direct feature-space correspondences.

On Abdomen MR--CT, the landmark task was first of all affected by coarser voxel resolution. This dataset has $2.0\times2.0\times2.0$\,mm$^3$ isotropic resolution compared with $0.7\times0.7\times0.7$\,mm$^3$ in HCP. Differences in voxel size therefore led to larger errors in millimeters. Second, limited field-of-views, especially in MR images, made the correspondence problem harder, and sometimes led to outliers. 
For this reason, Table~\ref{tab:landmark_all_groups_medianpair} reports median-pair aggregation; however, the alternative pooled-mean aggregation is provided in Appendix~\ref{sec:app:landmark_pooledmean} for completeness. 

In median values, DINOv3 \wpls{} gave the lowest errors in all three transfer categories and both datasets, 
with DINOv3 \pcatd{} second in each case. The handcrafted and CNN baselines were markedly weaker for pointwise correspondence. \wpls{} using DINOv3 features, especially, led to a clear improvement. 
This shows again the separation between features that can support deformable optimization from features whose raw nearest-neighbor structure is geometrically precise without registration.

Reiterating kNN segmentation conclusions, a strong encoder dependence was observed in the single-voxel landmark task. 
MedSAM2 and SAM3 backbones did not provide results at the same accuracy level as DINO backbones. Even though \wpls{} was observed to improve the performance of their features, the improvement did not close the gap. 
Figure~\ref{fig:knn_lm_medpair_g_trajectory} shows the same pattern jointly with kNN segmentation: along the Axial PCA $\rightarrow$ \pcatd{} $\rightarrow$ \wpls{} trajectory, \wpls{} generally moved ViT encoders toward higher G-Dice and lower G-distance at the same time. Therefore, correspondence-aware projection improved modality-stable neighborhood structure in both semantic and geometric terms, but precise point correspondence still depended strongly on the underlying ViT backbone.

\subsection{Computational Characteristics}
\label{ssec:results_compute}

\begin{table}
\centering
\caption{Computational cost of feature fitting and transform for one paired two-volume sample. Fit time and Fit GPU refer to offline projector fitting; Total time and Max GPU also include the immediately subsequent transform. Feature Transform columns report applying the fitted extractor to both volumes after fitting; per-volume transform cost is approximately half of the reported value. Downstream registration, kNN search, and landmark matching are not included.}
\label{tab:benchmark_compute}
\vspace{4pt}
\scriptsize
\setlength{\tabcolsep}{2.2pt}
\begin{adjustbox}{width=\textwidth}
\begin{tabular}{lll|cccc|cccc|cc}
\hline
\multirow{2}{*}{\textbf{Dataset}} & \multirow{2}{*}{\textbf{Source}} & \multirow{2}{*}{\textbf{Method}} & \multicolumn{4}{c|}{\textbf{Pair-Fit}} & \multicolumn{4}{c|}{\textbf{Dataset-Fit}} & \multicolumn{2}{c}{\textbf{Feature Transform}} \\
 & & & Fit [s] & Fit GPU [GB] & Total [s] & Max GPU [GB] & Fit [s] & Fit GPU [GB] & Total [s] & Max GPU [GB] & Time [s] & GPU [GB] \\
\hline\hline
\multirow{6}{*}{\shortstack{Abdomen\\MR--CT}} & \multirow{2}{*}{CNN} & MIND & -- & -- & -- & -- & -- & -- & -- & -- & $0.21$ & $1.07$ \\
 &  & Anatomix & -- & -- & -- & -- & -- & -- & -- & -- & $3.56$ & $3.54$ \\
\cline{2-13}
 & \multirow{4}{*}{ViT} & DINOv2 & $73.3$ & $7.93$ & $115.7$ & $12.7$ & $442.2$ & $33.1$ & $484.6$ & $33.1$ & $42.4$ & $12.7$ \\
 &  & DINOv3 & $85.8$ & $7.92$ & $132.1$ & $12.7$ & $524.1$ & $33.4$ & $570.4$ & $33.4$ & $46.3$ & $12.7$ \\
 &  & MedSAM2 & $38.6$ & $11.9$ & $62.9$ & $16.7$ & $238.1$ & $21.8$ & $262.5$ & $21.8$ & $24.3$ & $16.7$ \\
 &  & SAM3 & $94.0$ & $10.7$ & $146.6$ & $15.5$ & $576.1$ & $16.5$ & $628.8$ & $16.5$ & $52.7$ & $15.5$ \\
\hline
\multirow{6}{*}{\shortstack{HCP\\T2w--T1w}} & \multirow{2}{*}{CNN} & MIND & -- & -- & -- & -- & -- & -- & -- & -- & $0.71$ & $3.04$ \\
 &  & Anatomix & -- & -- & -- & -- & -- & -- & -- & -- & $24.5$ & $4.31$ \\
\cline{2-13}
 & \multirow{4}{*}{ViT} & DINOv2 & $117.4$ & $21.3$ & $189.7$ & $35.2$ & $680.7$ & $43.1$ & $752.9$ & $43.1$ & $72.2$ & $35.2$ \\
 &  & DINOv3 & $104.1$ & $21.3$ & $173.6$ & $35.2$ & $608.0$ & $43.0$ & $677.5$ & $43.0$ & $69.5$ & $35.2$ \\
 &  & MedSAM2 & $46.2$ & $22.1$ & $84.1$ & $36.0$ & $259.0$ & $43.8$ & $296.8$ & $43.8$ & $37.8$ & $36.0$ \\
 &  & SAM3 & $127.4$ & $24.0$ & $195.5$ & $38.0$ & $744.8$ & $40.7$ & $812.9$ & $40.7$ & $68.1$ & $38.0$ \\
\hline
\end{tabular}
\end{adjustbox}
\end{table}

We finally present the computational characteristics of VoxCor. Table~\ref{tab:benchmark_compute} reports offline fitting cost, transform-time feature extraction cost, and GPU allocation for one paired two-volume sample; costs for the downstream registration, kNN search, and landmark matching are not included. The fitting stage corresponds to estimating the projection parameters, including the correspondence-aware \wpls{} projection, rather than training or fine-tuning a neural network. On Abdomen MR--CT, Pair-Fit required approximately $39$--$94$\,s depending on the encoder, while Dataset-Fit on six pairs required approximately $238$--$576$\,s. On HCP T2w--T1w, the larger $256^3$ volumes increased this to approximately $46$--$127$\,s for Pair-Fit and $259$--$745$\,s for Dataset-Fit. Thus, fitting is not negligible, but remains short compared with training a new model and, in the Dataset-Fit regime, is amortized across subsequent transformed volumes. The higher Dataset-Fit memory use, reaching $33.4$\,GB on Abdomen MR--CT and $43.8$\,GB on HCP T2w--T1w, reflects the larger number of fit pairs and the cost of fitting a reusable projection.

After fitting, transforming new volumes requires frozen ViT inference, interpolation/resizing, and application of the stored linear projections. The transform columns in Table~\ref{tab:benchmark_compute} report the cost for a paired two-volume sample, so the approximate per-volume transform cost is half of the listed value. This stage was significantly more expensive than handcrafted or CNN descriptors: ViT-based transforms required approximately $24$--$53$\,s and $12.7$--$16.7$\,GB on Abdomen MR--CT, and $38$--$72$\,s and $35.2$--$38.0$\,GB on HCP T2w--T1w, whereas MIND required less than $1$\,s and Anatomix required $3.6$\,s on Abdomen and $24.5$\,s on HCP. Therefore, VoxCor trades additional transform-time computation and high-memory GPU requirements for reusable cross-modal and cross-subject feature spaces. Reducing this memory footprint and improving transform-time efficiency are important directions for future work.

\section{Discussion}
\label{sec:discussion}

VoxCor shows that frozen 2D ViT features can be converted into reusable volumetric correspondence spaces without encoder fine-tuning. The strongest evidence for this claim comes from the Dataset-Fit kNN segmentation and registration-free correspondence experiments, where the fitted projections were applied to held-out volumes and correspondences between key and query volumes were obtained directly by nearest-neighbor search in feature space. In this setting, \wpls{} provided its clearest advantage in the categories involving modality transfer, especially Generalization, where both subject identity and imaging modality changed. This indicates that triplanar aggregation alone is beneficial, but that correspondence-aware projection is important for making feature neighborhoods more anatomy-dominated and less contrast-dominated. At the same time, VoxCor features remained competitive in deformable registration, showing that the same representation can support both direct correspondence queries and optimization-based alignment.

The comparison with MIND and Anatomix highlights why these complementary evaluations were necessary. MIND remained a strong descriptor for deformable registration, particularly when coupled with ConvexAdam, but it performed poorly when used as a standalone nearest-neighbor feature space for kNN segmentation or landmark localization. This suggests that a local self-similarity descriptor can provide an effective optimization signal without necessarily defining anatomically meaningful global neighborhoods. Anatomix showed the opposite tendency more clearly: its segmentation-oriented 3D features were stronger than MIND for direct correspondence, and Anatomix+MIND was excellent in deformable registration and particularly competitive for Abdomen MR--CT kNN segmentation, but this combination did not consistently improve single-point landmark precision. In contrast, DINO-based ViT features combined with \wpls{} were more consistent across the direct correspondence tasks, especially on HCP T2w--T1w. These differences suggest that registration performance, dense label-transfer accuracy, and pointwise geometric precision reward related but distinct properties of a feature representation.

The main limitations of VoxCor are computational cost, dependence on fitting-time correspondences, and evaluation scope. Although fitting the projection is short compared with training a new model and can be amortized in the Dataset-Fit regime, transform-time feature extraction remains more expensive than MIND or Anatomix and requires high-memory GPUs, especially for high-resolution volumes such as HCP T2w--T1w. Reducing this memory footprint through chunked feature extraction, more compact feature materialization, or region-of-interest-based transforms is an important direction for future work. In addition, \wpls{} relies on fitting-time correspondences: these were assumed for subject-matched HCP T2w--T1w pairs and automatically generated by MIND+\gica{} for Abdomen MR--CT, so the fitted projection may inherit errors from the correspondence source. Future work should investigate more robust correspondence estimation, uncertainty-aware fitting, and iterative refinement of the feature space. Finally, our experiments covered two datasets with complementary but limited regimes; broader validation across anatomies, pathologies, scanners, and larger cohorts will be needed to establish VoxCor as a general-purpose multimodal correspondence layer.

\section{Conclusion}
\label{sec:conclusion}

We introduced VoxCor, a training-free fit--transform framework for constructing reusable voxelwise feature spaces from frozen 2D ViT foundation models. VoxCor combines triplanar feature extraction with closed-form projection fitting: per-axis PCA first produces compact sagittal, coronal, and axial feature volumes, and a correspondence-aware \wpls{} produces modality-specific projection matrices that map the concatenated triplanar features into a feature subspace shared by given modalities. Because the projection stage operates on voxelwise feature volumes rather than on a specific encoder architecture, the framework is model-agnostic and can be applied to future 2D or 3D foundation models that provide dense spatial features. At transform time, new volumes are processed by frozen ViT inference and stored linear projections only, without encoder fine-tuning, foreground masks, or registration.

Across deformable registration, voxelwise kNN segmentation, and registration-free correspondence, VoxCor showed that adapted frozen ViT features can support both optimization-based alignment and direct feature-space correspondence. The clearest gains appeared in the categories involving modality transfer, especially when both subject identity and modality changed, and DINO-based backbones provided the most precise semantic and geometric correspondences. These results support correspondence-aware projection of frozen ViT features as a promising route toward reusable multimodal voxel correspondence, while also highlighting the need for more efficient transform-time feature extraction, lower memory use, and broader validation across clinical imaging settings.

\section{Acknowledgements}

We acknowledge The LOOP Zurich -- Medical Research Center, Zurich, Switzerland and Georg and Berta Schwyzer-Winiker Foundation for the financial support for this project.

\newpage

\bibliographystyle{unsrtnat}
\bibliography{ref}


\newpage
\appendix

\section{Method Details}
\label{sec:app:method_details}

This section provides implementation details for the auxiliary components used by VoxCor during fitting. 
The main Method section defines the triplanar representation, the correspondence-aware \wpls{} projection, the \pcatd{} comparison, foreground masking, fitting-time correspondence generation, and \bandslice{} initialization. 
Here, we specify only details that are either implementation-specific or would otherwise interrupt the main presentation. 
Foreground masks, MIND-based correspondence generation, and \bandslice{} initialization are used only during fitting or registration evaluation. 
They are not required when transforming new volumes after the projection parameters have been fitted.

\subsection{Preprocessing and Foreground Mask Generation}
\label{sec:app:preprocess_mask}

\paragraph{Dataset-specific intensity normalization:}
Before feature extraction, each volume is normalized independently. 
For Abdomen MR--CT, we adopted the same intensity preprocessing as DINO-Reg~\cite{song2024dinoreg}: MR volumes were clipped at the 97th percentile, CT volumes were windowed with level $50$ and width $400$, and both modalities were subsequently rescaled to the range $[0,1]$.
For HCP T2w--T1w, both modalities were clipped at the 99th percentile and rescaled to $[0,1]$. 
The same normalized volumes were used for ViT feature extraction and for the auxiliary MIND-based processing steps.

\paragraph{Raw MIND foreground mask:}
We describe the foreground mask construction referenced in Section~\ref{ssec:auxiliary}. 
Let $V_n\in\mathbb{R}^{D\times H\times W}$ be a fitting-time volume and let
$m_{V_n}\in\mathbb{R}^{D\times H\times W\times C_M}$ denote its MIND descriptor volume, computed with the parameters specified in Appendix~\ref{ssec:app:internal_mind_gica}. 
A voxel $v$ is labeled as background when all MIND descriptor channels exceed a fixed threshold $\tau$:
\begin{equation}
\begin{aligned}
\mathrm{bg}_0(v)
&=
\mathbb{I}
\left[
m_{V_n}(v,c) > \tau
\;\;
\forall c \in \{1,\ldots,C_M\}
\right],
\\
\mathrm{fg}_0(v)
&=
1-\mathrm{bg}_0(v).
\end{aligned}
\end{equation}
Intuitively, MIND descriptor values are close to one in near-constant regions, because local patch differences within the descriptor neighborhood are small. 
Thus, requiring all descriptor channels to exceed a high threshold identifies homogeneous background regions.

\paragraph{Hole filling:}
The raw foreground mask may contain enclosed background pockets inside the body, for example gas or low-signal regions in abdominal MR/CT. 
We remove these by 6-connected boundary-flood hole filling. 
Let $\partial\Omega$ denote the set of voxels touching the volume boundary and let $\mathcal{N}_6$ denote the 6-connected neighborhood operator. 
The reachable background is defined as the fixed point of
\begin{equation}
\begin{aligned}
\mathcal{R}^{(0)}
&=
\mathrm{bg}_0 \cap \partial\Omega,
\\
\mathcal{R}^{(t+1)}
&=
\mathcal{R}^{(t)}
\cup
\left(
\mathrm{bg}_0
\cap
\mathcal{N}_6(\mathcal{R}^{(t)})
\right).
\end{aligned}
\end{equation}
A background voxel is therefore reachable if it is connected to the image boundary through a path of 6-connected background neighbors. 
Any background voxel that is not reachable is treated as an enclosed hole and added to the foreground:
\begin{equation}
\begin{aligned}
\mathrm{fg}(v)
=
\mathrm{fg}_0(v)
\vee
\left(
\mathrm{bg}_0(v)
\wedge
\neg \mathcal{R}^{(\infty)}(v)
\right).
\end{aligned}
\end{equation}
The fixed point is reached in finitely many iterations because $\mathcal{R}^{(t)}$ is monotonically increasing on a finite voxel grid.

In all experiments, we used $\tau=0.99$ for the raw MIND foreground threshold, MIND radius $r=2$, MIND dilation $d=2$, 6-connectivity for hole filling, no explicit iteration cap beyond the default $D+H+W$ fixed-point limit, and no final dilation.

\paragraph{Use of the mask:}
The final mask $\mathrm{fg}$ was used during fitting in four places. 
First, when fitting the per-axis joint-modality PCA, the foreground mask was pooled to the ViT token grid, and only foreground patch tokens contributed to the PCA statistics. 
Second, when fitting \wpls{}, foreground masks defined the valid correspondence region used for the weighted cross-covariance and variance estimates. 
Third, when fitting \pcatd{}, only foreground triplanar features from both modalities were pooled for the shared PCA fit. 
Fourth, the masks were used by the MIND-based registration procedure that generated fitting-time correspondences when such registration was required. 
Dataset-provided anatomical segmentations were never used for mask generation and were reserved exclusively for evaluation. 
At transform time, no foreground mask is required; the stored per-axis PCA and final projection parameters are applied densely to all patch tokens and voxels, including background regions.

\subsection{Fitting-Time Correspondence Generation}
\label{sec:app:correspondence_generation}

VoxCor/\wpls{} requires voxelwise correspondences only during fitting. 
When correspondence could already be assumed, as in the HCP T2w--T1w setting in our experiments, we used direct voxel matching, and the correspondence warp was the identity. 
When correspondence could not be assumed, as in Abdomen MR--CT, we generated fitting-time correspondences using MIND+\gica{} registration as described below. 
These correspondences were never estimated for new transform-time volumes. 
The \pcatd{} comparison did not use this step.

For each fit-stage pair $(I_n,J_n)$ requiring registration-based correspondence generation, we computed MIND descriptors with the parameters specified in the experimental setup. 
We then performed fixed-parameter MIND-based deformable registration using \gica{}, where \gica{} denotes \bandslice{} global initialization followed by ConvexAdam refinement~\cite{siebert2024convexadam}. 
The fixed hyperparameters used for this internal MIND+\gica{} procedure are reported in Table~\ref{tab:internal_mind_params}. 
We denote the final displacement field returned after the ConvexAdam refinement stage, initialized by the preceding \bandslice{} global transform, by $\disp_{\mathrm{adam}}$.

The valid correspondence region is defined by intersecting the foreground mask of the fixed role with the warped foreground mask of the moving role:
\begin{equation}
\begin{aligned}
\mask_{\cap}
=
\mask_I
\cap
\mathrm{warp}
\left(
\mask_J,
\disp_{\mathrm{adam}}
\right),
\end{aligned}
\end{equation}
where the warped mask is thresholded after interpolation. 
The \wpls{} statistics are accumulated only over pooled voxels inside $\mask_{\cap}$. 
This avoids using background regions and out-of-field voxels as correspondence pairs.

Although we use MIND+\gica{} in this work, the VoxCor projection formulation does not depend on this specific registration algorithm. 
Any spatial alignment procedure that provides sufficiently accurate fitting-time correspondences could be substituted.

\subsection{Projection-Fitting Implementation Details}
\label{sec:app:projection_details}

\paragraph{Coarse-grid projection fitting:}
The final \wpls{} and \pcatd{} projections are fitted on a coarsened version of the triplanar feature volume rather than on the full-resolution voxel grid. 
We use average pooling with factor $\gridsp=4$, yielding a coarse grid of size $d\times h\times w$ and $R=d h w$ pooled feature vectors per volume, as in Section~\ref{ssec:wpls}. 
This is primarily a memory-budget choice: storing and processing full-resolution $D\times H\times W\times 3\nfeats$ triplanar features across all fit-stage pairs is not practical. 
The coarsened grid also reduces sensitivity to voxel-level noise. 
After fitting, the learned projections are applied densely to the full-resolution triplanar feature volume at transform time.

\paragraph{\wpls{} correspondence region:}
For \wpls{}, the moving-role triplanar feature volume and foreground mask are warped into the fixed-role coordinate frame using the fitting-time displacement field $\disp_{\mathrm{adam}}$. 
The valid correspondence region is the foreground intersection $\mask_{\cap}$ defined above. 
The role-specific means are estimated in a separate first pass from each role's own foreground mask: $\mu_I$ from the fixed-role foreground voxels and $\mu_J$ from the moving-role foreground voxels. 
After this centering step, the feature variances, voxel weights, and weighted cross-covariance are estimated from pooled voxels inside $\mask_{\cap}$.

\paragraph{Voxel weighting:}
The \wpls{} weights are computed from the local multichannel gradient magnitude of the fixed-role triplanar features on the pooled grid. 
The weights are multiplied by the valid foreground correspondence mask, with masked-out voxels contributing zero.
This emphasizes voxels near feature-space transitions, which typically occur around anatomical boundaries and provide more discriminative correspondence anchors than homogeneous foreground regions.

\paragraph{\pcatd{} fitting:}
The \pcatd{} comparison uses the same triplanar features, foreground masks, and coarse-grid pooling as \wpls{}, but does not use fitting-time correspondences. 
It pools foreground features from both roles, estimates one shared mean, and fits one shared PCA projection. 
Unlike \wpls{}, \pcatd{} does not use gradient-magnitude voxel weights; it applies unweighted PCA to the pooled foreground triplanar features. 
The same mean and projection are then applied to all transformed volumes, irrespective of modality role. 
Thus, \pcatd{} isolates the effect of triplanar aggregation and dimensionality reduction without correspondence-aware fitting.

\paragraph{Fit--transform separation:}
The fit phase estimates and stores all projection parameters: the per-axis PCA means and projections
$\{(\mu^a,W^a)\}_{a\in\{S,C,A\}}$ and either the \wpls{} parameters
$(\mu_I,\mu_J,W_I,W_J)$ or the \pcatd{} parameters $(\mu,W)$.
The transform phase extracts frozen ViT features from a new volume, applies the stored per-axis PCA projections densely, concatenates the three anatomical axes, and applies the stored final projection.
No foreground mask, displacement field, registration, or paired volume is required at transform time.

\section{Experimental and Implementation Details}
\label{sec:app:experimental}

This section provides the experimental and implementation details that support Section~\ref{sec:experiments}. We first describe the datasets and segmentation-center landmark construction, then the encoder and feature configuration, the registration hyperparameter search, the VRAM-aware pre-screening, the fixed internal MIND+\gica{} parameters used for \wpls{} fitting, and the software and hardware setup. Method-side implementation details (preprocessing, foreground masks, fitting-time correspondence generation, and projection-fitting specifics) are kept in Appendix~\ref{sec:app:method_details}.

\subsection{Dataset Details}
\label{ssec:app:dataset_details}

\paragraph{Abdomen MR--CT (Intra-Subject).}
We used the training set of the public Learn2Reg Abdomen MR--CT dataset~\cite{hering2022learn2reg}, which consists of eight paired MR--CT volumes.
Each volume was resampled to $192 \times 160 \times 192$ with $2$\,mm isotropic spacing. Because only eight paired cases are available, we evaluated this dataset using Leave-2-Out Cross-Validation (L2OCV): in each fold, two pairs were held out for testing, while the remaining six pairs were used for feature fitting and hyperparameter selection. 
For evaluation, we used the provided manual segmentations of four abdominal organs: \textit{liver, spleen, left kidney, and right kidney}. This dataset is challenging due to pronounced appearance differences between MR and CT, non-rigid anatomical variation, and variability in abdominal field-of-view and organ appearance across cases.

\paragraph{HCP T2w--T1w (Inter-Subject).}
We used T2-weighted (T2w) and T1-weighted (T1w) brain MR volumes from the Human Connectome Project (HCP)~\cite{van2013wu}, each resampled to $256 \times 256 \times 256$ at $0.7$\,mm isotropic spacing.
Each subject provides one T2w and one T1w volume of the same underlying anatomy. For registration, we formed inter-subject cross-modal tasks by pairing different subjects and evaluating both directions, i.e.\ T2w from one subject to T1w from another subject and vice versa. Six subjects were used for feature fitting and hyperparameter selection. The main held-out set contained twelve subjects, from which we constructed the inter-subject registration tasks and the kNN segmentation evaluation blocks. 
For kNN segmentation, the twelve held-out subjects were grouped into six two-subject blocks, each containing both modalities for both subjects. For landmark localization only, we additionally included twelve further held-out HCP subjects, yielding twenty-four held-out subjects.
This expansion was used only for landmark evaluation, because nearest-neighbor matching at segmentation-center landmark locations was less computationally demanding than deformable registration or voxelwise kNN segmentation. The same subject ordering was preserved when constructing evaluation blocks, so the additional subjects extend the original held-out set without changing the evaluation protocol. For evaluation, we used FreeSurfer~\cite{freesurfer} segmentations and derived fourteen anatomical labels covering major brain structures: \textit{cerebellum gray matter, cerebellum white matter, cerebral gray matter, cerebral white matter, thalamus, hippocampus, amygdala, ventricles, caudate, putamen, pallidum, ventral diencephalon, cerebrospinal fluid, and brainstem}.

\subsection{Segmentation-Center Landmark Construction}
\label{ssec:app:landmark_annotations}

Landmark locations were derived automatically from the evaluation segmentations. For each selected anatomical label, we computed the center of mass of the corresponding binary segmentation mask in voxel coordinates and used this point as the landmark location for that volume and modality. If the center of mass is non-integer, the nearest voxel location was used for feature extraction, while Euclidean localization errors were computed in physical millimeters using the dataset spacing. These landmarks were used only for feature-space nearest-neighbor localization and were not used during feature fitting, registration hyperparameter selection, or kNN segmentation.

For Abdomen MR--CT, we used the centers of mass of the four annotated organ labels: \textit{liver, spleen, left kidney, and right kidney}. For HCP T2w--T1w, we used the centers of mass of the fourteen FreeSurfer labels listed in Appendix~\ref{ssec:app:dataset_details}.

\subsection{Dataset- and Encoder-Specific Parameters}
\label{ssec:app:data_variant_params}

Table~\ref{tab:data_variant_params} reports the dataset- and encoder-specific input resizing parameters used for ViT feature extraction, together with the \bandslice{} scale-regularization $\eta$ used for global initialization.
Following DINO-Reg~\cite{song2024dinoreg}, we used a scaling factor $\scale > 1$ to retain high-resolution information for scalable encoders. The remaining scale factors were chosen to yield comparable token-grid resolutions across encoders after accounting for patch size; for example, a DINOv2 scale of $5.3$ with $14{\times}14$ patches corresponds approximately to a DINOv3/MedSAM2 scale of $(5.3/14)\times16 \approx 6.0$ with $16{\times}16$ patches. Meanwhile, SAM3 used its native fixed-resolution input. Empirically, $\eta$ should be set close to $1$ when the fixed and moving images come from the same individual (Abdomen MR--CT), where $\eta{=}1$ effectively restricts \bandslice{} to translation without scaling, and it must be lower when they come from different individuals (HCP T2w--T1w).

\begin{table}
\centering
\caption{Dataset- and encoder-specific input resizing parameters and \bandslice{} scale regularization.}
\label{tab:data_variant_params}
\vspace{4pt}
\begin{tabular}{l|c||c|c|c}
\hline
Dataset & Model & Input / Scale & Patch & $\eta$ \\
\hline\hline
\multirow{4}{*}{\shortstack[l]{Abdomen\\MR--CT}}
 & DINOv2   & $\scale = 5.3$ & $14^2$ & 0.99 \\
 & DINOv3   & $\scale = 6.0$ & $16^2$ & 0.99 \\
 & MedSAM2  & $\scale = 6.0$ & $16^2$ & 0.99 \\
 & SAM3     & $1008^2$       & $14^2$ & 0.99 \\
\hline
\multirow{4}{*}{\shortstack[l]{HCP\\T2w--T1w}}
 & DINOv2   & $\scale = 4.0$ & $14^2$ & 0.1 \\
 & DINOv3   & $\scale = 4.0$ & $16^2$ & 0.1 \\
 & MedSAM2  & $\scale = 4.0$ & $16^2$ & 0.1 \\
 & SAM3     & $1008^2$       & $14^2$ & 0.1 \\
\hline
\end{tabular}
\end{table}

\subsection{Feature Dimensionality and Normalization}
\label{ssec:app:feature_norm}

Following DINO-Reg~\cite{song2024dinoreg}, we used 24 PCA output channels per anatomical axis and slice subsampling with stride 3. The final projected dimensionality was set equal to the per-axis PCA dimensionality,
\begin{equation}
\nprojfeats = \nfeats = 24,
\end{equation}
so that Axial PCA, \pcatd{}, and \wpls{} each produced 24-channel ViT representations.

\paragraph{Per-voxel L2 normalization.}
For all ViT-based registration features, we applied per-voxel $L_2$ normalization of the channel vector before ConvexAdam. Since this normalization changed the feature scale, the ConvexAdam regularization range for ViT-based methods was scaled by $0.1$ relative to the unnormalized setting (Appendix~\ref{ssec:app:hps}, Table~\ref{tab:hyperparams}).

\paragraph{+MIND hybrid construction.}
For ViT+MIND hybrids, we concatenated the first 16 projected ViT channels, scaled by $0.1$, with 12-channel MIND descriptors, yielding a 28-channel feature representation. The MIND descriptors used inside these +MIND registration hybrids used radius~$1$ and dilation~$2$, following the Anatomix+MIND comparison protocol; these parameters differ from the radius~$2$, dilation~$2$ MIND descriptors used internally for fitting-time correspondence generation (Appendix~\ref{ssec:app:internal_mind_gica}). Anatomix+MIND was constructed analogously by scaling Anatomix features by $0.1$ before concatenating them with the MIND descriptor.
MIND alone was used in its standard unnormalized form.

\subsection{Registration Hyperparameter Search}
\label{ssec:app:hps}

All deformable registration experiments used ConvexAdam~\cite{siebert2024convexadam}. The hyperparameter search space contained five structural parameters,
\[
\lambda,\quad \dispHW,\quad \gridsp,\quad \gridsp_{\mathrm{adam}},\quad \sigma_{\mathrm{gauss}},
\]
and two refinement parameters,
\[
n_{\mathrm{iter}},\quad n_{\mathrm{smooth}}.
\]
The full search space is shown in Table~\ref{tab:hyperparams}. In the L2-normalized ViT setting, the regularization range for $\lambda$ was scaled by $0.1$ as described in Appendix~\ref{ssec:app:feature_norm} above.

\begin{table}
\centering
\caption{ConvexAdam hyperparameter search space. In the \textbf{$L_2$ norm} setting used for ViT-based registration features, regularization weights are scaled by $0.1$.}
\label{tab:hyperparams}
\vspace{4pt}
\begin{tabular}{l|c|c}
\hline
\textbf{Parameter} & \textbf{$L_2$ norm} & \textbf{No norm} \\
\hline
$\lambda$ & $0.1 \!\times\! \{0.4, 0.6, \ldots, 1.6\}$ &
  $\{0.4, 0.6, \ldots, 1.6\}$ \\
$\dispHW$ & \multicolumn{2}{c}{$\{2, 3, 4, 5, 6, 7\}$} \\
$\gridsp$ & \multicolumn{2}{c}{$\{2, 3, 4, 5\}$} \\
$\gridsp_{\mathrm{adam}}$ & \multicolumn{2}{c}{$\{1, 2, 3, 4\}$} \\
$\sigma_{\mathrm{gauss}}$ &
  \multicolumn{2}{c}{$\{0.7, 1.0, 1.3, 1.6, 1.9, 2.2, 2.5, 2.8\}$} \\
\hline
$n_{\mathrm{iter}}$ & \multicolumn{2}{c}{$\{60, 80, 100, 120\}$} \\
$n_{\mathrm{smooth}}$ & \multicolumn{2}{c}{$\{0, 1, 2, 3\}$} \\
\hline
\end{tabular}
\end{table}

\paragraph{Sampling protocol.}
Not all structural parameter combinations were feasible within GPU memory. We therefore first applied the VRAM-aware screening described in Appendix~\ref{ssec:app:vram}. After excluding infeasible triples, we sampled $N = 400$ structural configurations uniformly at random from the remaining search space using seed~42. For each sampled structural configuration, refinement parameters were evaluated exhaustively over the $4 \times 4$ grid
\[
n_{\mathrm{iter}} \in \{60,80,100,120\},
\qquad
n_{\mathrm{smooth}} \in \{0,1,2,3\}.
\]
This yielded $M = 16$ refinement variants per structural configuration and $6{,}400$ evaluated configurations per method setting. In addition to the $6{,}400$ ConvexAdam configurations described above, the search included the convex-only variant and the global-initialization-only variant (no convex stage, no Adam stage) as additional candidates, allowing HPS to select a coarser or fully affine output when refinement did not improve validation Dice. Higher iteration counts and additional smoothing passes reused intermediate optimizer states from a single run, avoiding redundant computation.

For Abdomen MR--CT, the best configuration was selected by L2OCV Dice within each fold, using the fold-specific fitting/training split. For HCP T2w--T1w, hyperparameters were selected by the highest Dice on the training split used jointly for feature fitting and registration hyperparameter search, and the selected configuration was then applied to the held-out inter-subject cross-modal registration tasks.

\paragraph{ConvexAdam-MIND search space.}
For ConvexAdam-MIND, the search space additionally included MIND-specific parameters for the convex and Adam stages separately:
\[
r_c \in \{1,2,3\}, \qquad d_c \in \{1,2,3\},
\]
\[
r_a \in \{1,2\}, \qquad d_a \in \{1,2\},
\]
where $r_c$ and $d_c$ are the MIND radius and dilation in the convex stage, and $r_a$ and $d_a$ are the corresponding parameters in the Adam stage. The same $N = 400$ structural sampling and $M = 16$ refinement protocol was used, based on the 12-channel MIND VRAM feasibility screen.

\subsection{VRAM-Aware Hyperparameter Pre-Screening}
\label{ssec:app:vram}

We pre-screened every $(\gridsp, \gridsp_{\mathrm{adam}}, \dispHW)$ triple for memory feasibility on an NVIDIA A6000 GPU with 48\,GB memory. Screening was performed separately for each dataset resolution and target channel count. ViT-based methods were screened in the 28-channel setting corresponding to the largest registration feature configuration, namely \wpls{}+MIND. ConvexAdam-MIND was screened separately in its native 12-channel setting. Table~\ref{tab:avoid_params} reports the number of infeasible triples excluded from the 96 possible structural triples. This screening was used only to remove configurations that could not be executed reliably under the available memory budget. It did not otherwise rank or tune hyperparameters.

\begin{table}
\centering
\caption{VRAM pre-screening: number of infeasible $(\gridsp,\gridsp_{\mathrm{adam}},\dispHW)$ triples excluded from the search out of 96 total triples for each dataset and channel count on an NVIDIA A6000 GPU with 48\,GB memory.}
\label{tab:avoid_params}
\vspace{4pt}
\begin{tabular}{l|c|c}
\hline
\textbf{Setting} & \textbf{Abdomen MR--CT} & \textbf{HCP T2w--T1w} \\
 & $(192{\times}160{\times}192)$ & $(256^3)$ \\
\hline
28 ch (ViT-based) & 8 / 96 & 28 / 96 \\
12 ch (MIND) & 4 / 96 & 12 / 96 \\
\hline
\end{tabular}
\end{table}

\subsection{Fixed Internal MIND+\gica{} for \wpls{} Fitting}
\label{ssec:app:internal_mind_gica}

The MIND+\gica{} procedure used to generate fitting-time correspondences for \wpls{} operates with fixed parameters and was not part of the registration hyperparameter search. The parameters were chosen to lie approximately in the middle of the standard ConvexAdam-MIND search space, rather than being tuned for a specific dataset or encoder. In particular, we used the central MIND setting $r{=}2$, $d{=}2$ and intermediate ConvexAdam settings for the displacement search radius, grid spacings, regularization, iteration count, smoothing, and Gaussian kernel width. This provides a stable correspondence generator for \wpls{} fitting while keeping the fitting-time registration procedure separate from the downstream registration hyperparameter search. Table~\ref{tab:internal_mind_params} reports these fixed settings. The resulting displacement fields were used only for accumulating the \wpls{} cross-covariance during offline fitting; they were not estimated for transform-time volumes and were not used by \pcatd{}.

\begin{table}
\centering
\caption{Fixed parameters for the internal MIND+\gica{} procedure used during \wpls{} fitting. These parameters are not part of the registration hyperparameter search.}
\label{tab:internal_mind_params}
\vspace{4pt}
\begin{tabular}{l|c}
\hline
\textbf{Component} & \textbf{Parameters} \\
\hline
MIND & radius\,=\,2, dilation\,=\,2, mask-aware \\
\hline
ConvexAdam & $\lambda = 1.0$, $\dispHW = 5$, $\gridsp = 3$,\\
           & $\gridsp_{\mathrm{adam}} = 2$, $n_{\mathrm{iter}} = 80$ \\
           & $n_{\mathrm{smooth}} = 1$, $\sigma_{\mathrm{gauss}} = 1.6$ \\
\hline
\bandslice{} global & Dataset-dependent; see Table~\ref{tab:data_variant_params} \\
\hline
\end{tabular}
\end{table}

\subsection{Software and Hardware}
\label{ssec:app:software_hardware}

All experiments were implemented in PyTorch. ViT encoders were executed in bfloat16 precision, with xFormers~\cite{xFormers2022} memory-efficient attention enabled where supported. Most experiments were run on a single NVIDIA A6000 GPU with 48\,GB memory. A small number of large registration configurations required an NVIDIA A100 GPU with 80\,GB memory. Evaluation scripts supported checkpointed execution for SLURM-based workflows, allowing interrupted hyperparameter search and test evaluations to resume without repeating completed configurations.

\section{Complete Registration Results}
\label{sec:app:reg_all}

Tables~\ref{tab:registration_full_abdmrct} and~\ref{tab:registration_full_hcpt2t1} provide the complete numerical deformable registration results underlying the headline summaries in Tables~\ref{tab:reg_main_base} and~\ref{tab:reg_main_mind}. Results are separated by dataset. CNN baselines are reported as fixed feature configurations without a fitting regime: MIND, Anatomix, and Anatomix+MIND. For ViT-based methods, Axial, \pcatd{}, and \wpls{} features are reported with and without appended MIND features, under both Dataset-Fit and Pair-Fit fitting regimes. Each setting is evaluated with direct ConvexAdam (CA) and Globally-Initialized ConvexAdam (\gica{}). The three reported metrics are Dice, HD95 in millimeters, and sdLogJ. Higher Dice is better, whereas lower HD95 and sdLogJ are better. For readability, the main text separates the headline Pair-Fit+\gica{} results into base and MIND-augmented tables, while these appendix tables expose the full registration grid used for the analysis.

\begin{table}
\centering
\caption{Complete numerical deformable registration results on Abdomen MR--CT. CNN baselines are reported as fixed feature configurations without a fitting regime. ViT-based methods are reported using Axial, \pcatd{}, and \wpls{} features, with and without appended MIND features, under both Dataset-Fit and Pair-Fit settings. ConvexAdam denotes direct deformable registration, and Globally-Initialized ConvexAdam denotes registration with global initialization. Values are reported as mean$\pm$standard deviation. Higher Dice is better, whereas lower HD95 and sdLogJ are better. Bold marks the best mean Dice and HD95 values in each column; for sdLogJ, lower values indicate smoother fields but should be interpreted jointly with Dice and HD95.}
\label{tab:registration_full_abdmrct}
\vspace{4pt}
\footnotesize
\setlength{\tabcolsep}{2.0pt}
\resizebox{\textwidth}{!}{%
\begin{tabular}{l|l|l||ccc|ccc}
\hline
\multirow{2}{*}{\textbf{Encoder}} & \multirow{2}{*}{\textbf{Method}} & \multirow{2}{*}{\textbf{Setting}} & \multicolumn{3}{c|}{\textbf{ConvexAdam}} & \multicolumn{3}{c}{\textbf{Globally-Initialized ConvexAdam}} \\
 &  &  & Dice $\uparrow$ & HD95 [mm] $\downarrow$ & sdLogJ $\downarrow$ & Dice $\uparrow$ & HD95 [mm] $\downarrow$ & sdLogJ $\downarrow$ \\
\hline\hline
\multirow{3}{*}{--} & MIND & - & $0.778{\pm}0.157$ & $17.672{\pm}14.599$ & $0.180{\pm}0.028$ & $0.839{\pm}0.073$ & $13.874{\pm}10.840$ & $0.163{\pm}0.030$ \\
 & Anatomix & - & $0.727{\pm}0.221$ & $18.525{\pm}16.342$ & $0.121{\pm}0.019$ & $0.803{\pm}0.119$ & $14.036{\pm}\phantom{0}8.377$ & $0.119{\pm}0.015$ \\
 & Anatomix+MIND & - & $0.771{\pm}0.206$ & $19.448{\pm}16.513$ & $0.168{\pm}0.020$ & $\mathbf{0.868{\pm}0.059}$ & $\phantom{0}9.556{\pm}\phantom{0}5.655$ & $0.155{\pm}0.018$ \\
\hline
\multirow{12}{*}{DINOv2} & Axial & Dataset-Fit & $0.800{\pm}0.109$ & $17.604{\pm}13.538$ & $0.173{\pm}0.031$ & $0.837{\pm}0.061$ & $12.069{\pm}\phantom{0}7.671$ & $0.150{\pm}0.025$ \\
 & Axial & Pair-Fit & $0.792{\pm}0.124$ & $18.099{\pm}14.212$ & $0.243{\pm}0.090$ & $0.841{\pm}0.058$ & $11.471{\pm}\phantom{0}6.727$ & $0.154{\pm}0.025$ \\
 & Axial+MIND & Dataset-Fit & $0.810{\pm}0.116$ & $17.683{\pm}14.084$ & $0.192{\pm}0.025$ & $0.854{\pm}0.055$ & $10.493{\pm}\phantom{0}6.117$ & $0.146{\pm}0.022$ \\
 & Axial+MIND & Pair-Fit & $0.809{\pm}0.115$ & $17.516{\pm}14.510$ & $0.206{\pm}0.039$ & $0.858{\pm}0.055$ & $\phantom{0}9.965{\pm}\phantom{0}5.373$ & $0.134{\pm}0.017$ \\
 & \pcatd{} & Dataset-Fit & $0.714{\pm}0.221$ & $24.333{\pm}20.494$ & $0.338{\pm}0.158$ & $0.689{\pm}0.249$ & $26.012{\pm}23.464$ & $0.153{\pm}0.026$ \\
 & \pcatd{} & Pair-Fit & $0.718{\pm}0.198$ & $23.167{\pm}18.094$ & $0.305{\pm}0.115$ & $0.703{\pm}0.230$ & $22.712{\pm}19.323$ & $0.164{\pm}0.027$ \\
 & \pcatd{}+MIND & Dataset-Fit & $0.713{\pm}0.228$ & $24.708{\pm}21.200$ & $0.229{\pm}0.037$ & $0.686{\pm}0.259$ & $26.548{\pm}24.032$ & $0.167{\pm}0.039$ \\
 & \pcatd{}+MIND & Pair-Fit & $0.713{\pm}0.218$ & $24.170{\pm}19.815$ & $0.235{\pm}0.061$ & $0.684{\pm}0.242$ & $25.904{\pm}21.856$ & $0.146{\pm}0.023$ \\
 & \wpls{} & Dataset-Fit & $0.793{\pm}0.139$ & $16.590{\pm}13.667$ & $0.245{\pm}0.060$ & $0.776{\pm}0.154$ & $17.633{\pm}13.395$ & $0.159{\pm}0.022$ \\
 & \wpls{} & Pair-Fit & $0.845{\pm}0.054$ & $\mathbf{11.477{\pm}\phantom{0}8.478}$ & $0.205{\pm}0.024$ & $0.839{\pm}0.060$ & $11.454{\pm}\phantom{0}7.529$ & $0.150{\pm}0.020$ \\
 & \wpls{}+MIND & Dataset-Fit & $0.786{\pm}0.172$ & $17.387{\pm}15.174$ & $0.191{\pm}0.029$ & $0.761{\pm}0.193$ & $18.503{\pm}15.576$ & $0.141{\pm}0.020$ \\
 & \wpls{}+MIND & Pair-Fit & $\mathbf{0.847{\pm}0.064}$ & $11.618{\pm}10.125$ & $0.185{\pm}0.030$ & $0.844{\pm}0.061$ & $11.004{\pm}\phantom{0}6.126$ & $0.139{\pm}0.016$ \\
\hline
\multirow{12}{*}{DINOv3} & Axial & Dataset-Fit & $0.788{\pm}0.110$ & $18.383{\pm}11.882$ & $0.211{\pm}0.065$ & $0.784{\pm}0.129$ & $16.871{\pm}12.470$ & $0.151{\pm}0.032$ \\
 & Axial & Pair-Fit & $0.786{\pm}0.103$ & $18.434{\pm}11.529$ & $0.205{\pm}0.047$ & $0.781{\pm}0.128$ & $16.798{\pm}11.572$ & $0.150{\pm}0.036$ \\
 & Axial+MIND & Dataset-Fit & $0.773{\pm}0.175$ & $16.801{\pm}14.987$ & $0.129{\pm}0.026$ & $0.853{\pm}0.066$ & $10.305{\pm}\phantom{0}5.684$ & $0.104{\pm}0.025$ \\
 & Axial+MIND & Pair-Fit & $0.777{\pm}0.169$ & $16.456{\pm}14.582$ & $0.130{\pm}0.027$ & $0.853{\pm}0.066$ & $10.146{\pm}\phantom{0}5.721$ & $0.100{\pm}0.020$ \\
 & \pcatd{} & Dataset-Fit & $0.700{\pm}0.214$ & $23.999{\pm}16.043$ & $0.186{\pm}0.024$ & $0.656{\pm}0.267$ & $26.486{\pm}24.509$ & $0.128{\pm}0.029$ \\
 & \pcatd{} & Pair-Fit & $0.710{\pm}0.193$ & $22.732{\pm}16.504$ & $0.200{\pm}0.040$ & $0.662{\pm}0.248$ & $26.160{\pm}24.899$ & $0.127{\pm}0.031$ \\
 & \pcatd{}+MIND & Dataset-Fit & $0.759{\pm}0.217$ & $18.374{\pm}17.595$ & $0.145{\pm}0.020$ & $0.727{\pm}0.230$ & $20.993{\pm}21.238$ & $0.116{\pm}0.035$ \\
 & \pcatd{}+MIND & Pair-Fit & $0.763{\pm}0.190$ & $19.176{\pm}17.527$ & $0.145{\pm}0.021$ & $0.739{\pm}0.192$ & $20.115{\pm}18.391$ & $0.106{\pm}0.032$ \\
 & \wpls{} & Dataset-Fit & $0.753{\pm}0.172$ & $20.736{\pm}15.429$ & $0.195{\pm}0.039$ & $0.756{\pm}0.192$ & $19.096{\pm}15.859$ & $0.193{\pm}0.091$ \\
 & \wpls{} & Pair-Fit & $0.811{\pm}0.097$ & $14.484{\pm}\phantom{0}9.395$ & $0.167{\pm}0.029$ & $0.826{\pm}0.073$ & $12.091{\pm}\phantom{0}5.477$ & $0.138{\pm}0.014$ \\
 & \wpls{}+MIND & Dataset-Fit & $0.781{\pm}0.184$ & $17.129{\pm}15.480$ & $0.145{\pm}0.019$ & $0.805{\pm}0.140$ & $14.326{\pm}12.884$ & $0.099{\pm}0.021$ \\
 & \wpls{}+MIND & Pair-Fit & $0.827{\pm}0.088$ & $13.613{\pm}10.816$ & $0.142{\pm}0.018$ & $0.863{\pm}0.056$ & $\phantom{0}9.463{\pm}\phantom{0}4.739$ & $0.099{\pm}0.022$ \\
\hline
\multirow{12}{*}{MedSAM2} & Axial & Dataset-Fit & $0.778{\pm}0.141$ & $19.081{\pm}13.346$ & $0.226{\pm}0.055$ & $0.833{\pm}0.101$ & $12.726{\pm}\phantom{0}7.352$ & $0.154{\pm}0.032$ \\
 & Axial & Pair-Fit & $0.789{\pm}0.131$ & $17.363{\pm}11.530$ & $0.207{\pm}0.037$ & $0.835{\pm}0.097$ & $12.401{\pm}\phantom{0}6.954$ & $0.149{\pm}0.033$ \\
 & Axial+MIND & Dataset-Fit & $0.752{\pm}0.215$ & $18.123{\pm}17.085$ & $0.120{\pm}0.026$ & $0.831{\pm}0.091$ & $12.192{\pm}\phantom{0}6.514$ & $0.089{\pm}0.016$ \\
 & Axial+MIND & Pair-Fit & $0.753{\pm}0.214$ & $17.985{\pm}17.179$ & $0.122{\pm}0.029$ & $0.810{\pm}0.102$ & $13.559{\pm}\phantom{0}6.649$ & $0.075{\pm}0.019$ \\
 & \pcatd{} & Dataset-Fit & $0.685{\pm}0.244$ & $25.191{\pm}20.800$ & $0.179{\pm}0.040$ & $0.807{\pm}0.127$ & $15.307{\pm}\phantom{0}9.289$ & $0.169{\pm}0.024$ \\
 & \pcatd{} & Pair-Fit & $0.726{\pm}0.194$ & $22.232{\pm}17.892$ & $0.188{\pm}0.041$ & $0.822{\pm}0.109$ & $14.878{\pm}\phantom{0}9.038$ & $0.206{\pm}0.078$ \\
 & \pcatd{}+MIND & Dataset-Fit & $0.758{\pm}0.204$ & $17.399{\pm}16.143$ & $0.122{\pm}0.029$ & $0.824{\pm}0.086$ & $12.794{\pm}\phantom{0}6.286$ & $0.081{\pm}0.021$ \\
 & \pcatd{}+MIND & Pair-Fit & $0.759{\pm}0.203$ & $17.325{\pm}16.161$ & $0.121{\pm}0.026$ & $0.818{\pm}0.108$ & $12.799{\pm}\phantom{0}6.802$ & $0.077{\pm}0.025$ \\
 & \wpls{} & Dataset-Fit & $0.798{\pm}0.144$ & $17.323{\pm}13.671$ & $0.185{\pm}0.036$ & $0.829{\pm}0.115$ & $12.010{\pm}\phantom{0}7.499$ & $0.149{\pm}0.019$ \\
 & \wpls{} & Pair-Fit & $0.814{\pm}0.131$ & $13.893{\pm}11.461$ & $0.161{\pm}0.030$ & $0.824{\pm}0.132$ & $10.967{\pm}\phantom{0}7.465$ & $0.142{\pm}0.027$ \\
 & \wpls{}+MIND & Dataset-Fit & $0.757{\pm}0.204$ & $17.252{\pm}16.226$ & $0.120{\pm}0.027$ & $0.824{\pm}0.085$ & $13.167{\pm}\phantom{0}6.354$ & $0.086{\pm}0.014$ \\
 & \wpls{}+MIND & Pair-Fit & $0.760{\pm}0.206$ & $17.016{\pm}16.201$ & $0.122{\pm}0.028$ & $0.829{\pm}0.085$ & $12.413{\pm}\phantom{0}6.297$ & $0.090{\pm}0.017$ \\
\hline
\multirow{12}{*}{SAM3} & Axial & Dataset-Fit & $0.689{\pm}0.178$ & $25.923{\pm}13.809$ & $0.195{\pm}0.032$ & $0.773{\pm}0.114$ & $18.123{\pm}\phantom{0}8.890$ & $0.154{\pm}0.019$ \\
 & Axial & Pair-Fit & $0.692{\pm}0.172$ & $25.004{\pm}12.863$ & $0.195{\pm}0.035$ & $0.778{\pm}0.110$ & $17.223{\pm}\phantom{0}7.605$ & $0.141{\pm}0.012$ \\
 & Axial+MIND & Dataset-Fit & $0.768{\pm}0.186$ & $16.922{\pm}15.081$ & $0.125{\pm}0.029$ & $0.810{\pm}0.151$ & $13.261{\pm}\phantom{0}9.765$ & $0.082{\pm}0.022$ \\
 & Axial+MIND & Pair-Fit & $0.771{\pm}0.185$ & $16.676{\pm}15.101$ & $0.124{\pm}0.030$ & $0.813{\pm}0.151$ & $12.686{\pm}\phantom{0}9.633$ & $0.085{\pm}0.027$ \\
 & \pcatd{} & Dataset-Fit & $0.605{\pm}0.246$ & $33.709{\pm}22.309$ & $0.206{\pm}0.057$ & $0.616{\pm}0.245$ & $31.720{\pm}22.504$ & $0.169{\pm}0.027$ \\
 & \pcatd{} & Pair-Fit & $0.617{\pm}0.236$ & $32.929{\pm}22.123$ & $0.225{\pm}0.074$ & $0.610{\pm}0.255$ & $30.934{\pm}22.572$ & $0.163{\pm}0.029$ \\
 & \pcatd{}+MIND & Dataset-Fit & $0.747{\pm}0.201$ & $19.866{\pm}18.772$ & $0.131{\pm}0.026$ & $0.680{\pm}0.287$ & $28.420{\pm}35.512$ & $0.101{\pm}0.025$ \\
 & \pcatd{}+MIND & Pair-Fit & $0.749{\pm}0.198$ & $19.383{\pm}18.729$ & $0.131{\pm}0.027$ & $0.706{\pm}0.281$ & $26.974{\pm}36.991$ & $0.100{\pm}0.030$ \\
 & \wpls{} & Dataset-Fit & $0.700{\pm}0.210$ & $24.670{\pm}16.425$ & $0.243{\pm}0.037$ & $0.701{\pm}0.206$ & $24.482{\pm}16.391$ & $0.184{\pm}0.025$ \\
 & \wpls{} & Pair-Fit & $0.776{\pm}0.134$ & $17.182{\pm}11.686$ & $0.165{\pm}0.020$ & $0.794{\pm}0.132$ & $13.212{\pm}\phantom{0}6.245$ & $0.145{\pm}0.023$ \\
 & \wpls{}+MIND & Dataset-Fit & $0.778{\pm}0.201$ & $17.124{\pm}16.726$ & $0.148{\pm}0.018$ & $0.816{\pm}0.099$ & $12.690{\pm}\phantom{0}7.315$ & $0.079{\pm}0.019$ \\
 & \wpls{}+MIND & Pair-Fit & $0.808{\pm}0.119$ & $14.188{\pm}12.358$ & $0.131{\pm}0.029$ & $0.860{\pm}0.061$ & $\mathbf{\phantom{0}9.307{\pm}\phantom{0}4.437}$ & $0.103{\pm}0.032$ \\
\hline
\end{tabular}
}
\end{table}

\begin{table}
\centering
\caption{Complete numerical deformable registration results on HCP T2w--T1w. CNN baselines are reported as fixed feature configurations without a fitting regime. ViT-based methods are reported using Axial, \pcatd{}, and \wpls{} features, with and without appended MIND features, under both Dataset-Fit and Pair-Fit settings. ConvexAdam denotes direct deformable registration, and Globally-Initialized ConvexAdam denotes registration with global initialization. Values are reported as mean$\pm$standard deviation. Higher Dice is better, whereas lower HD95 and sdLogJ are better. Bold marks the best mean Dice and HD95 values in each column; for sdLogJ, lower values indicate smoother fields but should be interpreted jointly with Dice and HD95. Some entries report sdLogJ = 0.000; in these cases the selected hyperparameter configuration's refinement stage did not improve over the BandSlice global initialization, so the final displacement reduces to an affine transform and the log-Jacobian determinant is constant.}
\label{tab:registration_full_hcpt2t1}
\vspace{4pt}
\footnotesize
\setlength{\tabcolsep}{2.0pt}
\resizebox{\textwidth}{!}{%
\begin{tabular}{l|l|l||ccc|ccc}
\hline
\multirow{2}{*}{\textbf{Encoder}} & \multirow{2}{*}{\textbf{Method}} & \multirow{2}{*}{\textbf{Setting}} & \multicolumn{3}{c|}{\textbf{ConvexAdam}} & \multicolumn{3}{c}{\textbf{Globally-Initialized ConvexAdam}} \\
 &  &  & Dice $\uparrow$ & HD95 [mm] $\downarrow$ & sdLogJ $\downarrow$ & Dice $\uparrow$ & HD95 [mm] $\downarrow$ & sdLogJ $\downarrow$ \\
\hline\hline
\multirow{3}{*}{--} & MIND & - & $0.791{\pm}0.011$ & $\phantom{0}1.980{\pm}\phantom{0}0.228$ & $0.085{\pm}0.014$ & $0.794{\pm}0.011$ & $\phantom{0}1.934{\pm}\phantom{0}0.202$ & $0.067{\pm}0.007$ \\
 & Anatomix & - & $0.737{\pm}0.022$ & $\phantom{0}2.358{\pm}\phantom{0}0.328$ & $0.050{\pm}0.012$ & $0.736{\pm}0.018$ & $\phantom{0}2.345{\pm}\phantom{0}0.271$ & $0.046{\pm}0.008$ \\
 & Anatomix+MIND & - & $0.794{\pm}0.011$ & $\phantom{0}1.937{\pm}\phantom{0}0.211$ & $0.076{\pm}0.007$ & $0.794{\pm}0.011$ & $\phantom{0}1.933{\pm}\phantom{0}0.219$ & $0.071{\pm}0.009$ \\
\hline
\multirow{12}{*}{DINOv2} & Axial & Dataset-Fit & $0.744{\pm}0.013$ & $\phantom{0}2.273{\pm}\phantom{0}0.205$ & $0.058{\pm}0.007$ & $0.743{\pm}0.014$ & $\phantom{0}2.267{\pm}\phantom{0}0.215$ & $0.058{\pm}0.007$ \\
 & Axial & Pair-Fit & $0.744{\pm}0.013$ & $\phantom{0}2.270{\pm}\phantom{0}0.220$ & $0.058{\pm}0.007$ & $0.742{\pm}0.014$ & $\phantom{0}2.278{\pm}\phantom{0}0.238$ & $0.058{\pm}0.007$ \\
 & Axial+MIND & Dataset-Fit & $0.776{\pm}0.012$ & $\phantom{0}2.079{\pm}\phantom{0}0.227$ & $0.072{\pm}0.006$ & $0.774{\pm}0.012$ & $\phantom{0}2.071{\pm}\phantom{0}0.228$ & $0.055{\pm}0.008$ \\
 & Axial+MIND & Pair-Fit & $0.774{\pm}0.012$ & $\phantom{0}2.088{\pm}\phantom{0}0.231$ & $0.072{\pm}0.006$ & $0.771{\pm}0.013$ & $\phantom{0}2.101{\pm}\phantom{0}0.236$ & $0.048{\pm}0.007$ \\
 & \pcatd{} & Dataset-Fit & $0.752{\pm}0.013$ & $\phantom{0}2.177{\pm}\phantom{0}0.259$ & $0.055{\pm}0.008$ & $0.750{\pm}0.013$ & $\phantom{0}2.169{\pm}\phantom{0}0.239$ & $0.056{\pm}0.009$ \\
 & \pcatd{} & Pair-Fit & $0.750{\pm}0.014$ & $\phantom{0}2.196{\pm}\phantom{0}0.265$ & $0.055{\pm}0.008$ & $0.749{\pm}0.014$ & $\phantom{0}2.188{\pm}\phantom{0}0.255$ & $0.055{\pm}0.008$ \\
 & \pcatd{}+MIND & Dataset-Fit & $0.767{\pm}0.013$ & $\phantom{0}2.103{\pm}\phantom{0}0.268$ & $0.063{\pm}0.007$ & $0.765{\pm}0.014$ & $\phantom{0}2.100{\pm}\phantom{0}0.265$ & $0.059{\pm}0.011$ \\
 & \pcatd{}+MIND & Pair-Fit & $0.766{\pm}0.013$ & $\phantom{0}2.113{\pm}\phantom{0}0.247$ & $0.057{\pm}0.007$ & $0.764{\pm}0.014$ & $\phantom{0}2.102{\pm}\phantom{0}0.245$ & $0.052{\pm}0.008$ \\
 & \wpls{} & Dataset-Fit & $0.787{\pm}0.015$ & $\phantom{0}1.971{\pm}\phantom{0}0.244$ & $0.092{\pm}0.007$ & $0.785{\pm}0.014$ & $\phantom{0}1.959{\pm}\phantom{0}0.227$ & $0.063{\pm}0.008$ \\
 & \wpls{} & Pair-Fit & $0.787{\pm}0.013$ & $\phantom{0}2.003{\pm}\phantom{0}0.252$ & $0.092{\pm}0.009$ & $0.784{\pm}0.014$ & $\phantom{0}1.970{\pm}\phantom{0}0.222$ & $0.061{\pm}0.007$ \\
 & \wpls{}+MIND & Dataset-Fit & $0.799{\pm}0.013$ & $\mathbf{\phantom{0}1.870{\pm}\phantom{0}0.218}$ & $0.087{\pm}0.006$ & $0.797{\pm}0.013$ & $\mathbf{\phantom{0}1.883{\pm}\phantom{0}0.208}$ & $0.075{\pm}0.010$ \\
 & \wpls{}+MIND & Pair-Fit & $0.797{\pm}0.012$ & $\phantom{0}1.908{\pm}\phantom{0}0.221$ & $0.090{\pm}0.008$ & $0.794{\pm}0.012$ & $\phantom{0}1.928{\pm}\phantom{0}0.212$ & $0.062{\pm}0.008$ \\
\hline
\multirow{12}{*}{DINOv3} & Axial & Dataset-Fit & $0.726{\pm}0.012$ & $\phantom{0}2.427{\pm}\phantom{0}0.257$ & $0.049{\pm}0.007$ & $0.724{\pm}0.013$ & $\phantom{0}2.445{\pm}\phantom{0}0.243$ & $0.051{\pm}0.006$ \\
 & Axial & Pair-Fit & $0.725{\pm}0.012$ & $\phantom{0}2.445{\pm}\phantom{0}0.271$ & $0.050{\pm}0.007$ & $0.722{\pm}0.016$ & $\phantom{0}2.477{\pm}\phantom{0}0.267$ & $0.051{\pm}0.006$ \\
 & Axial+MIND & Dataset-Fit & $0.790{\pm}0.012$ & $\phantom{0}1.979{\pm}\phantom{0}0.240$ & $0.073{\pm}0.005$ & $0.785{\pm}0.013$ & $\phantom{0}1.987{\pm}\phantom{0}0.231$ & $0.039{\pm}0.004$ \\
 & Axial+MIND & Pair-Fit & $0.791{\pm}0.012$ & $\phantom{0}1.981{\pm}\phantom{0}0.232$ & $0.075{\pm}0.005$ & $0.789{\pm}0.012$ & $\phantom{0}1.955{\pm}\phantom{0}0.226$ & $0.047{\pm}0.005$ \\
 & \pcatd{} & Dataset-Fit & $0.732{\pm}0.015$ & $\phantom{0}2.371{\pm}\phantom{0}0.257$ & $0.044{\pm}0.008$ & $0.730{\pm}0.019$ & $\phantom{0}2.391{\pm}\phantom{0}0.276$ & $0.048{\pm}0.007$ \\
 & \pcatd{} & Pair-Fit & $0.735{\pm}0.019$ & $\phantom{0}2.352{\pm}\phantom{0}0.280$ & $0.045{\pm}0.008$ & $0.731{\pm}0.025$ & $\phantom{0}2.391{\pm}\phantom{0}0.322$ & $0.049{\pm}0.007$ \\
 & \pcatd{}+MIND & Dataset-Fit & $0.791{\pm}0.012$ & $\phantom{0}1.984{\pm}\phantom{0}0.238$ & $0.073{\pm}0.005$ & $0.786{\pm}0.013$ & $\phantom{0}1.976{\pm}\phantom{0}0.221$ & $0.038{\pm}0.005$ \\
 & \pcatd{}+MIND & Pair-Fit & $0.792{\pm}0.012$ & $\phantom{0}1.976{\pm}\phantom{0}0.228$ & $0.072{\pm}0.005$ & $0.787{\pm}0.013$ & $\phantom{0}1.967{\pm}\phantom{0}0.238$ & $0.044{\pm}0.005$ \\
 & \wpls{} & Dataset-Fit & $0.777{\pm}0.016$ & $\phantom{0}2.101{\pm}\phantom{0}0.273$ & $0.079{\pm}0.007$ & $0.773{\pm}0.016$ & $\phantom{0}2.083{\pm}\phantom{0}0.259$ & $0.052{\pm}0.007$ \\
 & \wpls{} & Pair-Fit & $0.777{\pm}0.013$ & $\phantom{0}2.069{\pm}\phantom{0}0.210$ & $0.073{\pm}0.008$ & $0.772{\pm}0.015$ & $\phantom{0}2.069{\pm}\phantom{0}0.226$ & $0.050{\pm}0.007$ \\
 & \wpls{}+MIND & Dataset-Fit & $0.793{\pm}0.012$ & $\phantom{0}1.953{\pm}\phantom{0}0.222$ & $0.073{\pm}0.005$ & $0.787{\pm}0.013$ & $\phantom{0}1.959{\pm}\phantom{0}0.219$ & $0.038{\pm}0.004$ \\
 & \wpls{}+MIND & Pair-Fit & $0.793{\pm}0.012$ & $\phantom{0}1.960{\pm}\phantom{0}0.222$ & $0.073{\pm}0.005$ & $0.787{\pm}0.013$ & $\phantom{0}1.962{\pm}\phantom{0}0.225$ & $0.038{\pm}0.004$ \\
\hline
\multirow{12}{*}{MedSAM2} & Axial & Dataset-Fit & $0.637{\pm}0.037$ & $\phantom{0}3.451{\pm}\phantom{0}0.520$ & $0.069{\pm}0.006$ & $0.645{\pm}0.031$ & $\phantom{0}3.299{\pm}\phantom{0}0.524$ & $0.078{\pm}0.011$ \\
 & Axial & Pair-Fit & $0.634{\pm}0.037$ & $\phantom{0}3.499{\pm}\phantom{0}0.532$ & $0.069{\pm}0.006$ & $0.645{\pm}0.031$ & $\phantom{0}3.302{\pm}\phantom{0}0.542$ & $0.077{\pm}0.010$ \\
 & Axial+MIND & Dataset-Fit & $0.789{\pm}0.012$ & $\phantom{0}2.020{\pm}\phantom{0}0.268$ & $0.072{\pm}0.006$ & $0.784{\pm}0.012$ & $\phantom{0}1.991{\pm}\phantom{0}0.230$ & $0.039{\pm}0.004$ \\
 & Axial+MIND & Pair-Fit & $0.789{\pm}0.012$ & $\phantom{0}2.019{\pm}\phantom{0}0.265$ & $0.072{\pm}0.006$ & $0.784{\pm}0.012$ & $\phantom{0}1.996{\pm}\phantom{0}0.235$ & $0.039{\pm}0.004$ \\
 & \pcatd{} & Dataset-Fit & $0.612{\pm}0.045$ & $\phantom{0}3.718{\pm}\phantom{0}0.548$ & $0.080{\pm}0.006$ & $0.636{\pm}0.033$ & $\phantom{0}3.295{\pm}\phantom{0}0.453$ & $0.090{\pm}0.010$ \\
 & \pcatd{} & Pair-Fit & $0.614{\pm}0.037$ & $\phantom{0}3.656{\pm}\phantom{0}0.495$ & $0.071{\pm}0.006$ & $0.636{\pm}0.040$ & $\phantom{0}3.560{\pm}\phantom{0}0.630$ & $0.000{\pm}0.000$ \\
 & \pcatd{}+MIND & Dataset-Fit & $0.789{\pm}0.012$ & $\phantom{0}2.019{\pm}\phantom{0}0.262$ & $0.072{\pm}0.006$ & $0.786{\pm}0.013$ & $\phantom{0}1.991{\pm}\phantom{0}0.228$ & $0.039{\pm}0.004$ \\
 & \pcatd{}+MIND & Pair-Fit & $0.789{\pm}0.012$ & $\phantom{0}2.019{\pm}\phantom{0}0.261$ & $0.072{\pm}0.006$ & $0.787{\pm}0.013$ & $\phantom{0}1.980{\pm}\phantom{0}0.235$ & $0.045{\pm}0.004$ \\
 & \wpls{} & Dataset-Fit & $\mathbf{0.799{\pm}0.011}$ & $\phantom{0}1.902{\pm}\phantom{0}0.204$ & $0.092{\pm}0.008$ & $\mathbf{0.798{\pm}0.011}$ & $\phantom{0}1.914{\pm}\phantom{0}0.197$ & $0.105{\pm}0.008$ \\
 & \wpls{} & Pair-Fit & $0.799{\pm}0.011$ & $\phantom{0}1.890{\pm}\phantom{0}0.199$ & $0.087{\pm}0.008$ & $0.797{\pm}0.011$ & $\phantom{0}1.912{\pm}\phantom{0}0.191$ & $0.094{\pm}0.011$ \\
 & \wpls{}+MIND & Dataset-Fit & $0.790{\pm}0.012$ & $\phantom{0}2.018{\pm}\phantom{0}0.262$ & $0.073{\pm}0.005$ & $0.788{\pm}0.013$ & $\phantom{0}1.970{\pm}\phantom{0}0.231$ & $0.045{\pm}0.004$ \\
 & \wpls{}+MIND & Pair-Fit & $0.790{\pm}0.012$ & $\phantom{0}2.007{\pm}\phantom{0}0.252$ & $0.073{\pm}0.006$ & $0.788{\pm}0.012$ & $\phantom{0}1.972{\pm}\phantom{0}0.236$ & $0.045{\pm}0.005$ \\
\hline
\multirow{12}{*}{SAM3} & Axial & Dataset-Fit & $0.664{\pm}0.029$ & $\phantom{0}3.127{\pm}\phantom{0}0.501$ & $0.054{\pm}0.008$ & $0.663{\pm}0.031$ & $\phantom{0}3.193{\pm}\phantom{0}0.460$ & $0.067{\pm}0.006$ \\
 & Axial & Pair-Fit & $0.660{\pm}0.030$ & $\phantom{0}3.265{\pm}\phantom{0}0.519$ & $0.066{\pm}0.008$ & $0.657{\pm}0.031$ & $\phantom{0}3.284{\pm}\phantom{0}0.478$ & $0.056{\pm}0.006$ \\
 & Axial+MIND & Dataset-Fit & $0.787{\pm}0.012$ & $\phantom{0}2.021{\pm}\phantom{0}0.261$ & $0.071{\pm}0.006$ & $0.781{\pm}0.012$ & $\phantom{0}2.051{\pm}\phantom{0}0.261$ & $0.044{\pm}0.006$ \\
 & Axial+MIND & Pair-Fit & $0.788{\pm}0.013$ & $\phantom{0}2.014{\pm}\phantom{0}0.290$ & $0.069{\pm}0.006$ & $0.781{\pm}0.013$ & $\phantom{0}2.045{\pm}\phantom{0}0.261$ & $0.044{\pm}0.006$ \\
 & \pcatd{} & Dataset-Fit & $0.622{\pm}0.042$ & $\phantom{0}3.535{\pm}\phantom{0}0.556$ & $0.061{\pm}0.008$ & $0.618{\pm}0.042$ & $\phantom{0}3.563{\pm}\phantom{0}0.542$ & $0.046{\pm}0.008$ \\
 & \pcatd{} & Pair-Fit & $0.623{\pm}0.041$ & $\phantom{0}3.538{\pm}\phantom{0}0.556$ & $0.062{\pm}0.008$ & $0.617{\pm}0.041$ & $\phantom{0}3.573{\pm}\phantom{0}0.532$ & $0.046{\pm}0.008$ \\
 & \pcatd{}+MIND & Dataset-Fit & $0.780{\pm}0.014$ & $\phantom{0}2.060{\pm}\phantom{0}0.278$ & $0.069{\pm}0.006$ & $0.774{\pm}0.014$ & $\phantom{0}2.096{\pm}\phantom{0}0.251$ & $0.038{\pm}0.006$ \\
 & \pcatd{}+MIND & Pair-Fit & $0.782{\pm}0.014$ & $\phantom{0}2.044{\pm}\phantom{0}0.287$ & $0.074{\pm}0.006$ & $0.775{\pm}0.014$ & $\phantom{0}2.088{\pm}\phantom{0}0.245$ & $0.038{\pm}0.006$ \\
 & \wpls{} & Dataset-Fit & $0.749{\pm}0.031$ & $\phantom{0}2.426{\pm}\phantom{0}0.384$ & $0.082{\pm}0.008$ & $0.748{\pm}0.032$ & $\phantom{0}2.450{\pm}\phantom{0}0.398$ & $0.075{\pm}0.009$ \\
 & \wpls{} & Pair-Fit & $0.765{\pm}0.019$ & $\phantom{0}2.262{\pm}\phantom{0}0.247$ & $0.096{\pm}0.009$ & $0.761{\pm}0.021$ & $\phantom{0}2.244{\pm}\phantom{0}0.246$ & $0.077{\pm}0.011$ \\
 & \wpls{}+MIND & Dataset-Fit & $0.793{\pm}0.013$ & $\phantom{0}1.980{\pm}\phantom{0}0.243$ & $0.091{\pm}0.006$ & $0.789{\pm}0.014$ & $\phantom{0}1.955{\pm}\phantom{0}0.243$ & $0.050{\pm}0.005$ \\
 & \wpls{}+MIND & Pair-Fit & $0.794{\pm}0.012$ & $\phantom{0}1.949{\pm}\phantom{0}0.204$ & $0.078{\pm}0.005$ & $0.788{\pm}0.013$ & $\phantom{0}1.970{\pm}\phantom{0}0.225$ & $0.043{\pm}0.005$ \\
\hline
\end{tabular}%
}
\end{table}

\section{Additional kNN Segmentation Results}
\label{sec:app:knn}

This section provides additional kNN segmentation results that support Section~\ref{ssec:results_knn}: an all-encoder version of the direction-specific radar plot, full direction-specific kNN tables that separate the two transfer directions within each category, and a sensitivity analysis over the number of nearest neighbors $k$.

\subsection{All-Encoder Direction-Specific Radar Plots}
\label{sec:app:knn_radar_all}

\begin{figure}[h!]
    \centering
    \includegraphics[width=\textwidth]{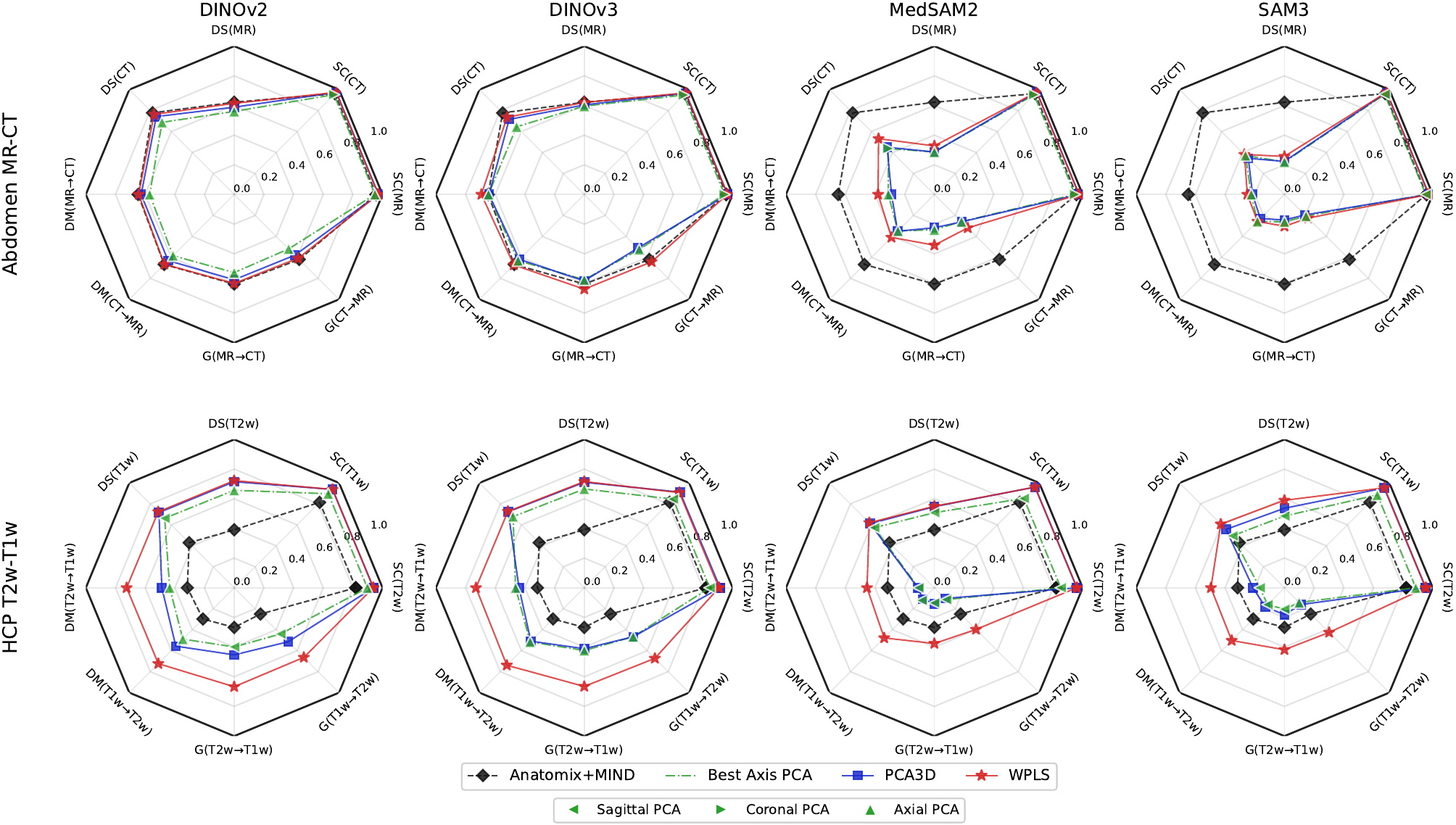}
    \caption{All-encoder voxelwise kNN segmentation Dice radar plots under the \textbf{Dataset-Fit} setting with $k=7$. 
    Rows correspond to datasets, and columns correspond to frozen ViT encoders. 
    Axes show direction-specific Self-Consistency (SC), Different Subject (DS), Different Modality (DM), and Generalization (G) groups. 
    Higher values indicate better label transfer by direct nearest-neighbor search in feature space. 
    The plots compare Anatomix+MIND, Best Axis PCA, \pcatd{}, and \wpls{}. 
    Marker shapes on the Best Axis PCA curve indicate which single-axis PCA feature was selected for each direction-specific group.}
    \label{fig:knn_radar_all_encoders}
\end{figure}

Figure~\ref{fig:knn_radar_all_encoders} extends the main-text DINOv3 radar (Fig.~\ref{fig:knn_radar}) to all four frozen encoders, complementing Table~\ref{tab:knn_all_groups} and the direction-specific tables below. The all-encoder view makes visible the encoder-dependent ordering reported in the main text: \wpls{} consistently expands the Generalization region for every encoder on HCP T2w--T1w, with the largest absolute gains for the encoders whose single-axis features are weakest (MedSAM2, SAM3), while the DINO-based encoders lead in absolute G-Dice across both datasets.

\subsection{Direction-Specific kNN Segmentation Results}
\label{sec:app:knn_dirres}

Tables~\ref{tab:knn_directional_abdmrct} and~\ref{tab:knn_directional_hcpt2t1} report the direction-specific kNN segmentation results corresponding to the pooled kNN table in Section~\ref{ssec:results_knn} and the radar visualizations in Figs.~\ref{fig:knn_radar} and~\ref{fig:knn_radar_all_encoders}. These tables separate the two directions within each SC, DS, DM, and G group, making modality and direction asymmetries visible while keeping the main text compact. They also report the three single-axis PCA features (Sagittal, Coronal, Axial) explicitly, rather than the Best Axis PCA oracle used in the main table. All results are evaluated under the \textbf{Dataset-Fit} setting with $k=7$ and cosine-similarity nearest-neighbor search.

\begin{table}
\centering
\caption{Directional voxelwise kNN segmentation Dice ($k=7$) on Abdomen MR--CT under the \textbf{Dataset-Fit} setting. Each cell reports the mean $\pm$ standard deviation computed over all per-volume Dice values in the corresponding direction-specific group. Higher is better. Bold marks the best result per column.}
\label{tab:knn_directional_abdmrct}
\vspace{4pt}
\scriptsize
\setlength{\tabcolsep}{2.0pt}
\resizebox{\textwidth}{!}{%
\begin{tabular}{l|l||cc|cc|cc|cc}
\hline
\multirow{2}{*}{\textbf{Encoder}} & \multirow{2}{*}{\textbf{Method}} & \multicolumn{2}{c|}{\textbf{Self-Consistency}} & \multicolumn{2}{c|}{\textbf{Different Subject}} & \multicolumn{2}{c|}{\textbf{Different Modality}} & \multicolumn{2}{c}{\textbf{Generalization}} \\
 &  & MR $\uparrow$ & CT $\uparrow$ & MR $\uparrow$ & CT $\uparrow$ & MR$\to$CT $\uparrow$ & CT$\to$MR $\uparrow$ & MR$\to$CT $\uparrow$ & CT$\to$MR $\uparrow$ \\
\hline\hline
\multirow{3}{*}{--} & MIND & 0.485$\pm$0.042 & 0.440$\pm$0.054 & 0.036$\pm$0.013 & 0.091$\pm$0.039 & 0.065$\pm$0.023 & 0.063$\pm$0.019 & 0.045$\pm$0.009 & 0.040$\pm$0.007 \\
 & Anatomix & 0.853$\pm$0.092 & 0.841$\pm$0.057 & 0.500$\pm$0.173 & 0.603$\pm$0.124 & 0.550$\pm$0.158 & 0.564$\pm$0.127 & 0.423$\pm$0.146 & 0.498$\pm$0.117 \\
 & Anatomix+MIND & 0.957$\pm$0.025 & 0.961$\pm$0.008 & \textbf{0.622$\pm$0.163} & \textbf{0.778$\pm$0.067} & 0.648$\pm$0.190 & 0.667$\pm$0.168 & 0.604$\pm$0.156 & 0.620$\pm$0.135 \\
\hline
\multirow{5}{*}{DINOv2} & Axial & 0.949$\pm$0.006 & 0.951$\pm$0.004 & 0.560$\pm$0.140 & 0.688$\pm$0.102 & 0.570$\pm$0.140 & 0.583$\pm$0.121 & 0.527$\pm$0.100 & 0.519$\pm$0.090 \\
 & Sagittal & 0.948$\pm$0.010 & 0.951$\pm$0.006 & 0.261$\pm$0.111 & 0.424$\pm$0.133 & 0.287$\pm$0.127 & 0.282$\pm$0.107 & 0.237$\pm$0.086 & 0.235$\pm$0.079 \\
 & Coronal & 0.948$\pm$0.009 & 0.952$\pm$0.005 & 0.424$\pm$0.231 & 0.638$\pm$0.106 & 0.488$\pm$0.164 & 0.479$\pm$0.163 & 0.398$\pm$0.113 & 0.417$\pm$0.102 \\
 & \pcatd{} & 0.972$\pm$0.004 & 0.973$\pm$0.002 & 0.589$\pm$0.155 & 0.743$\pm$0.091 & 0.627$\pm$0.157 & 0.630$\pm$0.174 & 0.575$\pm$0.116 & 0.579$\pm$0.119 \\
 & \wpls{} & 0.972$\pm$0.004 & 0.974$\pm$0.002 & 0.615$\pm$0.163 & 0.764$\pm$0.085 & 0.644$\pm$0.156 & 0.666$\pm$0.154 & 0.600$\pm$0.121 & 0.609$\pm$0.119 \\
\hline
\multirow{5}{*}{DINOv3} & Axial & 0.943$\pm$0.006 & 0.945$\pm$0.005 & 0.596$\pm$0.139 & 0.645$\pm$0.121 & 0.646$\pm$0.133 & 0.631$\pm$0.133 & 0.575$\pm$0.134 & 0.524$\pm$0.131 \\
 & Sagittal & 0.945$\pm$0.009 & 0.947$\pm$0.006 & 0.269$\pm$0.150 & 0.418$\pm$0.100 & 0.338$\pm$0.154 & 0.363$\pm$0.134 & 0.263$\pm$0.143 & 0.290$\pm$0.107 \\
 & Coronal & 0.946$\pm$0.009 & 0.948$\pm$0.005 & 0.377$\pm$0.216 & 0.639$\pm$0.114 & 0.443$\pm$0.242 & 0.466$\pm$0.239 & 0.384$\pm$0.220 & 0.376$\pm$0.132 \\
 & \pcatd{} & 0.966$\pm$0.005 & 0.966$\pm$0.004 & 0.604$\pm$0.182 & 0.716$\pm$0.137 & 0.642$\pm$0.145 & 0.619$\pm$0.206 & 0.582$\pm$0.141 & 0.511$\pm$0.106 \\
 & \wpls{} & 0.965$\pm$0.005 & 0.969$\pm$0.004 & 0.620$\pm$0.174 & 0.736$\pm$0.121 & \textbf{0.694$\pm$0.150} & \textbf{0.672$\pm$0.169} & \textbf{0.638$\pm$0.171} & \textbf{0.642$\pm$0.158} \\
\hline
\multirow{5}{*}{MedSAM2} & Axial & 0.946$\pm$0.006 & 0.949$\pm$0.005 & 0.287$\pm$0.160 & 0.432$\pm$0.110 & 0.310$\pm$0.084 & 0.347$\pm$0.106 & 0.238$\pm$0.071 & 0.259$\pm$0.070 \\
 & Sagittal & 0.947$\pm$0.009 & 0.949$\pm$0.006 & 0.179$\pm$0.111 & 0.304$\pm$0.135 & 0.202$\pm$0.101 & 0.256$\pm$0.101 & 0.152$\pm$0.052 & 0.201$\pm$0.073 \\
 & Coronal & 0.948$\pm$0.008 & 0.951$\pm$0.005 & 0.261$\pm$0.142 & 0.438$\pm$0.150 & 0.252$\pm$0.102 & 0.288$\pm$0.120 & 0.199$\pm$0.087 & 0.235$\pm$0.070 \\
 & \pcatd{} & \textbf{0.976$\pm$0.003} & \textbf{0.977$\pm$0.002} & 0.286$\pm$0.176 & 0.451$\pm$0.145 & 0.289$\pm$0.114 & 0.350$\pm$0.140 & 0.224$\pm$0.089 & 0.261$\pm$0.081 \\
 & \wpls{} & 0.975$\pm$0.003 & \textbf{0.976$\pm$0.002} & 0.327$\pm$0.193 & 0.532$\pm$0.171 & 0.379$\pm$0.192 & 0.409$\pm$0.148 & 0.342$\pm$0.196 & 0.318$\pm$0.138 \\
\hline
\multirow{5}{*}{SAM3} & Axial & 0.952$\pm$0.005 & 0.955$\pm$0.004 & 0.219$\pm$0.097 & 0.372$\pm$0.130 & 0.224$\pm$0.054 & 0.258$\pm$0.075 & 0.184$\pm$0.049 & 0.206$\pm$0.060 \\
 & Sagittal & 0.953$\pm$0.009 & 0.955$\pm$0.005 & 0.149$\pm$0.078 & 0.248$\pm$0.094 & 0.129$\pm$0.041 & 0.171$\pm$0.045 & 0.103$\pm$0.029 & 0.138$\pm$0.034 \\
 & Coronal & 0.948$\pm$0.008 & 0.951$\pm$0.005 & 0.154$\pm$0.110 & 0.287$\pm$0.082 & 0.121$\pm$0.036 & 0.148$\pm$0.066 & 0.111$\pm$0.041 & 0.118$\pm$0.040 \\
 & \pcatd{} & 0.973$\pm$0.004 & 0.974$\pm$0.003 & 0.223$\pm$0.120 & 0.345$\pm$0.111 & 0.214$\pm$0.069 & 0.230$\pm$0.067 & 0.172$\pm$0.048 & 0.195$\pm$0.062 \\
 & \wpls{} & 0.974$\pm$0.004 & 0.975$\pm$0.002 & 0.256$\pm$0.140 & 0.382$\pm$0.123 & 0.250$\pm$0.074 & 0.257$\pm$0.075 & 0.215$\pm$0.061 & 0.225$\pm$0.069 \\
\hline
\end{tabular}%
}
\end{table}

\begin{table}[h!]
\centering
\caption{Directional voxelwise kNN segmentation Dice ($k=7$) on HCP T2w--T1w under the \textbf{Dataset-Fit} setting. Each cell reports the mean $\pm$ standard deviation computed over all per-volume Dice values in the corresponding direction-specific group. Higher is better. Bold marks the best result per column.}
\label{tab:knn_directional_hcpt2t1}
\vspace{4pt}
\scriptsize
\setlength{\tabcolsep}{2.0pt}
\resizebox{\textwidth}{!}{%
\begin{tabular}{l|l||cc|cc|cc|cc}
\hline
\multirow{2}{*}{\textbf{Encoder}} & \multirow{2}{*}{\textbf{Method}} & \multicolumn{2}{c|}{\textbf{Self-Consistency}} & \multicolumn{2}{c|}{\textbf{Different Subject}} & \multicolumn{2}{c|}{\textbf{Different Modality}} & \multicolumn{2}{c}{\textbf{Generalization}} \\
 &  & T2w $\uparrow$ & T1w $\uparrow$ & T2w $\uparrow$ & T1w $\uparrow$ & T2w$\to$T1w $\uparrow$ & T1w$\to$T2w $\uparrow$ & T2w$\to$T1w $\uparrow$ & T1w$\to$T2w $\uparrow$ \\
\hline\hline
\multirow{3}{*}{--} & MIND & 0.497$\pm$0.017 & 0.494$\pm$0.019 & 0.100$\pm$0.003 & 0.109$\pm$0.004 & 0.119$\pm$0.006 & 0.113$\pm$0.006 & 0.088$\pm$0.003 & 0.083$\pm$0.003 \\
 & Anatomix & 0.318$\pm$0.023 & 0.298$\pm$0.019 & 0.276$\pm$0.021 & 0.273$\pm$0.019 & 0.183$\pm$0.013 & 0.206$\pm$0.014 & 0.178$\pm$0.014 & 0.193$\pm$0.016 \\
 & Anatomix+MIND & 0.820$\pm$0.012 & 0.815$\pm$0.012 & 0.391$\pm$0.022 & 0.430$\pm$0.021 & 0.316$\pm$0.018 & 0.298$\pm$0.020 & 0.267$\pm$0.012 & 0.250$\pm$0.018 \\
\hline
\multirow{5}{*}{DINOv2} & Axial & 0.898$\pm$0.003 & 0.897$\pm$0.003 & 0.657$\pm$0.014 & 0.660$\pm$0.014 & 0.435$\pm$0.020 & 0.489$\pm$0.021 & 0.390$\pm$0.021 & 0.438$\pm$0.024 \\
 & Sagittal & 0.897$\pm$0.003 & 0.896$\pm$0.003 & 0.641$\pm$0.020 & 0.663$\pm$0.021 & 0.434$\pm$0.012 & 0.487$\pm$0.014 & 0.399$\pm$0.012 & 0.441$\pm$0.017 \\
 & Coronal & 0.892$\pm$0.004 & 0.891$\pm$0.004 & 0.629$\pm$0.017 & 0.641$\pm$0.014 & 0.404$\pm$0.022 & 0.448$\pm$0.025 & 0.368$\pm$0.030 & 0.404$\pm$0.019 \\
 & \pcatd{} & 0.939$\pm$0.002 & 0.938$\pm$0.002 & 0.715$\pm$0.019 & 0.717$\pm$0.014 & 0.488$\pm$0.021 & 0.556$\pm$0.015 & 0.452$\pm$0.023 & 0.515$\pm$0.013 \\
 & \wpls{} & 0.938$\pm$0.002 & 0.938$\pm$0.002 & \textbf{0.723$\pm$0.018} & 0.724$\pm$0.014 & 0.725$\pm$0.011 & 0.723$\pm$0.011 & \textbf{0.667$\pm$0.014} & 0.664$\pm$0.015 \\
\hline
\multirow{5}{*}{DINOv3} & Axial & 0.843$\pm$0.005 & 0.841$\pm$0.006 & 0.666$\pm$0.015 & 0.679$\pm$0.014 & 0.463$\pm$0.019 & 0.516$\pm$0.034 & 0.423$\pm$0.030 & 0.466$\pm$0.028 \\
 & Sagittal & 0.845$\pm$0.005 & 0.846$\pm$0.004 & 0.653$\pm$0.033 & 0.664$\pm$0.026 & 0.424$\pm$0.025 & 0.459$\pm$0.037 & 0.388$\pm$0.037 & 0.412$\pm$0.037 \\
 & Coronal & 0.838$\pm$0.005 & 0.837$\pm$0.006 & 0.641$\pm$0.025 & 0.657$\pm$0.020 & 0.449$\pm$0.021 & 0.481$\pm$0.022 & 0.406$\pm$0.020 & 0.429$\pm$0.032 \\
 & \pcatd{} & 0.916$\pm$0.003 & 0.912$\pm$0.004 & 0.711$\pm$0.021 & 0.725$\pm$0.017 & 0.440$\pm$0.019 & 0.509$\pm$0.026 & 0.411$\pm$0.025 & 0.467$\pm$0.022 \\
 & \wpls{} & 0.908$\pm$0.003 & 0.907$\pm$0.004 & 0.716$\pm$0.017 & \textbf{0.728$\pm$0.015} & \textbf{0.730$\pm$0.013} & \textbf{0.738$\pm$0.011} & 0.665$\pm$0.017 & \textbf{0.673$\pm$0.015} \\
\hline
\multirow{5}{*}{MedSAM2} & Axial & 0.851$\pm$0.005 & 0.851$\pm$0.006 & 0.466$\pm$0.029 & 0.518$\pm$0.024 & 0.101$\pm$0.007 & 0.096$\pm$0.010 & 0.097$\pm$0.007 & 0.090$\pm$0.010 \\
 & Sagittal & 0.852$\pm$0.004 & 0.852$\pm$0.004 & 0.506$\pm$0.023 & 0.573$\pm$0.021 & 0.107$\pm$0.007 & 0.121$\pm$0.010 & 0.102$\pm$0.010 & 0.113$\pm$0.006 \\
 & Coronal & 0.842$\pm$0.005 & 0.842$\pm$0.005 & 0.447$\pm$0.030 & 0.500$\pm$0.022 & 0.094$\pm$0.005 & 0.089$\pm$0.007 & 0.090$\pm$0.006 & 0.085$\pm$0.007 \\
 & \pcatd{} & \textbf{0.961$\pm$0.001} & \textbf{0.960$\pm$0.002} & 0.547$\pm$0.032 & 0.612$\pm$0.019 & 0.112$\pm$0.008 & 0.108$\pm$0.008 & 0.108$\pm$0.007 & 0.102$\pm$0.007 \\
 & \wpls{} & 0.960$\pm$0.001 & 0.960$\pm$0.002 & 0.550$\pm$0.032 & 0.620$\pm$0.016 & 0.454$\pm$0.013 & 0.478$\pm$0.018 & 0.376$\pm$0.017 & 0.397$\pm$0.012 \\
\hline
\multirow{5}{*}{SAM3} & Axial & 0.886$\pm$0.004 & 0.886$\pm$0.004 & 0.429$\pm$0.028 & 0.465$\pm$0.033 & 0.157$\pm$0.009 & 0.153$\pm$0.010 & 0.140$\pm$0.007 & 0.140$\pm$0.008 \\
 & Sagittal & 0.884$\pm$0.004 & 0.884$\pm$0.004 & 0.484$\pm$0.033 & 0.495$\pm$0.030 & 0.166$\pm$0.011 & 0.161$\pm$0.012 & 0.144$\pm$0.014 & 0.140$\pm$0.010 \\
 & Coronal & 0.878$\pm$0.004 & 0.878$\pm$0.004 & 0.464$\pm$0.042 & 0.470$\pm$0.045 & 0.163$\pm$0.010 & 0.143$\pm$0.010 & 0.144$\pm$0.012 & 0.129$\pm$0.011 \\
 & \pcatd{} & 0.952$\pm$0.002 & 0.951$\pm$0.002 & 0.536$\pm$0.041 & 0.557$\pm$0.035 & 0.212$\pm$0.017 & 0.182$\pm$0.016 & 0.181$\pm$0.015 & 0.159$\pm$0.012 \\
 & \wpls{} & 0.953$\pm$0.002 & 0.953$\pm$0.002 & 0.590$\pm$0.043 & 0.607$\pm$0.036 & 0.496$\pm$0.019 & 0.503$\pm$0.018 & 0.418$\pm$0.026 & 0.426$\pm$0.024 \\
\hline
\end{tabular}%
}
\end{table}

\subsection{Sensitivity to the Number of Neighbors}
\label{sec:app:knn_sensk}

To assess whether the voxelwise kNN segmentation results depended strongly on the choice of the number of neighbors, we repeated the evaluation for $k \in \{1,3,5,7,9,11\}$ under the same \textbf{Dataset-Fit} setting used in the main text. Rather than reporting all Dice values for every dataset, encoder, method, direction, and value of $k$, we summarize sensitivity using the absolute Dice range
\begin{equation}
\Delta_k \mathrm{Dice}
=
\max_{k \in \{1,3,5,7,9,11\}} \mathrm{Dice}(k)
-
\min_{k \in \{1,3,5,7,9,11\}} \mathrm{Dice}(k).
\end{equation}
Lower values indicate that the corresponding result is less sensitive to the choice of $k$.

\begin{figure*}
    \centering
    \includegraphics[width=\textwidth]{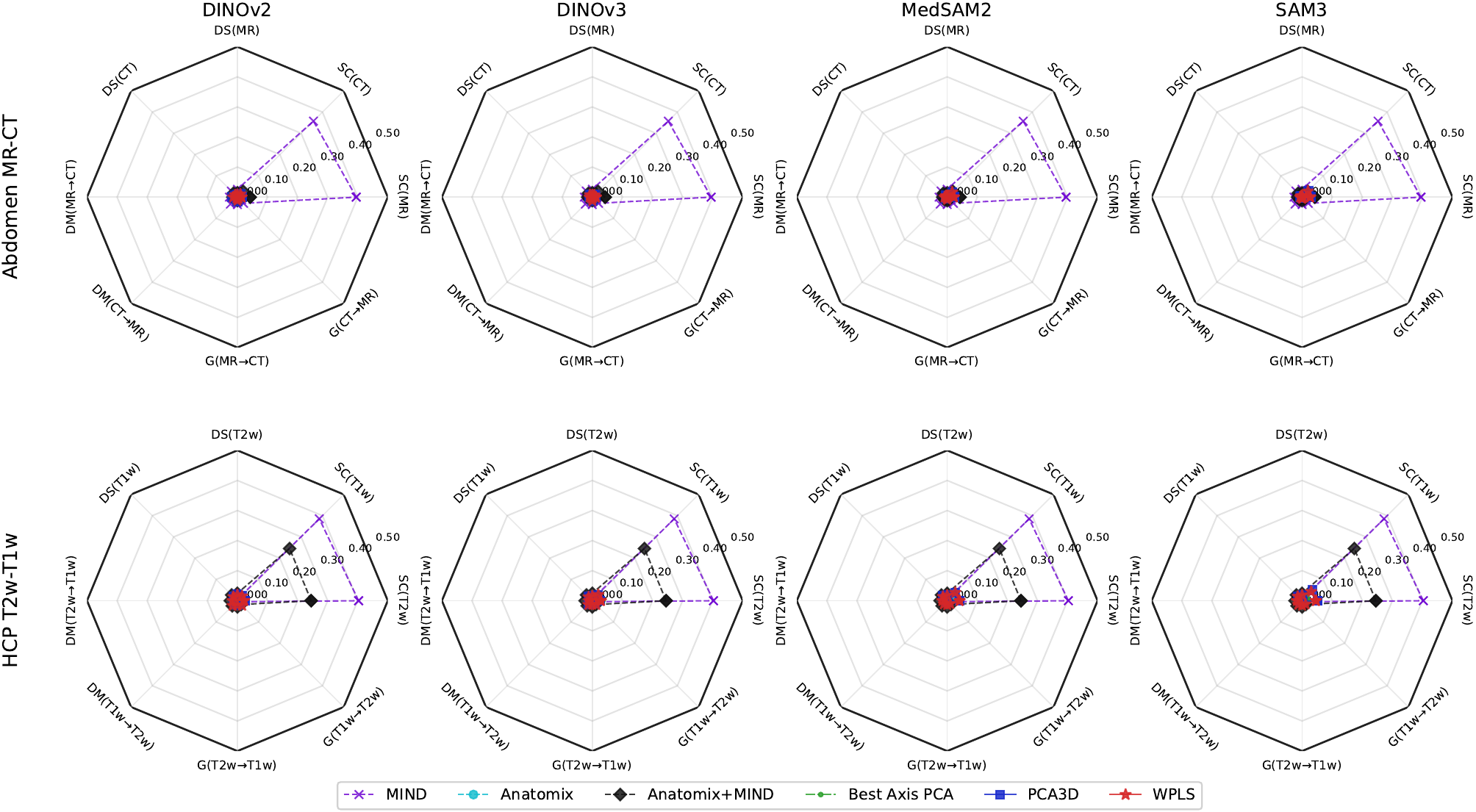}
    \caption{kNN sensitivity to the number of neighbors under the \textbf{Dataset-Fit} setting. Each radar axis reports the absolute Dice range
    $\Delta_k \mathrm{Dice} = \max_k \mathrm{Dice}(k) - \min_k \mathrm{Dice}(k)$
    over $k \in \{1,3,5,7,9,11\}$ for a direction-specific transfer category. Lower values indicate lower sensitivity to the choice of $k$. Most ViT-based representations show small variation across $k$, particularly in the DS, DM, and G transfer categories, indicating that the main-text results at $k=7$ are not driven by a particular neighbor count. Larger ranges are mainly observed for MIND-based self-consistency settings, where increasing $k$ changes local nearest-neighbor behavior more strongly.}
    \label{fig:knn_k_range_abs}
\end{figure*}

Figure~\ref{fig:knn_k_range_abs} shows the resulting sensitivity profiles across all encoders and direction-specific transfer categories. Overall, the main conclusions are stable across the evaluated range of $k$. For most ViT-based representations, especially in the DS, DM, and G transfer categories, $\Delta_k \mathrm{Dice}$ is small relative to the performance differences reported in the main text. This indicates that the use of $k=7$ in Section~\ref{ssec:results_knn} did not drive the observed ordering between Best Axis PCA, \pcatd{}, and \wpls{}.

The largest sensitivities are observed in the self-consistency categories, especially for local-descriptor-dominated representations such as MIND and Anatomix+MIND. In these settings, increasing $k$ changes how strongly very local neighborhoods are averaged, and highly local descriptors can assign similar features to nearby voxels without necessarily forming stable anatomical neighborhoods under broader averaging. These high self-consistency sensitivities are not central to the main correspondence claim, since the most informative settings are DS, DM, and especially G, where subject identity and/or modality change. In these harder transfer categories, the qualitative ranking remains stable: triplanar ViT features remain more robust than local descriptors, and \wpls{} remains competitive or strongest in the cross-subject, cross-modality categories. Quantitatively, MIND reaches $\Delta_k\mathrm{Dice} \approx 0.40$ in self-consistency on both datasets, and Anatomix+MIND reaches $\Delta_k\mathrm{Dice} \approx 0.25$ in self-consistency on HCP T2w--T1w, while no ViT-based representation exceeds $\Delta_k\mathrm{Dice} \approx 0.06$ in any category.

\newpage
\section{Complete Direction-Specific Landmark Localization Results}
\label{sec:app:landmark}

\subsection{Directional Landmark Localization with Median-Pair Aggregation}
\label{sec:app:landmark_medpair}

\begin{table}[h!]
\centering
\caption{Directional landmark localization error on Abdomen MR--CT under the \textbf{Dataset-Fit} setting using top-1 nearest-neighbor matching in feature space ($k=1$, $L_2$ distance). Distances are reported in millimeters as the mean$\pm$standard deviation across landmarks, where each landmark value is first computed as the median over held-out query--key pairs within each direction-specific group. Lower is better. Bold marks the best result per column.}
\label{tab:landmark_directional_abdmrct_medianpair}
\vspace{4pt}
\scriptsize
\setlength{\tabcolsep}{2.0pt}
\resizebox{\textwidth}{!}{%
\begin{tabular}{l|l||cc|cc|cc|cc}
\hline
\multirow{2}{*}{\textbf{Encoder}} & \multirow{2}{*}{\textbf{Method}} & \multicolumn{2}{c|}{\textbf{Self-Consistency}} & \multicolumn{2}{c|}{\textbf{Different Subject}} & \multicolumn{2}{c|}{\textbf{Different Modality}} & \multicolumn{2}{c}{\textbf{Generalization}} \\
 &  & MR $\downarrow$ & CT $\downarrow$ & MR $\downarrow$ & CT $\downarrow$ & MR$\to$CT $\downarrow$ & CT$\to$MR $\downarrow$ & MR$\to$CT $\downarrow$ & CT$\to$MR $\downarrow$ \\
\hline\hline
\multirow{3}{*}{--} & MIND & \phantom{0}92.71$\pm$\phantom{0}61.92 & 110.79$\pm$\phantom{0}53.21 & 168.35$\pm$\phantom{0}32.13 & 175.58$\pm$\phantom{0}33.00 & 135.74$\pm$\phantom{0}11.80 & 185.78$\pm$\phantom{0}41.50 & 157.05$\pm$\phantom{0}31.72 & 175.00$\pm$\phantom{0}27.67 \\
 & Anatomix & \phantom{00}2.21$\pm$\phantom{00}0.41 & \phantom{00}2.31$\pm$\phantom{00}0.40 & \phantom{0}47.29$\pm$\phantom{0}18.13 & \phantom{0}36.92$\pm$\phantom{0}17.93 & \phantom{0}46.02$\pm$\phantom{0}18.92 & \phantom{0}39.85$\pm$\phantom{0}24.53 & \phantom{0}79.95$\pm$\phantom{0}29.73 & \phantom{0}40.14$\pm$\phantom{0}18.46 \\
 & Anatomix+MIND & \phantom{00}2.21$\pm$\phantom{00}0.41 & \phantom{0}11.11$\pm$\phantom{0}17.94 & \phantom{0}63.22$\pm$\phantom{0}36.97 & \phantom{0}38.19$\pm$\phantom{0}17.17 & \phantom{0}54.90$\pm$\phantom{0}18.44 & \phantom{0}41.67$\pm$\phantom{0}20.78 & \phantom{0}88.63$\pm$\phantom{0}33.44 & \phantom{0}53.09$\pm$\phantom{0}23.90 \\
\hline
\multirow{5}{*}{DINOv2} & Axial & \phantom{00}2.10$\pm$\phantom{00}0.21 & \phantom{00}2.31$\pm$\phantom{00}0.40 & \phantom{0}28.22$\pm$\phantom{00}7.06 & \phantom{0}32.22$\pm$\phantom{0}15.61 & \phantom{0}26.61$\pm$\phantom{00}9.26 & \phantom{0}40.71$\pm$\phantom{0}28.18 & \phantom{0}33.40$\pm$\phantom{00}4.88 & \phantom{0}53.58$\pm$\phantom{0}39.95 \\
 & Sagittal & \phantom{00}2.31$\pm$\phantom{00}0.40 & \phantom{00}2.10$\pm$\phantom{00}0.21 & 100.05$\pm$\phantom{0}49.80 & \phantom{0}56.46$\pm$\phantom{0}41.00 & \phantom{0}70.20$\pm$\phantom{0}34.77 & 140.40$\pm$\phantom{0}56.32 & 126.91$\pm$\phantom{0}38.49 & 127.10$\pm$\phantom{0}55.52 \\
 & Coronal & \phantom{00}2.10$\pm$\phantom{00}0.21 & \phantom{00}2.21$\pm$\phantom{00}0.24 & \phantom{0}41.53$\pm$\phantom{0}17.25 & \phantom{0}34.52$\pm$\phantom{0}22.68 & \phantom{0}55.56$\pm$\phantom{0}22.67 & \phantom{0}47.38$\pm$\phantom{0}25.99 & \phantom{0}50.30$\pm$\phantom{0}16.27 & \phantom{0}41.98$\pm$\phantom{0}15.61 \\
 & \pcatd{} & \textbf{\phantom{00}2.00$\pm$\phantom{00}0.00} & \textbf{\phantom{00}2.00$\pm$\phantom{00}0.00} & \phantom{0}27.83$\pm$\phantom{0}10.96 & \phantom{0}26.05$\pm$\phantom{00}9.07 & \phantom{0}32.52$\pm$\phantom{0}12.95 & \phantom{0}25.56$\pm$\phantom{0}23.10 & \phantom{0}32.55$\pm$\phantom{0}10.12 & \phantom{0}30.84$\pm$\phantom{0}13.19 \\
 & \wpls{} & \textbf{\phantom{00}2.00$\pm$\phantom{00}0.00} & \textbf{\phantom{00}2.00$\pm$\phantom{00}0.00} & \phantom{0}30.34$\pm$\phantom{00}5.52 & \phantom{0}26.94$\pm$\phantom{0}11.32 & \phantom{0}23.60$\pm$\phantom{00}3.49 & \phantom{0}24.29$\pm$\phantom{0}13.15 & \phantom{0}25.11$\pm$\phantom{00}4.32 & \textbf{\phantom{0}27.18$\pm$\phantom{0}11.88} \\
\hline
\multirow{5}{*}{DINOv3} & Axial & \phantom{00}2.31$\pm$\phantom{00}0.40 & \phantom{00}2.10$\pm$\phantom{00}0.21 & \phantom{0}31.67$\pm$\phantom{00}9.72 & \phantom{0}24.86$\pm$\phantom{00}5.26 & \phantom{0}23.15$\pm$\phantom{00}5.04 & \phantom{0}25.05$\pm$\phantom{0}10.95 & \phantom{0}25.05$\pm$\phantom{00}1.72 & \phantom{0}47.85$\pm$\phantom{0}24.18 \\
 & Sagittal & \phantom{00}2.21$\pm$\phantom{00}0.24 & \phantom{00}2.10$\pm$\phantom{00}0.21 & \phantom{0}87.17$\pm$\phantom{0}44.17 & \phantom{0}54.79$\pm$\phantom{0}56.81 & 116.86$\pm$\phantom{0}74.38 & 100.78$\pm$\phantom{0}82.80 & 126.92$\pm$\phantom{0}53.95 & 111.13$\pm$\phantom{0}70.47 \\
 & Coronal & \phantom{00}2.66$\pm$\phantom{00}0.60 & \phantom{00}2.41$\pm$\phantom{00}0.34 & \phantom{0}53.27$\pm$\phantom{0}29.23 & \phantom{0}22.74$\pm$\phantom{00}6.64 & \phantom{0}40.80$\pm$\phantom{0}18.47 & \phantom{0}27.68$\pm$\phantom{00}9.13 & \phantom{0}65.18$\pm$\phantom{00}7.85 & \phantom{0}46.21$\pm$\phantom{0}17.10 \\
 & \pcatd{} & \textbf{\phantom{00}2.00$\pm$\phantom{00}0.00} & \textbf{\phantom{00}2.00$\pm$\phantom{00}0.00} & \phantom{0}26.04$\pm$\phantom{00}3.73 & \textbf{\phantom{0}19.81$\pm$\phantom{00}4.20} & \phantom{0}20.85$\pm$\phantom{00}4.99 & \phantom{0}22.58$\pm$\phantom{00}6.44 & \phantom{0}21.78$\pm$\phantom{0}10.79 & \phantom{0}30.32$\pm$\phantom{0}12.51 \\
 & \wpls{} & \textbf{\phantom{00}2.00$\pm$\phantom{00}0.00} & \textbf{\phantom{00}2.00$\pm$\phantom{00}0.00} & \textbf{\phantom{0}21.57$\pm$\phantom{00}7.45} & \phantom{0}20.43$\pm$\phantom{00}5.87 & \textbf{\phantom{0}19.85$\pm$\phantom{00}4.41} & \textbf{\phantom{0}19.32$\pm$\phantom{00}3.43} & \textbf{\phantom{0}21.12$\pm$\phantom{00}5.92} & \phantom{0}29.58$\pm$\phantom{00}5.24 \\
\hline
\multirow{5}{*}{MedSAM2} & Axial & \phantom{00}2.31$\pm$\phantom{00}0.40 & \phantom{00}2.10$\pm$\phantom{00}0.21 & \phantom{0}82.61$\pm$\phantom{0}30.23 & \phantom{0}84.32$\pm$\phantom{0}47.13 & 109.67$\pm$\phantom{0}30.93 & 119.07$\pm$\phantom{0}60.79 & 107.62$\pm$\phantom{0}37.44 & \phantom{0}98.83$\pm$\phantom{0}29.09 \\
 & Sagittal & \phantom{00}2.21$\pm$\phantom{00}0.24 & \phantom{00}2.10$\pm$\phantom{00}0.21 & 150.60$\pm$\phantom{0}62.71 & 120.47$\pm$\phantom{0}50.83 & \phantom{0}99.00$\pm$\phantom{0}27.38 & \phantom{0}96.61$\pm$\phantom{0}35.63 & 146.97$\pm$\phantom{0}27.45 & 146.11$\pm$\phantom{0}36.37 \\
 & Coronal & \phantom{00}2.66$\pm$\phantom{00}0.60 & \phantom{00}2.41$\pm$\phantom{00}0.34 & 134.91$\pm$\phantom{00}8.45 & \phantom{0}92.49$\pm$\phantom{0}24.51 & 101.23$\pm$\phantom{0}10.12 & 101.76$\pm$\phantom{0}26.80 & 116.35$\pm$\phantom{00}6.66 & 122.40$\pm$\phantom{0}32.60 \\
 & \pcatd{} & \textbf{\phantom{00}2.00$\pm$\phantom{00}0.00} & \textbf{\phantom{00}2.00$\pm$\phantom{00}0.00} & 106.38$\pm$\phantom{0}29.35 & \phantom{0}71.84$\pm$\phantom{0}38.08 & 106.85$\pm$\phantom{0}40.09 & \phantom{0}67.05$\pm$\phantom{0}34.45 & 104.28$\pm$\phantom{0}14.96 & \phantom{0}97.83$\pm$\phantom{0}43.10 \\
 & \wpls{} & \textbf{\phantom{00}2.00$\pm$\phantom{00}0.00} & \textbf{\phantom{00}2.00$\pm$\phantom{00}0.00} & \phantom{0}78.87$\pm$\phantom{0}36.89 & \phantom{0}63.41$\pm$\phantom{0}27.23 & \phantom{0}98.97$\pm$\phantom{0}39.28 & \phantom{0}67.34$\pm$\phantom{0}20.07 & \phantom{0}69.77$\pm$\phantom{0}13.99 & \phantom{0}77.72$\pm$\phantom{0}31.93 \\
\hline
\multirow{5}{*}{SAM3} & Axial & \phantom{00}2.10$\pm$\phantom{00}0.21 & \phantom{00}2.31$\pm$\phantom{00}0.40 & 142.48$\pm$\phantom{0}32.63 & \phantom{0}75.08$\pm$\phantom{0}36.27 & 127.10$\pm$\phantom{0}15.32 & 112.19$\pm$\phantom{0}42.80 & 133.79$\pm$\phantom{0}21.25 & 112.98$\pm$\phantom{0}34.86 \\
 & Sagittal & \phantom{00}2.10$\pm$\phantom{00}0.21 & \phantom{00}2.10$\pm$\phantom{00}0.21 & 145.05$\pm$\phantom{0}24.89 & \phantom{0}88.69$\pm$\phantom{0}18.09 & 147.99$\pm$\phantom{0}23.16 & 143.35$\pm$\phantom{0}29.07 & 130.72$\pm$\phantom{0}28.62 & 152.75$\pm$\phantom{0}25.13 \\
 & Coronal & \phantom{00}2.66$\pm$\phantom{00}0.60 & \phantom{00}2.31$\pm$\phantom{00}0.21 & 147.41$\pm$\phantom{0}28.29 & \phantom{0}87.56$\pm$\phantom{0}31.53 & 153.32$\pm$\phantom{0}18.66 & 140.88$\pm$\phantom{0}19.83 & 152.99$\pm$\phantom{0}22.81 & 140.43$\pm$\phantom{0}35.90 \\
 & \pcatd{} & \textbf{\phantom{00}2.00$\pm$\phantom{00}0.00} & \textbf{\phantom{00}2.00$\pm$\phantom{00}0.00} & 121.00$\pm$\phantom{0}29.60 & \phantom{0}72.81$\pm$\phantom{0}17.53 & 130.55$\pm$\phantom{0}20.66 & 108.20$\pm$\phantom{0}16.56 & 147.83$\pm$\phantom{0}29.33 & 115.65$\pm$\phantom{0}58.23 \\
 & \wpls{} & \textbf{\phantom{00}2.00$\pm$\phantom{00}0.00} & \textbf{\phantom{00}2.00$\pm$\phantom{00}0.00} & 117.31$\pm$\phantom{0}13.01 & \phantom{0}67.86$\pm$\phantom{0}29.50 & 138.03$\pm$\phantom{0}25.08 & \phantom{0}94.45$\pm$\phantom{0}23.34 & 134.97$\pm$\phantom{0}31.94 & 100.78$\pm$\phantom{0}48.92 \\
\hline
\end{tabular}%
}
\end{table}

\begin{table}[h!]
\centering
\caption{Directional landmark localization error on HCP T2w--T1w under the \textbf{Dataset-Fit} setting using top-1 nearest-neighbor matching in feature space ($k=1$, $L_2$ distance). Distances are reported in millimeters as the mean$\pm$standard deviation across landmarks, where each landmark value is first computed as the median over held-out query--key pairs within each direction-specific group. Lower is better. Bold marks the best result per column.}
\label{tab:landmark_directional_hcpt2t1_medianpair}
\vspace{4pt}
\scriptsize
\setlength{\tabcolsep}{2.0pt}
\resizebox{\textwidth}{!}{%
\begin{tabular}{l|l||cc|cc|cc|cc}
\hline
\multirow{2}{*}{\textbf{Encoder}} & \multirow{2}{*}{\textbf{Method}} & \multicolumn{2}{c|}{\textbf{Self-Consistency}} & \multicolumn{2}{c|}{\textbf{Different Subject}} & \multicolumn{2}{c|}{\textbf{Different Modality}} & \multicolumn{2}{c}{\textbf{Generalization}} \\
 &  & T2w $\downarrow$ & T1w $\downarrow$ & T2w $\downarrow$ & T1w $\downarrow$ & T2w$\to$T1w $\downarrow$ & T1w$\to$T2w $\downarrow$ & T2w$\to$T1w $\downarrow$ & T1w$\to$T2w $\downarrow$ \\
\hline\hline
\multirow{3}{*}{--} & MIND & 22.80$\pm$24.01 & 22.21$\pm$27.15 & 58.17$\pm$\phantom{0}8.47 & 54.62$\pm$\phantom{0}9.21 & 53.19$\pm$\phantom{0}8.53 & 58.71$\pm$\phantom{0}8.62 & 55.97$\pm$10.87 & 58.15$\pm$\phantom{0}8.59 \\
 & Anatomix & \phantom{0}0.73$\pm$\phantom{0}0.08 & \phantom{0}0.72$\pm$\phantom{0}0.08 & 21.16$\pm$11.73 & 17.19$\pm$\phantom{0}9.93 & 21.56$\pm$13.80 & 22.67$\pm$14.91 & 28.34$\pm$13.40 & 26.95$\pm$11.94 \\
 & Anatomix+MIND & \phantom{0}1.85$\pm$\phantom{0}2.96 & \phantom{0}1.65$\pm$\phantom{0}1.97 & 26.87$\pm$10.89 & 21.83$\pm$14.50 & 24.01$\pm$14.98 & 28.34$\pm$12.00 & 30.01$\pm$11.93 & 29.12$\pm$12.53 \\
\hline
\multirow{5}{*}{DINOv2} & Axial & \phantom{0}0.89$\pm$\phantom{0}0.20 & \phantom{0}0.92$\pm$\phantom{0}0.20 & \phantom{0}4.47$\pm$\phantom{0}1.04 & \phantom{0}4.31$\pm$\phantom{0}1.08 & \phantom{0}7.22$\pm$\phantom{0}4.02 & \phantom{0}7.08$\pm$\phantom{0}5.07 & \phantom{0}7.81$\pm$\phantom{0}3.12 & \phantom{0}7.82$\pm$\phantom{0}3.85 \\
 & Sagittal & \phantom{0}0.91$\pm$\phantom{0}0.23 & \phantom{0}0.89$\pm$\phantom{0}0.20 & \phantom{0}5.20$\pm$\phantom{0}1.80 & \phantom{0}5.30$\pm$\phantom{0}1.14 & 10.01$\pm$\phantom{0}8.84 & 10.43$\pm$\phantom{0}6.25 & 11.55$\pm$\phantom{0}8.35 & 11.49$\pm$\phantom{0}5.18 \\
 & Coronal & \phantom{0}0.93$\pm$\phantom{0}0.11 & \phantom{0}0.94$\pm$\phantom{0}0.19 & \phantom{0}5.95$\pm$\phantom{0}2.65 & \phantom{0}5.44$\pm$\phantom{0}1.51 & 13.48$\pm$\phantom{0}9.37 & 10.03$\pm$\phantom{0}5.27 & 16.54$\pm$\phantom{0}9.19 & 10.68$\pm$\phantom{0}4.46 \\
 & \pcatd{} & \textbf{\phantom{0}0.70$\pm$\phantom{0}0.00} & \textbf{\phantom{0}0.70$\pm$\phantom{0}0.00} & \phantom{0}3.95$\pm$\phantom{0}1.09 & \phantom{0}4.07$\pm$\phantom{0}0.88 & \phantom{0}7.08$\pm$\phantom{0}3.78 & \phantom{0}5.60$\pm$\phantom{0}3.48 & \phantom{0}8.33$\pm$\phantom{0}4.26 & \phantom{0}6.69$\pm$\phantom{0}2.53 \\
 & \wpls{} & \textbf{\phantom{0}0.70$\pm$\phantom{0}0.00} & \textbf{\phantom{0}0.70$\pm$\phantom{0}0.00} & \phantom{0}4.00$\pm$\phantom{0}1.01 & \phantom{0}4.00$\pm$\phantom{0}1.05 & \phantom{0}3.86$\pm$\phantom{0}1.26 & \phantom{0}4.14$\pm$\phantom{0}1.23 & \phantom{0}5.58$\pm$\phantom{0}1.57 & \phantom{0}5.91$\pm$\phantom{0}1.82 \\
\hline
\multirow{5}{*}{DINOv3} & Axial & \phantom{0}1.36$\pm$\phantom{0}0.32 & \phantom{0}1.36$\pm$\phantom{0}0.32 & \phantom{0}3.99$\pm$\phantom{0}0.99 & \phantom{0}4.46$\pm$\phantom{0}1.24 & \phantom{0}7.86$\pm$\phantom{0}6.95 & \phantom{0}5.67$\pm$\phantom{0}3.79 & \phantom{0}8.86$\pm$\phantom{0}6.81 & \phantom{0}7.02$\pm$\phantom{0}2.75 \\
 & Sagittal & \phantom{0}1.39$\pm$\phantom{0}0.26 & \phantom{0}1.39$\pm$\phantom{0}0.26 & \phantom{0}4.62$\pm$\phantom{0}0.92 & \phantom{0}4.38$\pm$\phantom{0}0.72 & \phantom{0}7.42$\pm$\phantom{0}6.25 & \phantom{0}6.86$\pm$\phantom{0}4.00 & \phantom{0}8.49$\pm$\phantom{0}6.77 & \phantom{0}8.31$\pm$\phantom{0}3.54 \\
 & Coronal & \phantom{0}1.44$\pm$\phantom{0}0.19 & \phantom{0}1.44$\pm$\phantom{0}0.19 & \phantom{0}4.57$\pm$\phantom{0}0.95 & \phantom{0}4.69$\pm$\phantom{0}1.05 & \phantom{0}5.20$\pm$\phantom{0}2.47 & \phantom{0}5.56$\pm$\phantom{0}2.78 & \phantom{0}7.59$\pm$\phantom{0}2.41 & \phantom{0}8.09$\pm$\phantom{0}6.19 \\
 & \pcatd{} & \textbf{\phantom{0}0.70$\pm$\phantom{0}0.00} & \textbf{\phantom{0}0.70$\pm$\phantom{0}0.00} & \phantom{0}3.83$\pm$\phantom{0}1.04 & \phantom{0}3.69$\pm$\phantom{0}1.04 & \phantom{0}6.27$\pm$\phantom{0}4.37 & \phantom{0}5.25$\pm$\phantom{0}2.20 & \phantom{0}7.11$\pm$\phantom{0}3.64 & \phantom{0}6.62$\pm$\phantom{0}2.16 \\
 & \wpls{} & \textbf{\phantom{0}0.70$\pm$\phantom{0}0.00} & \textbf{\phantom{0}0.70$\pm$\phantom{0}0.00} & \textbf{\phantom{0}3.58$\pm$\phantom{0}0.86} & \textbf{\phantom{0}3.63$\pm$\phantom{0}1.04} & \textbf{\phantom{0}2.67$\pm$\phantom{0}0.67} & \textbf{\phantom{0}3.15$\pm$\phantom{0}0.75} & \textbf{\phantom{0}4.48$\pm$\phantom{0}0.84} & \textbf{\phantom{0}4.48$\pm$\phantom{0}0.89} \\
\hline
\multirow{5}{*}{MedSAM2} & Axial & \phantom{0}1.36$\pm$\phantom{0}0.32 & \phantom{0}1.36$\pm$\phantom{0}0.32 & 13.15$\pm$\phantom{0}9.79 & 11.34$\pm$\phantom{0}6.27 & 39.30$\pm$13.93 & 36.06$\pm$15.23 & 43.18$\pm$15.57 & 38.78$\pm$12.06 \\
 & Sagittal & \phantom{0}1.39$\pm$\phantom{0}0.26 & \phantom{0}1.39$\pm$\phantom{0}0.26 & 13.36$\pm$10.51 & \phantom{0}9.15$\pm$\phantom{0}6.09 & 34.16$\pm$15.22 & 35.99$\pm$13.35 & 36.61$\pm$10.88 & 36.70$\pm$10.65 \\
 & Coronal & \phantom{0}1.44$\pm$\phantom{0}0.19 & \phantom{0}1.44$\pm$\phantom{0}0.19 & 21.37$\pm$\phantom{0}7.91 & 13.91$\pm$\phantom{0}6.29 & 34.64$\pm$11.78 & 49.66$\pm$18.97 & 34.07$\pm$10.22 & 48.71$\pm$14.98 \\
 & \pcatd{} & \textbf{\phantom{0}0.70$\pm$\phantom{0}0.00} & \textbf{\phantom{0}0.70$\pm$\phantom{0}0.00} & 10.30$\pm$\phantom{0}4.99 & \phantom{0}7.37$\pm$\phantom{0}2.44 & 34.05$\pm$10.50 & 43.54$\pm$16.97 & 35.88$\pm$\phantom{0}9.59 & 41.05$\pm$17.26 \\
 & \wpls{} & \textbf{\phantom{0}0.70$\pm$\phantom{0}0.00} & \textbf{\phantom{0}0.70$\pm$\phantom{0}0.00} & \phantom{0}8.47$\pm$\phantom{0}4.00 & \phantom{0}6.49$\pm$\phantom{0}1.62 & 17.93$\pm$\phantom{0}8.44 & 20.36$\pm$14.47 & 28.03$\pm$\phantom{0}8.44 & 24.78$\pm$12.36 \\
\hline
\multirow{5}{*}{SAM3} & Axial & \phantom{0}0.91$\pm$\phantom{0}0.20 & \phantom{0}0.92$\pm$\phantom{0}0.19 & 16.03$\pm$\phantom{0}7.83 & \phantom{0}9.72$\pm$\phantom{0}4.06 & 33.58$\pm$13.94 & 26.09$\pm$11.22 & 32.57$\pm$\phantom{0}7.56 & 28.46$\pm$10.88 \\
 & Sagittal & \phantom{0}1.06$\pm$\phantom{0}0.21 & \phantom{0}1.04$\pm$\phantom{0}0.20 & 20.34$\pm$11.62 & \phantom{0}8.75$\pm$\phantom{0}3.20 & 32.66$\pm$18.10 & 37.89$\pm$17.13 & 40.94$\pm$16.40 & 37.15$\pm$16.34 \\
 & Coronal & \phantom{0}0.93$\pm$\phantom{0}0.22 & \phantom{0}0.94$\pm$\phantom{0}0.22 & 14.03$\pm$\phantom{0}6.75 & 11.23$\pm$\phantom{0}4.26 & 33.43$\pm$20.45 & 33.08$\pm$19.80 & 38.58$\pm$15.62 & 30.98$\pm$12.66 \\
 & \pcatd{} & \textbf{\phantom{0}0.70$\pm$\phantom{0}0.00} & \textbf{\phantom{0}0.70$\pm$\phantom{0}0.00} & 12.64$\pm$12.14 & \phantom{0}7.36$\pm$\phantom{0}2.02 & 25.45$\pm$13.30 & 22.63$\pm$12.01 & 30.18$\pm$11.41 & 28.20$\pm$13.24 \\
 & \wpls{} & \textbf{\phantom{0}0.70$\pm$\phantom{0}0.00} & \textbf{\phantom{0}0.70$\pm$\phantom{0}0.00} & \phantom{0}7.69$\pm$\phantom{0}2.58 & \phantom{0}5.74$\pm$\phantom{0}1.64 & 15.82$\pm$\phantom{0}9.24 & 13.59$\pm$\phantom{0}6.17 & 17.85$\pm$\phantom{0}8.39 & 15.45$\pm$\phantom{0}5.43 \\
\hline
\end{tabular}%
}
\end{table}

\newpage

\subsection{Directional Landmark Localization with Pooled-Mean Aggregation}
\label{sec:app:landmark_pooledmean}

\begin{table}[h!]
\centering
\caption{Directional landmark localization error on Abdomen MR--CT under the \textbf{Dataset-Fit} setting using top-1 nearest-neighbor matching in feature space ($k=1$, $L_2$ distance). Distances are reported in millimeters as the mean$\pm$standard deviation, pooled over landmarks and held-out pairs within each direction-specific group. Lower is better. Bold marks the best result per column.}
\label{tab:landmark_directional_abdmrct}
\vspace{4pt}
\scriptsize
\setlength{\tabcolsep}{2.0pt}
\resizebox{\textwidth}{!}{%
\begin{tabular}{l|l||cc|cc|cc|cc}
\hline
\multirow{2}{*}{\textbf{Encoder}} & \multirow{2}{*}{\textbf{Method}} & \multicolumn{2}{c|}{\textbf{Self-Consistency}} & \multicolumn{2}{c|}{\textbf{Different Subject}} & \multicolumn{2}{c|}{\textbf{Different Modality}} & \multicolumn{2}{c}{\textbf{Generalization}} \\
 &  & MR $\downarrow$ & CT $\downarrow$ & MR $\downarrow$ & CT $\downarrow$ & MR$\to$CT $\downarrow$ & CT$\to$MR $\downarrow$ & MR$\to$CT $\downarrow$ & CT$\to$MR $\downarrow$ \\
\hline\hline
\multirow{3}{*}{--} & MIND & \phantom{0}99.25$\pm$\phantom{0}85.73 & 118.43$\pm$\phantom{0}92.52 & 172.21$\pm$\phantom{0}55.29 & 168.65$\pm$\phantom{0}62.58 & 149.87$\pm$\phantom{0}64.03 & 187.38$\pm$\phantom{0}61.69 & 161.58$\pm$\phantom{0}63.10 & 158.09$\pm$\phantom{0}66.04 \\
 & Anatomix & \phantom{00}2.61$\pm$\phantom{00}0.93 & \phantom{00}2.39$\pm$\phantom{00}0.55 & \phantom{0}53.10$\pm$\phantom{0}33.60 & \phantom{0}43.86$\pm$\phantom{0}33.33 & \phantom{0}63.43$\pm$\phantom{0}52.10 & \phantom{0}50.19$\pm$\phantom{0}37.71 & \phantom{0}78.11$\pm$\phantom{0}39.67 & \phantom{0}50.14$\pm$\phantom{0}35.06 \\
 & Anatomix+MIND & \phantom{0}11.07$\pm$\phantom{0}24.11 & \phantom{0}13.22$\pm$\phantom{0}21.26 & \phantom{0}68.13$\pm$\phantom{0}50.82 & \phantom{0}49.32$\pm$\phantom{0}37.98 & \phantom{0}65.44$\pm$\phantom{0}47.34 & \phantom{0}59.47$\pm$\phantom{0}45.98 & \phantom{0}80.28$\pm$\phantom{0}44.04 & \phantom{0}57.65$\pm$\phantom{0}37.79 \\
\hline
\multirow{5}{*}{DINOv2} & Axial & \phantom{00}2.42$\pm$\phantom{00}0.80 & \phantom{00}2.54$\pm$\phantom{00}0.72 & \phantom{0}45.17$\pm$\phantom{0}47.17 & \phantom{0}43.74$\pm$\phantom{0}47.03 & \phantom{0}49.80$\pm$\phantom{0}69.61 & \phantom{0}60.34$\pm$\phantom{0}65.85 & \phantom{0}53.21$\pm$\phantom{0}60.01 & \phantom{0}68.93$\pm$\phantom{0}60.81 \\
 & Sagittal & \phantom{00}2.59$\pm$\phantom{00}0.83 & \phantom{00}2.53$\pm$\phantom{00}0.91 & 103.63$\pm$\phantom{0}71.50 & \phantom{0}80.74$\pm$\phantom{0}75.51 & \phantom{0}93.36$\pm$\phantom{0}77.24 & 134.80$\pm$\phantom{0}81.32 & 130.33$\pm$\phantom{0}66.61 & 124.69$\pm$\phantom{0}81.01 \\
 & Coronal & \phantom{00}2.55$\pm$\phantom{00}0.88 & \phantom{00}2.57$\pm$\phantom{00}0.86 & \phantom{0}54.16$\pm$\phantom{0}43.90 & \phantom{0}40.68$\pm$\phantom{0}31.93 & \phantom{0}60.75$\pm$\phantom{0}38.80 & \phantom{0}50.19$\pm$\phantom{0}33.18 & \phantom{0}60.63$\pm$\phantom{0}44.26 & \phantom{0}57.63$\pm$\phantom{0}45.23 \\
 & \pcatd{} & \phantom{00}2.19$\pm$\phantom{00}0.59 & \phantom{00}2.03$\pm$\phantom{00}0.15 & \phantom{0}32.41$\pm$\phantom{0}18.95 & \phantom{0}37.39$\pm$\phantom{0}44.63 & \phantom{0}37.18$\pm$\phantom{0}26.88 & \phantom{0}36.72$\pm$\phantom{0}40.25 & \phantom{0}37.66$\pm$\phantom{0}23.26 & \phantom{0}34.18$\pm$\phantom{0}19.94 \\
 & \wpls{} & \phantom{00}2.06$\pm$\phantom{00}0.35 & \textbf{\phantom{00}2.00$\pm$\phantom{00}0.00} & \phantom{0}35.04$\pm$\phantom{0}19.44 & \phantom{0}27.56$\pm$\phantom{0}18.84 & \phantom{0}29.91$\pm$\phantom{0}22.17 & \phantom{0}29.06$\pm$\phantom{0}22.17 & \textbf{\phantom{0}30.54$\pm$\phantom{0}18.98} & \phantom{0}35.80$\pm$\phantom{0}34.71 \\
\hline
\multirow{5}{*}{DINOv3} & Axial & \phantom{00}2.75$\pm$\phantom{00}0.97 & \phantom{00}2.49$\pm$\phantom{00}0.87 & \phantom{0}37.80$\pm$\phantom{0}36.09 & \phantom{0}36.38$\pm$\phantom{0}30.60 & \phantom{0}45.33$\pm$\phantom{0}53.63 & \phantom{0}35.24$\pm$\phantom{0}43.11 & \phantom{0}34.80$\pm$\phantom{0}28.79 & \phantom{0}54.69$\pm$\phantom{0}48.20 \\
 & Sagittal & \phantom{00}2.49$\pm$\phantom{00}0.76 & \phantom{00}2.44$\pm$\phantom{00}0.69 & 102.98$\pm$\phantom{0}69.22 & \phantom{0}69.96$\pm$\phantom{0}66.83 & 109.60$\pm$\phantom{0}84.17 & \phantom{0}94.58$\pm$\phantom{0}88.77 & 114.98$\pm$\phantom{0}88.69 & 110.99$\pm$\phantom{0}78.16 \\
 & Coronal & \phantom{00}2.96$\pm$\phantom{00}0.95 & \phantom{00}2.90$\pm$\phantom{00}1.05 & \phantom{0}68.80$\pm$\phantom{0}52.02 & \phantom{0}28.04$\pm$\phantom{0}16.48 & \phantom{0}59.96$\pm$\phantom{0}55.84 & \phantom{0}45.82$\pm$\phantom{0}45.12 & \phantom{0}75.06$\pm$\phantom{0}57.12 & \phantom{0}55.33$\pm$\phantom{0}40.74 \\
 & \pcatd{} & \textbf{\phantom{00}2.00$\pm$\phantom{00}0.00} & \phantom{00}2.03$\pm$\phantom{00}0.15 & \phantom{0}32.93$\pm$\phantom{0}25.32 & \textbf{\phantom{0}21.37$\pm$\phantom{0}13.47} & \phantom{0}32.14$\pm$\phantom{0}41.75 & \textbf{\phantom{0}24.34$\pm$\phantom{0}15.50} & \phantom{0}32.83$\pm$\phantom{0}35.36 & \phantom{0}33.38$\pm$\phantom{0}20.70 \\
 & \wpls{} & \phantom{00}2.11$\pm$\phantom{00}0.40 & \phantom{00}2.03$\pm$\phantom{00}0.15 & \textbf{\phantom{0}32.11$\pm$\phantom{0}26.93} & \phantom{0}24.09$\pm$\phantom{0}15.19 & \textbf{\phantom{0}26.40$\pm$\phantom{0}24.58} & \phantom{0}24.67$\pm$\phantom{0}13.86 & \phantom{0}32.75$\pm$\phantom{0}35.44 & \textbf{\phantom{0}32.54$\pm$\phantom{0}21.83} \\
\hline
\multirow{5}{*}{MedSAM2} & Axial & \phantom{00}2.75$\pm$\phantom{00}0.97 & \phantom{00}2.46$\pm$\phantom{00}0.87 & \phantom{0}97.68$\pm$\phantom{0}63.99 & \phantom{0}83.20$\pm$\phantom{0}57.05 & 114.54$\pm$\phantom{0}69.05 & 121.31$\pm$\phantom{0}70.26 & 110.68$\pm$\phantom{0}63.13 & 102.07$\pm$\phantom{0}51.46 \\
 & Sagittal & \phantom{00}2.49$\pm$\phantom{00}0.76 & \phantom{00}2.44$\pm$\phantom{00}0.69 & 145.86$\pm$\phantom{0}81.30 & 110.32$\pm$\phantom{0}70.20 & 111.54$\pm$\phantom{0}69.02 & 117.02$\pm$\phantom{0}83.39 & 140.01$\pm$\phantom{0}73.53 & 146.51$\pm$\phantom{0}74.14 \\
 & Coronal & \phantom{00}2.98$\pm$\phantom{00}0.94 & \phantom{00}2.90$\pm$\phantom{00}1.05 & 123.20$\pm$\phantom{0}54.29 & \phantom{0}93.75$\pm$\phantom{0}64.15 & 105.66$\pm$\phantom{0}57.12 & 108.27$\pm$\phantom{0}66.82 & 119.83$\pm$\phantom{0}47.86 & 123.81$\pm$\phantom{0}52.37 \\
 & \pcatd{} & \phantom{00}2.06$\pm$\phantom{00}0.35 & \phantom{00}2.03$\pm$\phantom{00}0.15 & 104.47$\pm$\phantom{0}51.64 & \phantom{0}85.54$\pm$\phantom{0}65.06 & 110.57$\pm$\phantom{0}61.38 & \phantom{0}88.56$\pm$\phantom{0}63.63 & 109.19$\pm$\phantom{0}57.23 & 105.51$\pm$\phantom{0}75.34 \\
 & \wpls{} & \phantom{00}2.03$\pm$\phantom{00}0.15 & \phantom{00}2.13$\pm$\phantom{00}0.52 & \phantom{0}84.71$\pm$\phantom{0}49.27 & \phantom{0}76.44$\pm$\phantom{0}57.89 & \phantom{0}96.12$\pm$\phantom{0}57.74 & \phantom{0}93.83$\pm$\phantom{0}74.22 & \phantom{0}82.75$\pm$\phantom{0}61.48 & \phantom{0}91.48$\pm$\phantom{0}73.23 \\
\hline
\multirow{5}{*}{SAM3} & Axial & \phantom{00}2.70$\pm$\phantom{00}0.99 & \phantom{00}2.44$\pm$\phantom{00}0.79 & 134.88$\pm$\phantom{0}64.62 & \phantom{0}90.26$\pm$\phantom{0}63.96 & 120.14$\pm$\phantom{0}69.35 & 112.87$\pm$\phantom{0}71.19 & 125.57$\pm$\phantom{0}67.76 & 121.31$\pm$\phantom{0}63.71 \\
 & Sagittal & \phantom{00}2.32$\pm$\phantom{00}0.49 & \phantom{00}2.19$\pm$\phantom{00}0.36 & 146.24$\pm$\phantom{0}70.13 & 100.28$\pm$\phantom{0}69.46 & 148.97$\pm$\phantom{0}64.43 & 149.34$\pm$\phantom{0}74.14 & 129.35$\pm$\phantom{0}66.79 & 157.03$\pm$\phantom{0}61.14 \\
 & Coronal & \phantom{00}2.98$\pm$\phantom{00}0.94 & \phantom{00}2.87$\pm$\phantom{00}1.06 & 149.47$\pm$\phantom{0}63.94 & 101.90$\pm$\phantom{0}69.71 & 154.02$\pm$\phantom{0}42.02 & 137.05$\pm$\phantom{0}57.21 & 153.19$\pm$\phantom{0}44.93 & 140.64$\pm$\phantom{0}60.42 \\
 & \pcatd{} & \phantom{00}2.06$\pm$\phantom{00}0.35 & \phantom{00}2.07$\pm$\phantom{00}0.37 & 121.52$\pm$\phantom{0}69.88 & \phantom{0}86.03$\pm$\phantom{0}63.03 & 137.84$\pm$\phantom{0}68.77 & 104.42$\pm$\phantom{0}56.07 & 144.58$\pm$\phantom{0}68.80 & 123.53$\pm$\phantom{0}74.73 \\
 & \wpls{} & \phantom{00}2.12$\pm$\phantom{00}0.49 & \phantom{00}2.07$\pm$\phantom{00}0.37 & 116.83$\pm$\phantom{0}52.74 & \phantom{0}82.70$\pm$\phantom{0}61.93 & 133.65$\pm$\phantom{0}74.55 & 101.03$\pm$\phantom{0}59.18 & 136.17$\pm$\phantom{0}68.44 & 109.90$\pm$\phantom{0}70.60 \\
\hline
\end{tabular}%
}
\end{table}

\begin{table}[h!]
\centering
\caption{Directional landmark localization error on HCP T2w--T1w under the \textbf{Dataset-Fit} setting using top-1 nearest-neighbor matching in feature space ($k=1$, $L_2$ distance). Distances are reported in millimeters as the mean$\pm$standard deviation, pooled over landmarks and held-out pairs within each direction-specific group. Lower is better. Bold marks the best result per column.}
\label{tab:landmark_directional_hcpt2t1}
\vspace{4pt}
\scriptsize
\setlength{\tabcolsep}{2.0pt}
\resizebox{\textwidth}{!}{%
\begin{tabular}{l|l||cc|cc|cc|cc}
\hline
\multirow{2}{*}{\textbf{Encoder}} & \multirow{2}{*}{\textbf{Method}} & \multicolumn{2}{c|}{\textbf{Self-Consistency}} & \multicolumn{2}{c|}{\textbf{Different Subject}} & \multicolumn{2}{c|}{\textbf{Different Modality}} & \multicolumn{2}{c}{\textbf{Generalization}} \\
 &  & T2w $\downarrow$ & T1w $\downarrow$ & T2w $\downarrow$ & T1w $\downarrow$ & T2w$\to$T1w $\downarrow$ & T1w$\to$T2w $\downarrow$ & T2w$\to$T1w $\downarrow$ & T1w$\to$T2w $\downarrow$ \\
\hline\hline
\multirow{3}{*}{--} & MIND & 30.34$\pm$33.77 & 30.10$\pm$33.30 & 55.98$\pm$23.33 & 52.02$\pm$26.26 & 52.39$\pm$24.39 & 55.88$\pm$25.33 & 55.70$\pm$23.47 & 55.90$\pm$23.28 \\
 & Anatomix & \phantom{0}1.24$\pm$\phantom{0}3.99 & \phantom{0}1.40$\pm$\phantom{0}3.88 & 23.21$\pm$19.68 & 20.39$\pm$18.87 & 24.31$\pm$20.07 & 23.66$\pm$20.25 & 29.30$\pm$21.39 & 27.85$\pm$19.86 \\
 & Anatomix+MIND & \phantom{0}8.27$\pm$16.91 & \phantom{0}6.10$\pm$13.10 & 29.28$\pm$22.33 & 25.51$\pm$21.93 & 28.23$\pm$23.10 & 29.60$\pm$22.77 & 30.88$\pm$21.18 & 31.39$\pm$21.55 \\
\hline
\multirow{5}{*}{DINOv2} & Axial & \phantom{0}1.01$\pm$\phantom{0}0.36 & \phantom{0}1.02$\pm$\phantom{0}0.36 & \phantom{0}5.76$\pm$\phantom{0}5.12 & \phantom{0}5.23$\pm$\phantom{0}4.06 & \phantom{0}9.07$\pm$\phantom{0}7.59 & \phantom{0}8.67$\pm$\phantom{0}8.01 & 10.64$\pm$\phantom{0}9.77 & \phantom{0}9.84$\pm$\phantom{0}7.41 \\
 & Sagittal & \phantom{0}1.03$\pm$\phantom{0}0.41 & \phantom{0}1.00$\pm$\phantom{0}0.35 & \phantom{0}6.74$\pm$\phantom{0}6.01 & \phantom{0}6.26$\pm$\phantom{0}4.23 & 12.29$\pm$13.20 & 11.98$\pm$10.23 & 14.34$\pm$13.33 & 13.19$\pm$\phantom{0}9.46 \\
 & Coronal & \phantom{0}1.03$\pm$\phantom{0}0.35 & \phantom{0}1.03$\pm$\phantom{0}0.35 & \phantom{0}8.55$\pm$\phantom{0}9.92 & \phantom{0}6.86$\pm$\phantom{0}5.89 & 17.07$\pm$17.64 & 12.49$\pm$12.54 & 20.25$\pm$20.32 & 13.91$\pm$12.42 \\
 & \pcatd{} & \phantom{0}0.73$\pm$\phantom{0}0.13 & \phantom{0}0.74$\pm$\phantom{0}0.15 & \phantom{0}4.48$\pm$\phantom{0}2.67 & \phantom{0}4.65$\pm$\phantom{0}2.77 & \phantom{0}7.89$\pm$\phantom{0}5.62 & \phantom{0}6.41$\pm$\phantom{0}5.19 & \phantom{0}8.82$\pm$\phantom{0}5.87 & \phantom{0}7.70$\pm$\phantom{0}5.36 \\
 & \wpls{} & \phantom{0}0.73$\pm$\phantom{0}0.13 & \phantom{0}0.74$\pm$\phantom{0}0.17 & \phantom{0}4.47$\pm$\phantom{0}2.65 & \phantom{0}4.72$\pm$\phantom{0}3.04 & \phantom{0}4.71$\pm$\phantom{0}3.80 & \phantom{0}4.73$\pm$\phantom{0}3.36 & \phantom{0}6.54$\pm$\phantom{0}4.50 & \phantom{0}6.26$\pm$\phantom{0}3.90 \\
\hline
\multirow{5}{*}{DINOv3} & Axial & \phantom{0}1.39$\pm$\phantom{0}0.57 & \phantom{0}1.39$\pm$\phantom{0}0.57 & \phantom{0}4.72$\pm$\phantom{0}3.10 & \phantom{0}4.99$\pm$\phantom{0}3.12 & \phantom{0}8.10$\pm$\phantom{0}7.61 & \phantom{0}6.66$\pm$\phantom{0}5.67 & \phantom{0}9.40$\pm$\phantom{0}8.17 & \phantom{0}8.22$\pm$\phantom{0}6.08 \\
 & Sagittal & \phantom{0}1.40$\pm$\phantom{0}0.60 & \phantom{0}1.40$\pm$\phantom{0}0.60 & \phantom{0}5.28$\pm$\phantom{0}3.16 & \phantom{0}4.76$\pm$\phantom{0}2.76 & \phantom{0}7.63$\pm$\phantom{0}7.06 & \phantom{0}7.33$\pm$\phantom{0}5.60 & \phantom{0}9.25$\pm$\phantom{0}7.70 & \phantom{0}8.78$\pm$\phantom{0}5.61 \\
 & Coronal & \phantom{0}1.42$\pm$\phantom{0}0.57 & \phantom{0}1.42$\pm$\phantom{0}0.57 & \phantom{0}5.67$\pm$\phantom{0}4.81 & \phantom{0}5.88$\pm$\phantom{0}6.05 & \phantom{0}7.33$\pm$\phantom{0}8.18 & \phantom{0}7.85$\pm$\phantom{0}8.77 & \phantom{0}9.63$\pm$\phantom{0}8.94 & \phantom{0}9.49$\pm$\phantom{0}9.16 \\
 & \pcatd{} & \phantom{0}0.75$\pm$\phantom{0}0.18 & \phantom{0}0.75$\pm$\phantom{0}0.19 & \phantom{0}4.32$\pm$\phantom{0}2.57 & \phantom{0}4.06$\pm$\phantom{0}2.33 & \phantom{0}6.54$\pm$\phantom{0}5.29 & \phantom{0}6.03$\pm$\phantom{0}4.27 & \phantom{0}7.68$\pm$\phantom{0}4.89 & \phantom{0}7.24$\pm$\phantom{0}4.48 \\
 & \wpls{} & \phantom{0}0.74$\pm$\phantom{0}0.15 & \phantom{0}0.76$\pm$\phantom{0}0.19 & \textbf{\phantom{0}3.95$\pm$\phantom{0}2.28} & \textbf{\phantom{0}3.92$\pm$\phantom{0}2.19} & \textbf{\phantom{0}3.06$\pm$\phantom{0}2.13} & \textbf{\phantom{0}3.44$\pm$\phantom{0}2.24} & \textbf{\phantom{0}4.81$\pm$\phantom{0}2.81} & \textbf{\phantom{0}4.98$\pm$\phantom{0}2.88} \\
\hline
\multirow{5}{*}{MedSAM2} & Axial & \phantom{0}1.39$\pm$\phantom{0}0.57 & \phantom{0}1.39$\pm$\phantom{0}0.57 & 18.82$\pm$18.78 & 17.72$\pm$18.25 & 39.43$\pm$20.40 & 37.80$\pm$22.44 & 42.21$\pm$20.32 & 38.58$\pm$21.03 \\
 & Sagittal & \phantom{0}1.40$\pm$\phantom{0}0.60 & \phantom{0}1.40$\pm$\phantom{0}0.60 & 19.27$\pm$20.78 & 12.45$\pm$13.71 & 38.03$\pm$21.29 & 37.29$\pm$22.39 & 39.38$\pm$20.26 & 39.28$\pm$22.22 \\
 & Coronal & \phantom{0}1.42$\pm$\phantom{0}0.57 & \phantom{0}1.42$\pm$\phantom{0}0.57 & 24.21$\pm$19.36 & 21.34$\pm$20.66 & 37.86$\pm$21.95 & 50.68$\pm$27.19 & 38.54$\pm$20.81 & 49.48$\pm$25.10 \\
 & \pcatd{} & \phantom{0}0.73$\pm$\phantom{0}0.14 & \phantom{0}0.75$\pm$\phantom{0}0.17 & 16.73$\pm$17.58 & 13.46$\pm$16.01 & 34.77$\pm$18.29 & 42.01$\pm$23.87 & 36.77$\pm$17.79 & 42.30$\pm$24.18 \\
 & \wpls{} & \phantom{0}0.73$\pm$\phantom{0}0.14 & \phantom{0}0.75$\pm$\phantom{0}0.19 & 14.43$\pm$16.17 & 11.36$\pm$12.91 & 21.38$\pm$18.24 & 23.28$\pm$20.57 & 28.41$\pm$18.46 & 27.32$\pm$20.91 \\
\hline
\multirow{5}{*}{SAM3} & Axial & \phantom{0}1.02$\pm$\phantom{0}0.35 & \phantom{0}1.02$\pm$\phantom{0}0.36 & 22.69$\pm$21.91 & 14.60$\pm$15.66 & 36.28$\pm$23.60 & 30.28$\pm$21.75 & 35.72$\pm$20.90 & 33.23$\pm$23.08 \\
 & Sagittal & \phantom{0}1.14$\pm$\phantom{0}0.47 & \phantom{0}1.14$\pm$\phantom{0}0.47 & 23.72$\pm$22.11 & 14.76$\pm$17.31 & 37.09$\pm$25.53 & 37.38$\pm$23.94 & 41.14$\pm$24.84 & 38.78$\pm$23.03 \\
 & Coronal & \phantom{0}1.11$\pm$\phantom{0}0.48 & \phantom{0}1.11$\pm$\phantom{0}0.48 & 19.80$\pm$19.24 & 18.11$\pm$18.28 & 37.22$\pm$27.00 & 37.41$\pm$28.72 & 39.25$\pm$23.69 & 37.76$\pm$25.32 \\
 & \pcatd{} & \textbf{\phantom{0}0.72$\pm$\phantom{0}0.13} & \phantom{0}0.73$\pm$\phantom{0}0.15 & 16.77$\pm$19.04 & 11.74$\pm$13.81 & 31.97$\pm$24.10 & 27.98$\pm$21.73 & 34.54$\pm$23.78 & 32.99$\pm$24.21 \\
 & \wpls{} & \phantom{0}0.73$\pm$\phantom{0}0.12 & \textbf{\phantom{0}0.73$\pm$\phantom{0}0.13} & 13.48$\pm$16.70 & \phantom{0}8.66$\pm$\phantom{0}9.88 & 20.42$\pm$19.04 & 16.78$\pm$16.12 & 22.16$\pm$19.90 & 18.62$\pm$16.57 \\
\hline
\end{tabular}%
}
\end{table}

\end{document}